\documentclass{article}

\usepackage[preprint]{neurips_2026}

\usepackage[utf8]{inputenc} % allow utf-8 input
\usepackage[T1]{fontenc}    % use 8-bit T1 fonts
\usepackage{hyperref}
\usepackage{url}            % simple URL typesetting
\usepackage{booktabs}       % professional-quality tables
\usepackage{amsfonts}       % blackboard math symbols
\usepackage{nicefrac}       % compact symbols for 1/2, etc.
\usepackage{microtype}      % microtypography

\usepackage{subcaption}
\usepackage{graphicx}
\usepackage{multirow}
\usepackage{amsmath,amssymb,amsfonts}
\usepackage{amsthm}
\usepackage{mathrsfs}
\usepackage[dvipsnames]{xcolor}
\usepackage{textcomp}
\usepackage{manyfoot}
\usepackage{booktabs}
\usepackage{algorithm}
\usepackage{algorithmicx}
\usepackage{algpseudocode}
\usepackage{listings}
\usepackage{diagbox}
\usepackage{wrapfig}

\usepackage{authblk}

\graphicspath{{figs}}

\usepackage{tikz}
\usetikzlibrary{
  arrows.meta,
  calc,
  fit,
  positioning,
  matrix,
  backgrounds,
  decorations.pathreplacing
}

\tikzset{
  matrix_node/.style={
    matrix of math nodes,
    left delimiter={[},
    right delimiter={]},
    nodes={minimum size=6mm, anchor=center}
  },
  arrow_style/.style={->, thick}
}

\definecolor{ff}{RGB}{193,231,247}   % FeedForward
\definecolor{sa}{RGB}{255,225,188}  % Self-Attention
\definecolor{ln}{RGB}{242,244,192}   % LayerNorm

\definecolor{plotBlue}{RGB}{31,119,180}  % as in matplotlib
\definecolor{plotOrange}{RGB}{255,127,14}  % as in matplotlib

\newtheorem{assumption}{Assumption}

\newtheorem{property}{Property}

\newtheorem{theorem}{Theorem} % continuous numbers

\newtheorem{lemma}[theorem]{Lemma}
\newtheorem{proposition}[theorem]{Proposition}
\newtheorem{definition}[theorem]{Definition}

\title{Closing the Curvature Gap: Full Transformer Hessians}

    \author[1]{Egor Petrov}
    \author[2]{Nikita Kiselev}
    \author[2]{Vladislav Meshkov}
    \author[3]{Andrey Grabovoy}
    
    \affil[1]{Yandex}
    \affil[2]{Independent}
    \affil[3]{Moscow Center for Advanced Studies}
\begin{document}

\maketitle

\begin{abstract}
    The optimization landscape of Transformer models remains poorly understood despite their widespread adoption. While recent studies have derived curvature properties for isolated self-attention mechanisms, a comprehensive theoretical characterization of the full Transformer block, accounting for the interactions between Layer Normalization, Feed-Forward Networks (FFNs), and residual connections, is missing. In this work, we close this gap by deriving the exact, closed-form Hessian for the complete Transformer block under arbitrary twice-differentiable loss functions. We utilize rigorous matrix calculus to handle the non-linearities of LayerNorm and row-wise activations, establishing explicit spectral norm bounds for the resulting Hessian blocks. Our analysis reveals how different architectural components contribute distinct curvature mechanisms, identifying the specific curvature contributions of particular sub-layers. Furthermore, empirical validation against automatic differentiation confirms the exactness of the derived formulas up to numerical precision and shows substantial computational speedups for the closed-form Jacobian evaluations.
\end{abstract}

\section{Introduction} \label{sec:intro}

The Transformer architecture \citep{vaswani2023attentionneed} has become the dominant model family in modern deep learning, with major impact across natural language processing, vision, and multimodal learning \citep{devlin2019bert,brown2020language,dosovitskiy2020image}. Its success, however, comes with a distinctive training recipe. In contrast to many classical neural architectures, practical Transformer training almost always relies on residual connections, Layer Normalization, careful initialization, learning-rate warmup, and adaptive optimizers such as Adam or AdamW \citep{kingma2017adam}. These ingredients are not incidental engineering details: they appear tightly connected to the stability and conditioning of Transformer optimization. Yet the mathematical mechanisms by which the architecture induces these optimization properties remain only partially understood.

A natural object for studying such mechanisms is the Hessian of the training objective. The Hessian describes the local second-order geometry of the loss, determines the quadratic approximation around a parameter point, and governs local curvature, conditioning, and the stability of gradient-based optimization \citep{nocedal2006numericaloptimization,nesterov2013convexoptimization}. In non-convex neural networks, Hessian structure also helps distinguish flat and sharp directions, identify saddle behavior, and explain the strong anisotropy observed in practical loss landscapes \citep{dauphin2014computationalcomplexity,sagun2017eigenvalues,fort2019emergentpropertieslocalgeometry}. For a composite model $\mathcal{L}(\mathbf{F}(\mathbf{w}))$, the Hessian naturally decomposes into a Gauss--Newton term, induced by the curvature of the loss, and a functional term, induced by the curvature of the model map itself \citep{schraudolph2002aproxhessian,martens2020kfac}. The latter is especially important in Transformers, whose architecture contains non-polynomial softmax interactions, normalization operations, residual paths, and nonlinear feed-forward sublayers.

Recent work has begun to uncover the second-order structure of Transformers by deriving Hessian expressions for isolated self-attention mechanisms \citep{ormaniec2024attentionhessian}. This line of analysis has shown that self-attention differs sharply from classical MLP and convolutional layers: the query, key, and value parameter blocks exhibit heterogeneous curvature, different input-growth behavior, and distinct dependencies on the attention mechanism. These results provide important theoretical building blocks. However, a practical Transformer block is not an isolated self-attention layer. Its optimization landscape is shaped by the composition of self-attention with residual connections, Layer Normalization, and Feed-Forward Networks. These components are precisely among the architectural features believed to stabilize Transformer training, but their exact second-order contribution has not been characterized in a unified analytical framework.

In this paper, we derive the exact Hessian structure of a full Transformer block. We consider a post-normalization Transformer block consisting of self-attention, residual connections, Layer Normalization, and a nonlinear FFN. Our analysis is carried out using row-wise vectorization and matrix calculus, allowing all derivatives to be expressed as explicit linear operators with consistent dimensions. We work with an arbitrary twice-differentiable loss function, and we identify the precise way in which the loss Hessian and loss gradient combine with the architecture-induced Jacobians and Hessians.

We summarize our main contributions as follows:
% \begin{enumerate}

1. \textbf{Exact Second-Order Characterization:} We derive closed-form Jacobians and Hessian blocks for the full Transformer block, including self-attention, residual connections, Layer Normalization, and a nonlinear FFN. This extends prior analytical Hessian studies of isolated self-attention to the complete architectural unit.

2. \textbf{LayerNorm and FFN second-order calculus.}
We derive explicit matrix-calculus formulas for the Jacobian and Hessian of row-wise LayerNorm and for the Jacobians and interaction terms of the FFN. These formulas expose how inverse variance, activation gates, and residual streams enter the full Hessian.
    
3. \textbf{Spectral-norm bounds and curvature heterogeneity.}
We establish upper bounds on the spectral norms of the Hessian blocks. The bounds reveal the distinct curvature mechanisms of different sublayers: high-order input dependence in attention, variance-sensitive curvature in LayerNorm, and structured cross-block interactions in the FFN.
    
4. \textbf{Empirical validation and computational advantage.}
We validate the derived formulas against PyTorch automatic differentiation and obtain numerical agreement up to floating-point precision. The closed-form expressions also yield substantial speedups for selected componentwise Jacobian evaluations, supporting their use as practical tools for Transformer curvature analysis.
% \end{enumerate}

% \textcolor{black}{\textbf{Outline.}} The rest of the paper is organized as follows. In Section \ref{sec:rw}, we review related work, categorizing existing research of Transformer optimization and matrix calculus. Section \ref{sec:prelim} establishes the row-wise vectorization notation and reviews the Gauss-Newton decomposition for arbitrary loss functions. In Section \ref{sec:method}, we systematically derive the full Transformer Hessian: beginning with the refined bounds for Self-Attention (Sec. \ref{subsec:hessian_self_attention}), deriving the novel operator calculus for Layer Normalization (Sec. \ref{subsec:layer_norm}) and Feed-Forward Networks (Sec. \ref{subsec:ffn}), and finally assembling the full block Hessian with spectral bounds (Sec. \ref{subsec:hessian_full}). Section \ref{sec:disc} discusses the implications of these bounds for scaling and stability. Detailed proofs and auxiliary lemmas are provided in Appendices \ref{app:properties}--\ref{app:additional_properties}.

\section{Related Work}\label{sec:rw}

\textbf{Hessian geometry and second-order optimization.}
The Hessian is a standard tool for studying local curvature, conditioning, and stability of optimization methods \citep{nocedal2006numericaloptimization,nesterov2013convexoptimization}. In neural networks, Hessian spectra are known to be highly anisotropic, with a few sharp directions and a large near-flat bulk, and have been linked to saddle points, sharpness, generalization, and mode connectivity \citep{dauphin2014computationalcomplexity,sagun2017eigenvalues,fort2019emergentpropertieslocalgeometry,garipov2018losssurfacesmodeconnectivity,draxler2019essentiallynobarriers}. Because exact Hessians are costly in large models, most practical methods rely on approximations such as Gauss--Newton, Hessian-vector products, quasi-Newton methods, K-FAC, or adaptive diagonal preconditioning \citep{pearlmutter1994multiplicationhessian,schraudolph2002aproxhessian,berahas2020lbfgs,martens2020kfac,kingma2017adam}. In contrast, our goal is to derive exact blockwise second-order formulas for the Transformer block.

\textbf{Transformer optimization and curvature.}
Transformers \citep{vaswani2023attentionneed} introduce curvature mechanisms absent in classical MLPs and CNNs, including token-token attention, bilinear query-key interactions, softmax normalization, residual streams, and LayerNorm. These components are closely connected to practical training choices such as adaptive optimization, learning-rate warmup, and normalization placement \citep{xiong2020prelayernorm,noci2022signalpropagationtransformerstheoretical,zhang2024why}. Recent works study Transformer stability, in-context learning, and loss-landscape behavior \citep{li2023transformersalgorithmsgeneralizationstability,zhang2025understandinggeneralizationincontextlearning,hoogland2025losslandscapedegeneracystagewise}, but many analyses remain empirical or omit parts of the full block.

\textbf{Analytical Hessians for attention and our extension.}
The closest work to ours is \citep{ormaniec2024attentionhessian}, which derives the Hessian of an isolated self-attention layer and identifies heterogeneous query, key, and value curvature. We extend this analytical program to the complete post-norm Transformer block by incorporating residual connections, LayerNorm, and the nonlinear FFN. Our derivation supports arbitrary twice-differentiable losses via the Gauss--Newton decomposition and yields explicit spectral-norm bounds for the assembled full-block Hessian.

\section{Preliminaries and General Notation}\label{sec:prelim}
% \subsection{Row-Wise Vectorization}
\paragraph{Row-Wise Vectorization.}
To align with the token-wise processing nature of Transformer architectures, we adopt the row-wise vectorization notation \(\mathrm{vec}_r(\cdot)\), following \citep{ormaniec2024attentionhessian, noci2022signalpropagationtransformerstheoretical}. Formally, for a matrix \(\mathbf{A} \in \mathbb{R}^{m \times n}\), the operator is defined such that the entries of the $i$-th row of $\mathbf{A}$ appear consecutively in the resulting vector. This transformation is visualized in Figure~\ref{fig:prelim_concepts}(a).

\begin{figure}[t]
    \centering
    \begin{minipage}[c]{0.47\textwidth}
        \centering
        \begin{tikzpicture}[
            >=stealth,
            matrix_node/.style={
                matrix of math nodes,
                left delimiter={[},
                right delimiter={]},
                nodes={minimum size=6mm, anchor=center}
            },
            arrow_style/.style={->, thick}
        ]
        % Matrix W
        \matrix (W) [matrix_node] {
            \textcolor{blue}{w_{11}} & \textcolor{blue}{w_{12}} \\
            \textcolor{red}{w_{21}} & \textcolor{red}{w_{22}} \\
        };
        \node[above=0.15cm of W] {$\mathbf{W}$};

        % Vector vec_r(W)
        \matrix (vecW) [matrix_node, right=1.4cm of W] {
            \textcolor{blue}{w_{11}} \\
            \textcolor{blue}{w_{12}} \\
            \textcolor{red}{w_{21}} \\
            \textcolor{red}{w_{22}} \\
        };
        \node[above=0.15cm of vecW] {$\mathrm{vec}_r(\mathbf{W})$};

        \draw[arrow_style, blue!70] ([xshift=3pt]W.east |- W-1-1.center) to[out=0,in=180] ([xshift=-3pt]vecW-1-1.west);
        \draw[arrow_style, blue!70] ([xshift=3pt]W.east |- W-1-1.center) to[out=0,in=180] ([xshift=-3pt]vecW-2-1.west);
        \draw[arrow_style, red!70]  ([xshift=3pt]W.east |- W-2-1.center) to[out=0,in=180] ([xshift=-3pt]vecW-3-1.west);
        \draw[arrow_style, red!70]  ([xshift=3pt]W.east |- W-2-1.center) to[out=0,in=180] ([xshift=-3pt]vecW-4-1.west);

        % Braces
        \draw[decorate, decoration={brace, amplitude=4pt}]
            ([xshift=3pt]vecW-1-1.north east) -- ([xshift=3pt]vecW-2-1.south east)
            node[midway, right=5pt, font=\scriptsize] {Row 1};
        \draw[decorate, decoration={brace, amplitude=4pt}]
            ([xshift=3pt]vecW-3-1.north east) -- ([xshift=3pt]vecW-4-1.south east)
            node[midway, right=5pt, font=\scriptsize] {Row 2};
        \end{tikzpicture}
        \smallskip
        (a) Row-wise vectorization
    \end{minipage}
    \hfill
    \begin{minipage}[c]{0.47\textwidth}
        \centering
        \small
        \renewcommand{\arraystretch}{1.3}
        \begin{tabular}{ll}
            \toprule
            \textbf{Object} & \textbf{Dimension} \\
            \midrule
            $\mathbf{F}$ & $n \times d$ \\
            $\mathrm{vec}_r(\mathbf{F})$ & $nd \times 1$ \\
            $\mathrm{vec}_r(\mathbf{W}_i)$ & $p_i q_i \times 1$ \\
            \midrule
            Jacobian $\frac{\partial \mathbf{F}}{\partial \mathbf{W}_i}$ & $nd \times p_i q_i$ \\
            Hessian $\frac{\partial^2 \mathbf{F}}{\partial \mathbf{W}_i \partial \mathbf{W}_j}$ & $(nd \cdot p_i q_i) \times p_j q_j$ \\
            \bottomrule
        \end{tabular}
        \smallskip
        (b) Dimensionality scheme
    \end{minipage}

    \vspace{0.4cm}
    \caption{\textbf{Notational preliminaries.} (a) Row-wise vectorization $\mathrm{vec}_r(\cdot)$. (b) Dimensionality of derivatives for weights $\mathbf{W}_i \in \mathbb{R}^{p_i \times q_i}$.}
    \label{fig:prelim_concepts}
\end{figure}
% \subsection{Matrix Calculus and Dimensions}
\paragraph{Matrix Calculus and Dimensions.}
We consider matrix-valued functions \(\mathbf{F}: \mathbb{R}^{p \times q} \to \mathbb{R}^{n \times d}\). Determining the dimensions of derivatives for such functions requires systematic unfolding. We verify dimensions using the layout convention where the derivative of a vector with respect to a transposed vector yields the Jacobian matrix. Let \(\mathbf{W}_i \in \mathbb{R}^{p_i \times q_i}\) and \(\mathbf{W}_j \in \mathbb{R}^{p_j \times q_j}\) be weight matrices. The first-order derivative (Jacobian) and the second-order derivative (Hessian Block) are defined via vectorization as follows:
\begin{equation}
    \frac{\partial \mathbf{F}}{\partial \mathbf{W}_i} := \frac{\partial \mathrm{vec}_r(\mathbf{F})}{\partial \mathrm{vec}_r(\mathbf{W}_i)^\top}, \quad 
    \frac{\partial^2 \mathbf{F}}{\partial \mathbf{W}_i \partial \mathbf{W}_j} := \frac{\partial \mathrm{vec}_r\left(\frac{\partial \mathbf{F}}{\partial \mathbf{W}_i}\right)}{\partial \mathrm{vec}_r(\mathbf{W}_j)^\top}.
\end{equation}
The dimensional analysis of these objects is summarized in Figure~\ref{fig:prelim_concepts}(b). 
% \subsection{Hessian Decomposition for Composite Functions}
\paragraph{Hessian Decomposition for Composite Functions.}
We consider an arbitrary, twice-differentiable scalar loss function \(\mathcal{L}: \mathbb{R}^{L \times d_V} \to \mathbb{R}\), which maps the network output to a scalar value (e.g., Mean Squared Error or Cross-Entropy). To analyze the curvature of \(\mathcal{L}(\mathbf{F}(\mathbf{X}; \mathbf{w}))\) with respect to parameter matrices \(\mathbf{W}_i \in \mathbb{R}^{p_i \times q_i}\) and \(\mathbf{W}_j \in \mathbb{R}^{p_j \times q_j}\), we utilize the Gauss-Newton decomposition \citep{schraudolph2002aproxhessian}, following which we obtain the $(i,j)$-th block of the full Hessian $\mathbf{H}$:

\begin{equation} \label{eq:gauss_decomposition}
\frac{\partial^2 (\mathcal{L} \circ \mathbf{F})}{\partial \mathbf{W}_i \partial \mathbf{W}_j} = 
\underbrace{\left(\frac{\partial \mathbf{F}}{\partial \mathbf{W}_i}\right)^\top \frac{\partial^2 \mathcal{L}}{\partial \mathbf{F}^2} \left(\frac{\partial \mathbf{F}}{\partial \mathbf{W}_j}\right)}_{\textbf{Outer Term}} 
+ 
\underbrace{\left( \frac{\partial \mathcal{L}}{\partial \mathbf{F}} \otimes \mathbf{I}_{p_i q_i} \right) \frac{\partial^2 \mathbf{F}}{\partial \mathbf{W}_i \partial \mathbf{W}_j}}_{\textbf{Functional Term}}
\end{equation}

\noindent This decomposition holds for any twice-differentiable composite function. The terms have distinct geometric interpretations:

\textbf{The Outer Term:} This component captures the curvature induced by the loss function $\mathcal{L}$ projected onto the parameter space via the first-order Jacobians of the model. In the neighborhood of a stationary point (where $\nabla \mathcal{L} \to 0$) or in linear models, this term dominates \citep{martens2020kfac}. It is symmetric and Positive Semi-Definite if the loss $\mathcal{L}$ is convex with respect to its inputs $\mathbf{F}$ \citep{boyd2004convex}.
    
\textbf{The Functional Term:} This component involves the gradient of the loss $\frac{\partial \mathcal{L}}{\partial \mathbf{F}}$, contracting with the model's Hessian. For highly non-linear architectures like Transformers, specifically due to LayerNorm and Softmax operations, this term is non-trivial and often breaks the convex property of the total Hessian \citep{dauphin2014computationalcomplexity}.

% In the subsequent sections, we derive analytic expressions for $\frac{\partial^2 \mathbf{F}}{\partial \mathbf{W}_i \partial \mathbf{W}_j}$.
\section{Transformer-block Decomposition} \label{sec:method}
\begin{wrapfigure}{r}{0.3\textwidth}
\centering
\begin{tikzpicture}[
      font=\small,
      >={Latex[length=1mm,width=1mm]},
      layer/.style={rounded corners=2pt, draw=black!70, minimum width=3cm, minimum height=6mm, align=center},
      layernorm/.style={layer, fill=ln},
      selfattn/.style={layer, fill=sa},
      feedforward/.style={layer, fill=ff},
]

\node[rounded corners=6pt, minimum width=40mm, minimum height=39.5mm, fill=black!3, anchor=north west] (transformer_block) {};
\node[above=2mm of transformer_block.north] (title) {\textbf{Transformer Block}};

\node[layer, below=4mm of transformer_block.south, anchor=north] (emb) {Embeddings};
\node[selfattn, above=3mm of transformer_block.south] (sa) {Self-Attention};
\node[layernorm, above=3mm of sa] (ln1) {LayerNorm};
\coordinate (sa_south_east) at ($(sa.south)+(17mm,-2mm)$);

\draw[->] (emb.north) -- (sa.south);
\draw[->] (sa.north) -- (ln1.south);
\draw[->] ($(sa.south)+(0mm,-2mm)$) -- (sa_south_east) |- ($(ln1.east)+(0mm,0mm)$);

\node[feedforward, above=3mm of ln1] (ff) {FeedForward};
\node[layernorm, above=3mm of ff] (ln2) {LayerNorm};
\coordinate (ff_south_east) at ($(ff.south)+(17mm,-2mm)$);

\draw[->] (ln1.north) -- (ff.south);
\draw[->] (ff.north) -- (ln2.south);
\draw[->] (ln2.north) -- ($(ln2.north)+(0mm,5mm)$);
\draw[->] ($(ff.south)+(0mm,-2mm)$) -- (ff_south_east) |- ($(ln2.east)+(0mm,0mm)$);

% \node[above=1mm of transformer_block.north west] {$L\times$};

\end{tikzpicture}
\caption{Schematic of the standard Post-Norm Transformer block architecture.}
\label{fig:transformer}
\end{wrapfigure}

To systematically derive the full Hessian, we first formally define the components of the Transformer block used in our analysis. We consider a standard single-head attention block with post-layer normalization, as depicted in Figure~\ref{fig:transformer}.

Let the input sequence of generic embeddings be \(\mathbf{X} \in \mathbb{R}^{L \times d_V}\). The core mechanism is the Self-Attention (SA) layer, defined as:
\begin{equation} \label{eq:self_attention}
    \mathrm{SA}(\mathbf{X}) = \mathbf{A}(\mathbf{X}) \mathbf{X} \mathbf{W}_V,
\end{equation}
where the attention probability matrix \(\mathbf{A}(\mathbf{X})\) is governed by the query and key projections:
\[
\mathbf{A}(\mathbf{X}) = \mathrm{softmax}\left( \frac{\mathbf{X} \mathbf{W}_Q \mathbf{W}_K^\top \mathbf{X}^\top}{\sqrt{d_K}} \right).
\]
Here, the weight matrices are \(\mathbf{W}_Q, \mathbf{W}_K \in \mathbb{R}^{d_V \times d_K}\) and \(\mathbf{W}_V \in \mathbb{R}^{d_V \times d_V}\). First, the attention output is added to the residual stream and normalized:
\begin{align}
    \mathbf{Y} &= \text{LayerNorm}\Big(\mathbf{X} + \mathrm{SA}(\mathbf{X})\Big). \label{eq:transformer_mid}
\end{align}
Subsequently, this intermediate representation $\mathbf{Y}$ flows through a non-linear Feed-Forward Network ($\mathrm{FFN}$) and a second normalization step to produce the final block output $\mathbf{Z}$:
\begin{align}
    \mathbf{Z} &= \text{LayerNorm}\Big(\mathbf{Y} + \mathrm{FFN}(\mathbf{Y})\Big). \label{eq:transformer}
\end{align}
The $\mathrm{FFN}(\cdot)$ typically consists of two linear transformations separated by a non-linear activation (e.g., ReLU or GELU). The Layer Normalization operation for an arbitrary input $\mathbf{U} \in \mathbb{R}^{m \times n}$ is defined row-wise:
\begin{align} \label{eq:layer_norm}
\text{LayerNorm}(\mathbf{U})_{i,j} = \boldsymbol{\gamma}_j \frac{\mathbf{U}_{i,j} - \mu_i}{\sqrt{\sigma_i^2}} + \boldsymbol{\beta}_j,
\end{align}
where $\mu_i$ and $\sigma_i^2$ are the mean and variance of the $i$-th row. For the sake of Hessian simplicity, we treat the affine parameters $\boldsymbol{\gamma}, \boldsymbol{\beta}$ as constants in our derivation, focusing on the curvature induced by the weight matrices $\mathbf{W}$.

\subsection{Self-Attention}\label{subsec:hessian_self_attention}

We start our analysis in the properties of the isolated Self-Attention (SA) mechanism. Following the foundational derivation from \citep{ormaniec2024attentionhessian}, the Hessian of the loss with respect to attention parameter blocks $\mathbf{W}_i, \mathbf{W}_j \in \{\mathbf{W}_Q, \mathbf{W}_K, \mathbf{W}_V\}$ admits the Gauss-Newton decomposition:
\[
\mathbf{H}(\mathbf{W}_i, \mathbf{W}_j) = 
\mathbf{H}_o(\mathbf{W}_i, \mathbf{W}_j) + \mathbf{H}_f(\mathbf{W}_i, \mathbf{W}_j).
\]
While prior works \citep{ormaniec2024attentionhessian, noci2022signalpropagationtransformerstheoretical}, primarily scrutinized this structure under the assumption of Mean Squared Error (MSE) loss, we observe that the expressions for the Jacobian and second-order kinematic derivatives derived in prior work remain valid for arbitrary twice-differentiable objectives, such as Cross-Entropy, provided the outer loss derivatives are substituted appropriately, which we utilize in our consequent analysis of the full Transformer block.

\paragraph{Hessian's norm estimation.} Next, we introduce a theorem for estimation the spectral norm (Definition \ref{def:matrix_norms}) of the Hessian for a single Self-Attention block.

\begin{theorem}\label{thm:self_attention_hessian_estimation}
Consider a single-head Self-Attention map
$\mathrm{SA}(\mathbf X)=\mathbf A(\mathbf X)\mathbf X\mathbf W_V$ with
$\mathbf A(\mathbf X)=\mathrm{softmax}\!\big(\mathbf X\mathbf W_Q\mathbf W_K^\top \mathbf X^\top/\sqrt{d_K}\big)$,
and an arbitrary twice-differentiable loss $\mathcal L$.
Let $\mathbf H^{(i,j)}$ denote the $(i,j)$-block of the loss Hessian w.r.t.
$\mathbf W_i,\mathbf W_j\in\{\mathbf W_Q,\mathbf W_K,\mathbf W_V\}$.
Then each block admits the Gauss--Newton decomposition
\[
\mathbf H^{(i,j)}=\mathbf H_o^{(i,j)}+\mathbf H_f^{(i,j)},
\]
and its spectral norm is bounded as
\[
\|\mathbf H^{(i,j)}\|_2 \le \|\mathbf H_o^{(i,j)}\|_2+\|\mathbf H_f^{(i,j)}\|_2
\le C_o^{(i,j)} + C_f^{(i,j)}.
\]
Moreover, up to multiplicative factors depending only on
$\|\mathbf W_Q\|_2,\|\mathbf W_K\|_2,\|\mathbf W_V\|_2$, $d_V$, $d_K$
(and on outer loss derivatives), the blocks scale with the input norm as

\[
\begin{array}{c|ccc}
 & Q & K & V \\ \hline
Q & \mathcal O(\|\mathbf X\|_2^{6}) & \mathcal O(\|\mathbf X\|_2^{6}) & \mathcal O(\|\mathbf X\|_2^{5}) \\
K & \cdot                         & \mathcal O(\|\mathbf X\|_2^{6}) & \mathcal O(\|\mathbf X\|_2^{5}) \\
V & \cdot                         & \cdot                         & \mathcal O(\|\mathbf X\|_2^{4})
\end{array}
\]
and the leading functional terms are damped by $\mathcal O(L^{-3})$. In particular, the full self-attention Hessian satisfies
\[
\|\mathbf H\|_2 \le 3 \cdot \max_{i,j\in\{Q,K,V\}}\big(C_o^{(i,j)}+C_f^{(i,j)}\big),
\]
where explicit constants $C_o^{(i,j)},C_f^{(i,j)}$ are given in
Appendix~\ref{app:proof_self_attention_hessian_estimation}.
\end{theorem}

\paragraph{Asymptotic Analysis.}
The bound $M$ reveals critical scaling properties of the optimization landscape. 

 \textbf{Input Sensitivity:} The bound is dominated by terms scaling as $\mathcal{O}(\|\mathbf{X}\|_2^6)$. This extreme sensitivity suggests that pre-LayerNorm placement (controlling $\|\mathbf{X}\|_2$ before attention) is crucial for stabilizing curvature, aligning with empirical observations in \citep{xiong2020prelayernorm}.

 \textbf{Sequence Length:} Several terms scale inversely with sequence length (e.g., $1/L^3$), a consequence of the softmax normalization factor. This implies that for very long sequences, the intrinsic curvature of the attention mechanism (specifically the softmax derivative contribution) may naturally dampen, potentially aiding stability in large-context training.

\subsection{LayerNorm}\label{subsec:layer_norm}

As defined in Eq. (\ref{eq:layer_norm}), Layer Normalization operates row-wise. To facilitate the derivation of second-order properties and ensure compatibility with the matrix calculus properties established in Appendix~\ref{app:properties}, we first reformulated this operation in compact matrix notation.

Let $\mathbf{X} \in \mathbb{R}^{L \times d_V}$. We define the centering operator $\mathbf{M}(\mathbf{X})$ and the inverse-standard-deviation operator $\mathbf{P}(\mathbf{X})$ as:
\[
\mathbf{M}(\mathbf{X}) = \mathbf{X} - \tfrac{1}{d_V}\mathbf{X}\mathbf{1}_{d_V}\mathbf{1}_{d_V}^\top, \quad
\sigma(\mathbf{X}) = \tfrac{1}{\sqrt{d_V}}\big(\mathbf{M}(\mathbf{X})^{\circ 2}\mathbf{1}_{d_V}\big)^{\circ 1/2}, \quad
\mathbf{P}(\mathbf{X}) = \mathrm{diag}^{-1}(\sigma(\mathbf{X})).
\]
Consequently, the LayerNorm operation (excluding affine parameters $\boldsymbol{\gamma}, \boldsymbol{\beta}$ which are treated as constants for the curvature analysis of weights) can be written as:
\[
\text{LayerNorm}(\mathbf{X}) = \mathbf{P}(\mathbf{X}) \mathbf{M}(\mathbf{X}).
\]

\begin{assumption} \label{as:non_zero_variance}
For all input matrices to LayerNorm, the per-row variances satisfy $\min_i \sigma_i^2 > 0$ to ensure differentiability.
\end{assumption}

We note that it's a technical assumption for the proof part simplification and numerical stability. The same effect can be achieved by adding some positive constant to the denominator, but it makes calculations less clear. We now derive the exact Jacobian and Hessian of this operator.

\begin{theorem} [Jacobian of LayerNorm] \label{thm:layernorm_derivative}
The Jacobian of $\text{LayerNorm}(\mathbf{X})$ with respect to $\mathbf{X}$ is given by:
\[
\frac{\partial \,\text{LayerNorm}(\mathbf{X})}{\partial \mathbf{X}}
= (\mathbf{P}(\mathbf{X}) \otimes \mathbf{I}_{d_V})\mathbf{G}
+ (\mathbf{I}_L \otimes \mathbf{M}(\mathbf{X})^\top) \frac{\partial \mathbf{P}(\mathbf{X})}{\partial \mathbf{X}},
\]
where $\mathbf{G} = \frac{\partial \mathbf{M}}{\partial \mathbf{X}} = \mathbf{I}_{Ld_V} - \tfrac{1}{d_V}(\mathbf{I}_L \otimes \mathbf{1}_{d_V \times d_V})$ is the constant centering projection.
The derivative of the inverse deviation $\mathbf{P}$ is:
\[
\frac{\partial \mathbf{P}}{\partial \mathbf{X}}
= \tfrac{1}{\sqrt{d_V}}
\Big( -\mathbf{D}^{-1} \otimes \mathbf{D}^{-\top} \Big)
\mathbf{E}_{diag}
\Big( 
\mathrm{diag}^{-1}\!\big(\mathrm{vec}_r^{1/2}(\mathbf{M}^{\circ 2}\mathbf{1}_{d_V})\big)
(\mathbf{I}_L \otimes \mathbf{1}_{d_V}^\top)
\mathrm{diag}(\mathrm{vec}_r(\mathbf{M}))
\mathbf{G}
\Big),
\]
with $\mathbf{D} = \mathrm{diag}(\sigma(\mathbf{X}))$ and $\mathbf{E}_{diag} = \big( \mathbf{e}_1 \otimes \mathbf{e}_1, \dots, \mathbf{e}_L \otimes \mathbf{e}_L \big)$ representing the derivative of the diagonal operator, revealing the block-diagonal structure of the Jacobian.
\end{theorem}

\begin{figure}[t]
    \centering
    \includegraphics[width=0.7\textwidth]{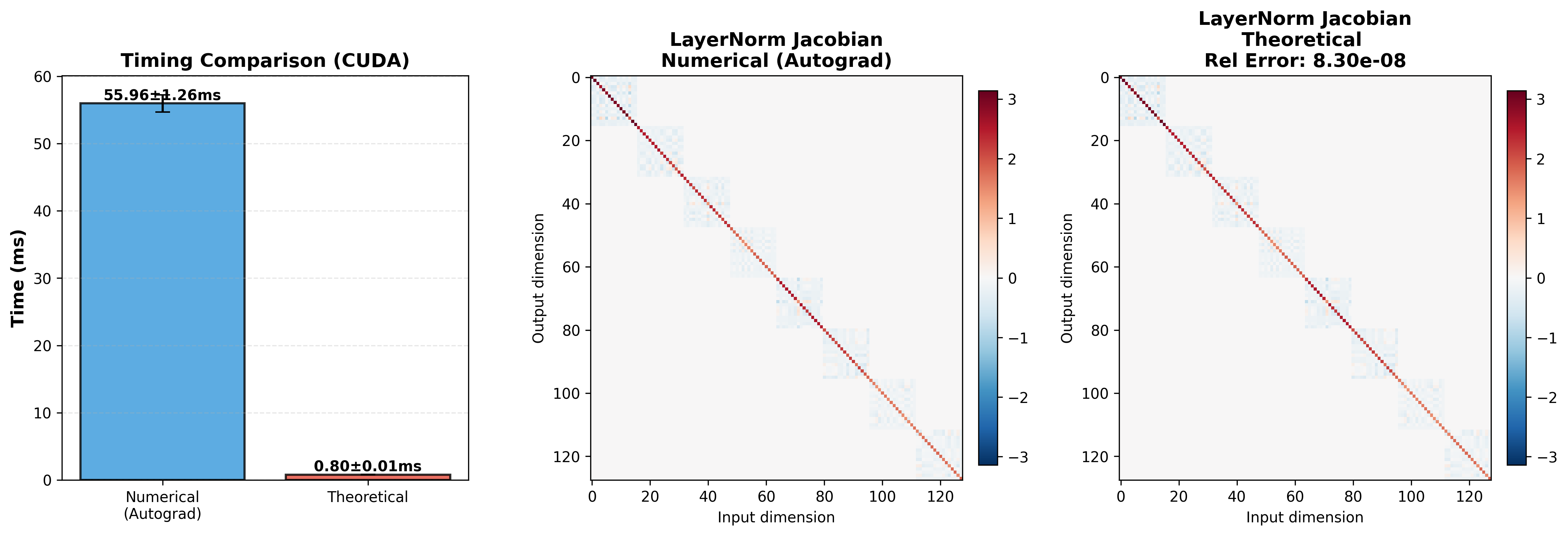}
    \vspace{-0.3em}
    \caption{LayerNorm Jacobian benchmarks. \textbf{Left:} $82\times$ speed‑up. \textbf{Center/Right:} the theoretical formula (Theorem~\ref{thm:layernorm_derivative}) reproduces the autograd Jacobian to a relative error of $8.3\times10^{-8}$.}
    \label{fig:ln}
\end{figure}

\begin{theorem} [Hessian of LayerNorm] \label{thm:layernorm_second_derivative} The Hessian $\frac{\partial^2 \text{LayerNorm}}{\partial \mathbf{X}^2}$ is given by the derivative of the Jacobian terms:
\begin{align*}
\frac{\partial^2 \text{LayerNorm}}{\partial \mathbf{X}^2} &= 
\left( \mathbf{I}_{L d_V} \otimes \mathbf{G}^\top \right) \frac{\partial (\mathbf{P}(\mathbf{X}) \otimes \mathbf{I}_{d_V})}{\partial \mathbf{X}} \\
&+ \left( (\mathbf{I}_L \otimes \mathbf{M}^\top ) \otimes \mathbf{I}_{L d_V} \right) \frac{\partial^2 \mathbf{P}}{\partial \mathbf{X}^2} + \left( \mathbf{I}_{L d_V} \otimes  \left(\frac{\partial \mathbf{P}}{\partial \mathbf{X}}\right)^\top \right) \frac{\partial (\mathbf{I}_L \otimes \mathbf{M}^\top )}{\partial\mathbf{X}}.
\end{align*}
Note that the term involving $\frac{\partial^2 \mathbf{M}}{\partial \mathbf{X}^2}$ vanishes because the centering operation is linear. Explicit forms for the remaining second-order terms are derived in the proof.
\end{theorem}

The detailed proofs for Theorems \ref{thm:layernorm_derivative} and \ref{thm:layernorm_second_derivative} rely on the expansion of Kronecker derivative properties and are provided in Appendices~\ref{app:proof_layernorm_derivative} and \ref{app:proof_layernorm_second_derivative}.

\subsection{Feed-Forward Network} \label{subsec:ffn}

The Feed-Forward Network (FFN) introduces essential non-linearity into the Transformer block. While some prior theoretical works allow for linear approximations of the FFN to simplify signal propagation analysis \citep{noci2022signalpropagationtransformerstheoretical}, we explicitly model the standard non-linear formulation:
\[
\mathrm{FFN}(\mathbf{Y}) = \sigma(\mathbf{Y}\mathbf{W}_1)\mathbf{W}_2,
\]
where $\mathbf{W}_1 \in \mathbb{R}^{d_V \times d_{ff}}$, $\mathbf{W}_2 \in \mathbb{R}^{d_{ff} \times d_V}$, and $\sigma(\cdot) = \mathrm{ReLU}(\cdot)$. To handle the non-differentiability of ReLU at zero, we utilize the following lemma.

\begin{lemma}[ReLU Jacobian and Hessian] \label{lemma:relu_derivative_hessian}
Let $\mathbf{X} \in \mathbb{R}^{m \times n}$. The element-wise $\mathrm{ReLU}$ is differentiable almost everywhere. Its Jacobian and Hessian with respect to the input $\mathbf{X}$ are given by:
\[ 
\frac{\partial \mathrm{ReLU}(\mathbf{X})}{\partial \mathbf{X}} = \mathbf{D}_\sigma(\mathbf{X}) := \mathrm{diag}\!\big(\mathrm{vec}_r(\mathbf{1}_{\{\mathbf{X}>0\}})\big), \quad \text{and} \quad \frac{\partial^2 \mathrm{ReLU}(\mathbf{X})}{\partial \mathbf{X}^2} = \mathbf{0}. 
\]
\end{lemma}
\begin{proof}
See Appendix \ref{app:additional_properties}.
\end{proof}

Using this result, we determine the exact first-order derivatives ($\mathbf{B}_i$ terms) needed for the functional Hessian in Eq.~(\ref{eq:gauss_decomposition}).

\begin{proposition}[FFN Jacobians] \label{prop:ffn_jacobians}
Let $\mathbf{V} = \sigma(\mathbf{Y}\mathbf{W}_1)\mathbf{W}_2$. The Jacobians with respect to the weights are:
\begin{align}
    \frac{\partial \mathbf{V}}{\partial \mathbf{W}_2} &= \sigma(\mathbf{Y}\mathbf{W}_1) \otimes \mathbf{I}_{d_V}, \label{eq:ffn_jac_w2} \\
    \frac{\partial \mathbf{V}}{\partial \mathbf{W}_1} &= (\mathbf{I}_L \otimes \mathbf{W}_2^\top)\, \mathbf{D}_\sigma\, (\mathbf{Y} \otimes \mathbf{I}_{d_{ff}}), \label{eq:ffn_jac_w1}
\end{align}
where $\mathbf{D}_\sigma = \mathrm{diag}(\mathrm{vec}_r(\mathbf{1}_{\{\mathbf{Y}\mathbf{W}_1 > 0\}}))$.
\end{proposition}

\begin{proof}
We apply the vectorization properties from Appendix~\ref{app:properties}. For $\mathbf{W}_2$, the function is of the linear form $\mathbf{A} \mathbf{W}_2 \mathbf{B}$ with $\mathbf{A} = \sigma(\mathbf{Y}\mathbf{W}_1)$ and constant $\mathbf{B} = \mathbf{I}_{d_V}$. A direct application of the Matrix-Product Derivative (Property~\ref{prop:matrix_product_derivative}) yields first result. For $\mathbf{W}_1$, we apply the chain rule through the activation function. The gradient of the output with respect to the post-activation matrix is $\mathbf{I}_L \otimes \mathbf{W}_2^\top$ by Property~\ref{prop:matrix_product_derivative}. The gradient of the activation is $\mathbf{D}_\sigma$ by Lemma~\ref{lemma:relu_derivative_hessian}. Finally, the gradient of the pre-activation $ \mathbf{Y}\mathbf{W}_1\mathbf{I}_{d_{ff}} $ with respect to the weights is $\mathbf{Y} \otimes \mathbf{I}_{d_{ff}}$. Multiplying these three Jacobian matrices yields the second result.
\end{proof}

% \begin{figure}[t]
%     \centering
%     % LEFT PANEL
%     \begin{minipage}[t]{0.48\textwidth}
%         \centering
%         \includegraphics[width=\linewidth]{figs/hessian_entries.pdf}
%         \caption{Hessian entries for an \textbf{initialized model}. Note the magnitude heterogeneity; the \textcolor{olive}{Values} corresponding blocks have larger values.}
%         \label{fig:hessian_entries}
%     \end{minipage}
%     \hfill
%     % RIGHT PANEL
%     \begin{minipage}[t]{0.48\textwidth}
%         \centering
%         \includegraphics[width=\linewidth]{figs/hessian_entries_trained.pdf}
%         \caption{Hessian entries after \textbf{training} for several epochs. While magnitudes increase, the \textcolor{olive}{Values-Values} blocks remain the high-curvature dominant terms.}
%         \label{fig:hessian_entries_trained}
%     \end{minipage}
%     \label{fig:hessian_viz}
% \end{figure}

\begin{figure}[t]
    \centering
    \includegraphics[width=0.75\textwidth]{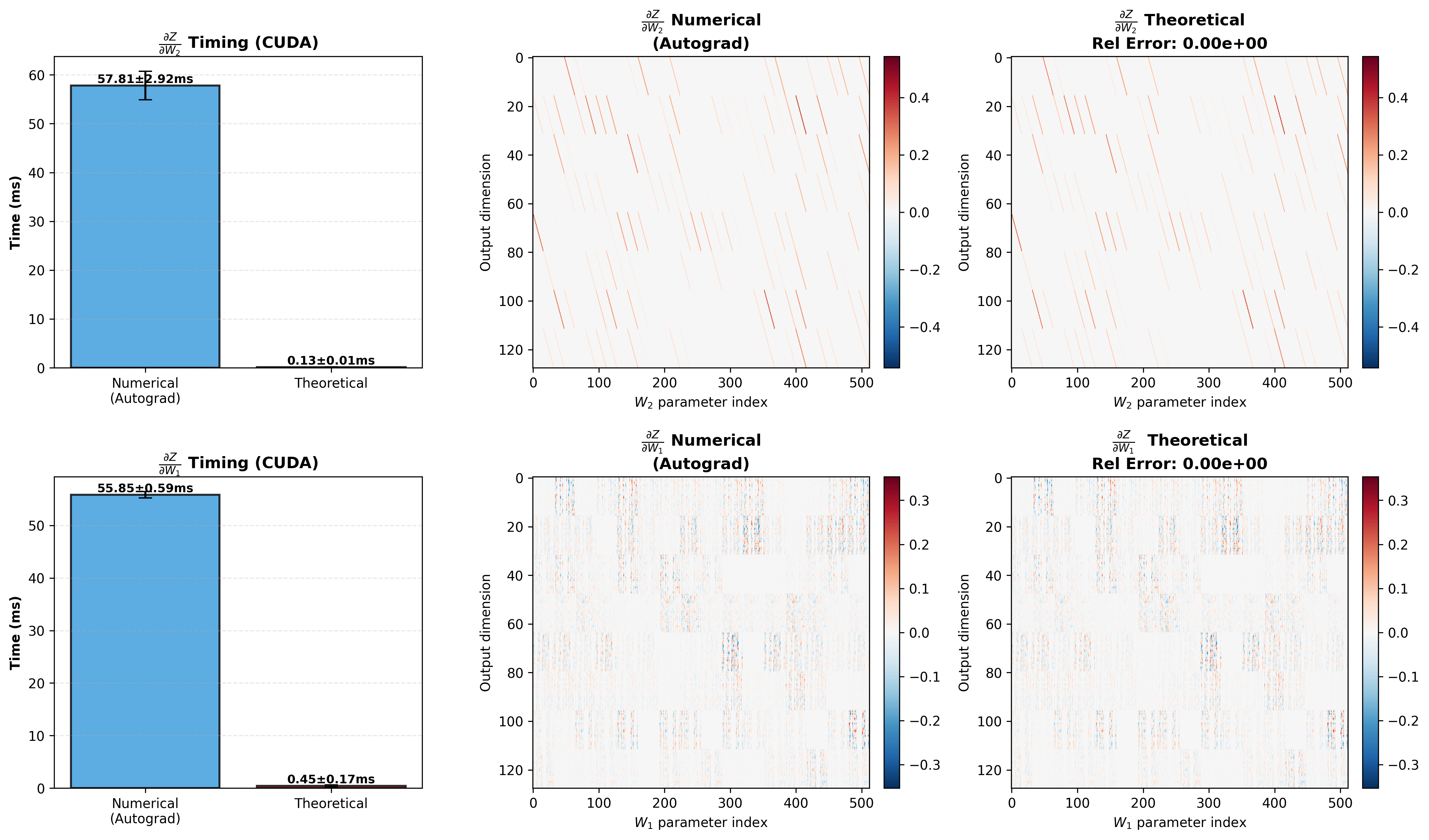}
    \vspace{-0.3em}
    \caption{FFN Jacobian benchmarks. \textbf{Left:} wall‑clock time – the theoretical formulas achieve a $444\times$ speed‑up for $\partial \textbf{Z}/\partial \textbf{W}_2$ (top) and $124\times$ for $\partial \textbf{Z}/\partial \textbf{W}_1$ (bottom). \textbf{Center/Right:} heat‑maps of the numerical (autograd) and theoretical Jacobians for a small configuration, demonstrating pixel‑wise identity.}
    \label{fig:ffn}
\end{figure}

\subsection{Full Transformer Block} \label{subsec:hessian_full}

Having characterized the differential properties of the LayerNorm operator (Section~\ref{subsec:layer_norm}) and the Feed-Forward Network (Section~\ref{subsec:ffn}), we now integrate these components with the Self-Attention mechanism to derive the full Transformer block Hessian, explicitly accounting for the Residual connections. Let the parameter set be $\mathbf{w} = \{\mathbf{W}_1, \mathbf{W}_2, \mathbf{W}_Q, \mathbf{W}_K, \mathbf{W}_V\}$. The block output is the composite signal defined in Eq. (\ref{eq:transformer}):
\[
\mathbf{Z} = \mathrm{LayerNorm}(\mathbf{S}), \quad \text{where} \quad \mathbf{S} = \mathbf{Y} + \mathrm{FFN}(\mathbf{Y}) \quad \text{and} \quad \mathbf{Y} = \mathrm{LayerNorm}(\mathbf{X} + \mathrm{SA}(\mathbf{X})).
\]

We first establish the first-order Jacobians, which serve as the linear operators for signal propagation.

\begin{theorem}[Transformer Block Jacobians] \label{thm:transformer_derivative}
For any weight matrix $\mathbf{W}_i \in \mathbf{w}$, the Jacobian of the output $\mathbf{Z}$ is given by the chain rule through the final LayerNorm:
\[
\frac{\partial \mathbf{Z}}{\partial \mathbf{W}_i} = \mathbf{J}_{Z} \mathbf{B}_i,
\]
where $\mathbf{J}_{Z} := \frac{\partial \mathrm{LayerNorm}(\mathbf{S})}{\partial \mathbf{S}}$ is the operator from Theorem~\ref{thm:layernorm_derivative}, and $\mathbf{B}_i := \frac{\partial \mathbf{S}}{\partial \mathbf{W}_i}$ represents the gradient flow into the final residual stream. Specifically:

1. \textbf{Feed-Forward Weights ($i \in \{1, 2\}$):} $\mathbf{B}_i$ matches the FFN Jacobians derived in Proposition \ref{prop:ffn_jacobians}:
   \begin{align*}
       \mathbf{B}_1 &= (\mathbf{I}_L \otimes \mathbf{W}_2^\top)\, \mathbf{D}_\sigma\, (\mathbf{Y} \otimes \mathbf{I}_{d_{ff}}), \\
       \mathbf{B}_2 &= \sigma(\mathbf{Y}\mathbf{W}_1) \otimes \mathbf{I}_{d_V}.
   \end{align*}

2. \textbf{Attention Weights ($i \in \{K, Q, V\}$):} The gradient flows back through the FFN residual and the first LayerNorm:
   \[
   \mathbf{B}_i = \mathbf{J}_{SY} \mathbf{J}_{Y} \mathbf{G}_i.
   \]
   Here, $\mathbf{J}_{SY} = \frac{\partial \mathbf{S}}{\partial \mathbf{Y}}$ explicitly includes the Residual Connection via the identity term:
   \[
   \mathbf{J}_{SY} = (\mathbf{I}_L \otimes \mathbf{W}_2^\top)\mathbf{D}_\sigma (\mathbf{I}_L \otimes \mathbf{W}_1^\top) + (\mathbf{I}_L \otimes \mathbf{I}_{d_V}).
   \]
   $\mathbf{J}_{Y}$ is the first LayerNorm Jacobian, and $\mathbf{G}_i$ are the Self-Attention Jacobians from \citep{noci2022signalpropagationtransformerstheoretical}.
\end{theorem}
\begin{proof}
See Appendix \ref{app:proof_transformer_derivative}. The derivation is a direct application of the chain rule using the matrix calculus properties established in Appendix \ref{app:properties}.
\end{proof}

We now state the explicit characterization of the Transformer Hessian blocks. This theorem connects the Functional Term from the Gauss-Newton decomposition (Eq. \ref{eq:gauss_decomposition}) to the specific architectural components.

\begin{theorem}[Transformer Block Hessian] \label{thm:transformer_hessian}
The Hessian block of the output $\mathbf{Z}$ with respect to parameter pair $(\mathbf{W}_i, \mathbf{W}_j)$ is:
\begin{equation} \label{eq:block_hessian_transformer}
\;\mathbf{H}_{\mathrm{tr}}^{(i,j)} := \frac{\partial^2 \mathbf{Z}}{\partial \mathbf{W}_i \partial \mathbf{W}_j}
= 
\underbrace{\left( \mathbf{J}_Z \otimes \mathbf{I}_{n_i} \right) \boldsymbol{\xi}_{ij}}_{\text{Inter-layer Interaction}}
  + 
\underbrace{\left( \mathbf{I}_{L d_V} \otimes \mathbf{B}_i^\top \right) \mathbf{H}_Z \mathbf{B}_j}_{\text{Normalization Curvature}}
\; 
\end{equation}
where $\mathbf{H}_Z$ is the LayerNorm Hessian (Theorem \ref{thm:layernorm_second_derivative}) and $\boldsymbol{\xi}_{ij} := \frac{\partial \mathbf{B}_i}{\partial \mathbf{W}_j}$ captures the second-order interactions between sub-layers. The interaction terms $\boldsymbol{\xi}_{ij}$ are:

% \begin{enumerate}
1. \textbf{FFN Curvature:} Due to the piecewise-linearity of ReLU (Lemma \ref{lemma:relu_derivative_hessian}), the second derivatives $\xi_{11}$ and $\xi_{22}$ vanish almost everywhere. Nevertheless, the mixed FFN interaction $\boldsymbol{\xi}_{12}$ and $\boldsymbol{\xi}_{21}$ remains non-zero, reflecting the bilinear coupling
between the two FFN weight matrices. (detailed in Appendix \ref{app:proof_transformer_hessian_estimate}).
    
2. \textbf{Attention-FFN Interaction:} Cross-terms $\boldsymbol{\xi}_{1k}$ capture how attention weights scale the gradients flowing into the FFN via $\mathbf{D}_\sigma$, inheriting the sparsity pattern.
    
3. \textbf{Attention Curvature:} Terms $\boldsymbol{\xi}_{k\ell}$ incorporate the second derivatives of the softmax function $\boldsymbol{\Phi}_{k\ell}$ derived in \citep{ormaniec2024attentionhessian}, modulated by the residual path $\mathbf{J}_{SY}$.
% \end{enumerate}
\end{theorem}

\textbf{Connection to the Loss Landscape.} 
Theorem \ref{thm:transformer_hessian} resolves the cryptic functional term $\frac{\partial^2 \mathbf{F}}{\partial \mathbf{W}^2}$ in the general Gauss-Newton decomposition. For a general twice-differentiable loss function $\mathcal{L}$, the full Hessian of the objective $\mathcal{J}(\mathbf{w}) = \mathcal{L}(\mathbf{Z}(\mathbf{w}))$ is:
\begin{equation} \label{eq:transformer_decomposition}
\frac{\partial^2 \mathcal{J}}{\partial \mathbf{W}_i \partial \mathbf{W}_j} 
= 
\underbrace{(\mathbf{J}_Z \mathbf{B}_i)^\top \frac{\partial^2 \mathcal{L}}{\partial \mathbf{Z}^2} (\mathbf{J}_Z \mathbf{B}_j)}_{\text{Loss Convexity}} 
+ 
\underbrace{\left( \frac{\partial \mathcal{L}}{\partial \mathbf{Z}} \otimes \mathbf{I}_{p_i q_i} \right) \mathbf{H}_{\text{tr}}^{(i,j)}}_{\text{Architectural Curvature}}.
\end{equation}
This decomposition reveals that even for a convex loss, where the first term is positively semi-defined, the optimization landscape might become non-convex due to $\mathbf{H}_{\text{tr}}^{(i,j)}$.

Finally, we provide a computable upper bound on the curvature, essential for determining step sizes and convergence rates.

\begin{theorem}[Spectral Bounds] \label{thm:transformer_hessian_estimate}
Let $\mathbf{H}_{\mathrm{tr}}$ be the full Hessian matrix for the block, composed of $m_b \times n_b = 25$ blocks. Its spectral norm is bounded by:
\begin{equation} \label{eq:full_bound_transformer}
\|\mathbf{H}_{\mathrm{tr}}\|_2
\;\le\;
5 \cdot \max_{i,j} \left(
    \|\frac{\partial^2 \mathcal{L}}{\partial \mathbf{Z}^2}\|_2 \|\mathbf{J}_Z \mathbf{B}_i\|_2 \|\mathbf{J}_Z \mathbf{B}_j\|_2 
    + 
    \big\|\frac{\partial \mathcal{L}}{\partial \mathbf{Z}}\big\|_2 \left( \|\mathbf{J}_Z\|_2 \|\boldsymbol{\xi}_{ij}\|_2 + \|\mathbf{B}_i\|_2 \|\mathbf{H}_Z\|_2 \|\mathbf{B}_j\|_2 \right)
\right).
\end{equation}
For Mean Squared Error, $\|\frac{\partial^2 \mathcal{L}}{\partial \mathbf{Z}^2}\|_2 = \frac{2}{L d_V}$. The terms $\|\boldsymbol{\xi}_{ij}\|_2$ and $\|\mathbf{H}_Z\|_2$ are explicitly bounded by polynomials of the input norm $\|\mathbf{X}\|_2$ and weight norms in Appendix \ref{app:proof_transformer_hessian_estimate}.
\end{theorem}
% At initialization (Figure~\ref{fig:hessian_entries}), the spectrum is highly anisotropic, with Value interactions clearly distinguishable from the background noise of FFN curvature. After training the model for several iterations until convergence (Figure~\ref{fig:hessian_entries_trained}), we observe that while the overall scale of the Hessian increases, the relative structural dominance of the $\mathbf{W}_V$-$\mathbf{W}_V$ block persists. This confirms that the specific polynomial scaling factors identified in Theorem~\ref{thm:self_attention_hessian_estimation} correctly predict the principal sources of curvature in the Transformer landscape. Additional layer-wise Hessians visualizations are present in Appendix \ref{app:exps}.

\section{Experimental Validation of Theoretical Formulas}\label{sec:experiments}

Since the main contribution of this work is theoretical, our experiments are designed primarily as numerical sanity checks of the derived formulas. We benchmark the closed-form Jacobians and Hessians against PyTorch automatic differentiation (autograd), and additionally report the evaluation speed benefits that arise from using explicit matrix formulas. All experiments use FP32 precision on a single NVIDIA A100 GPU.
% \subsection{Exactness against autograd}
\paragraph{Exactness against autograd.}
For each architectural component we compute the relative error
\[
\varepsilon_{\text{rel}} = \frac{\|\mathbf{J}_{\text{theory}} - \mathbf{J}_{\text{autograd}}\|_F}{\|\mathbf{J}_{\text{autograd}}\|_F + \epsilon},
\]
with $\epsilon$ a tiny constant to avoid division by zero. The results are summarised in Table~\ref{tab:errors}.

\begin{table}[t]
    \centering
    \small
    \caption{Empirical validation: theoretical formulas vs.\ autograd (FP32 precision). Every quantity achieves machine‑level agreement.}
    \label{tab:errors}
    \begin{tabular}{lc}
    \toprule
    \textbf{Component} & \textbf{Rel.\ Error} \\
    \midrule
    FFN Jacobian $\partial Z/\partial W_1, W_2$ (Prop.\ \ref{prop:ffn_jacobians}) & $< 10^{-15}$ \\
    LayerNorm Jacobian (Theorem \ref{thm:layernorm_derivative}) & $8.3 \times 10^{-8}$ \\
    LayerNorm Hessian (Theorem \ref{thm:layernorm_second_derivative}) & $1.5 \times 10^{-7}$ \\
    Transformer Jacobian (Theorem \ref{thm:transformer_derivative}) & $1.5 \times 10^{-7}$ \\
    Transformer Hessian (Theorem \ref{thm:transformer_hessian})\textsuperscript{$\dagger$} & $1.6 \times 10^{-7}$ \\
    \bottomrule
    \end{tabular}
    \\[0.3em]
    {\footnotesize \textsuperscript{$\dagger$}The functional term of the Gauss‑Newton decomposition is computed using autograd, because the required soft‑max second derivatives were calculated in the prior work \cite{ormaniec2024attentionhessian}. All FFN and LayerNorm contributions are evaluated via our purely theoretical formulas.}
\end{table}

The LayerNorm operators are the most sensitive to floating‑point rounding because they involve division by row‑wise standard deviations; nevertheless the relative error remains below $2\!\times\!10^{-7}$. The FFN Jacobians are algebraically exact and the error drops to the numerical noise floor. The full‑block Jacobian and Hessian inherit these properties, confirming that the chain‑rule assembly is correct and no approximation has been introduced.
% \subsection{Computational speed‑up}
\paragraph{Computational speed‑up.}
Beyond their theoretical value, the closed-form expressions also provide a practical evaluation advantage for componentwise derivative computations.  
We therefore time the computation of individual Jacobians and Hessians on an NVIDIA A100 GPU (FP32) using small synthetic configurations to isolate the component costs: $L=8$, $d_V=16$, $d_{ff}=32$ for the FFN ($\partial \textbf{Z}/\partial \textbf{W}_2$ and $\partial \textbf{Z}/\partial \textbf{W}_1$);  
$L=8$, $d_V=16$ for LayerNorm; and $L=6$, $d_V=16$, $d_{ff}=32$ for the full Transformer Jacobian.   Timing results are shown in Figure~\ref{fig:ln} (LayerNorm) and Figure~\ref{fig:ffn} (Feed‑Forward Network).

% \begin{figure}[t]
%     \centering
%     \includegraphics[width=0.75\textwidth]{figs/exp2_complete_cuda_a100.png}
%     \vspace{-0.3em}
%     \caption{FFN Jacobian benchmarks. \textbf{Left:} wall‑clock time – the theoretical formulas achieve a $444\times$ speed‑up for $\partial \textbf{Z}/\partial \textbf{W}_2$ (top) and $124\times$ for $\partial \textbf{Z}/\partial \textbf{W}_1$ (bottom). \textbf{Center/Right:} heat‑maps of the numerical (autograd) and theoretical Jacobians for a small configuration, demonstrating pixel‑wise identity.}
%     \label{fig:ffn}
% \end{figure}

% \begin{figure}[t]
%     \centering
%     \includegraphics[width=0.7\textwidth]{figs/exp3_complete_cuda_a100.png}
%     \vspace{-0.3em}
%     \caption{LayerNorm Jacobian benchmarks. \textbf{Left:} $82\times$ speed‑up. \textbf{Center/Right:} the theoretical formula (Theorem~\ref{thm:layernorm_derivative}) reproduces the autograd Jacobian to a relative error of $8.3\times10^{-8}$.}
%     \label{fig:ln}
% \end{figure}
Overall, these experiments confirm that the derived formulas agree with automatic differentiation up to numerical precision. They also show that explicit expressions can substantially reduce the cost of selected componentwise derivative evaluations, supporting their use as practical tools for Transformer curvature analysis.

For additional experiments, including the analysis of the Hessian evolution through training, see Appendix \ref{app:exps}.

\section{Discussion and Conclusion}\label{sec:disc}

We presented an exact second-order characterization of a full post-norm Transformer block with respect to its attention and FFN parameters. Unlike attention-only analyses, our derivation incorporates residual connections, LayerNorm, and the nonlinear FFN into a single matrix-calculus framework, and applies to arbitrary twice-differentiable losses through the Gauss--Newton decomposition.

Our results show that the full Transformer Hessian is shaped by several distinct architectural curvature sources. Self-attention contributes highly heterogeneous blocks, with query/key interactions exhibiting the strongest input-norm dependence. LayerNorm introduces additional curvature through inverse row-wise standard deviations, making the Hessian sensitive to variance collapse. The FFN contributes mainly through activation-gated Jacobians and cross-layer interactions, since ReLU has zero second derivative almost everywhere. Residual connections appear explicitly as identity terms in the Jacobian chain, controlling how curvature propagates across sublayers.

The derived spectral bounds make these mechanisms explicit and provide a principled way to reason about Transformer conditioning and training stability. They show that the full-block curvature is not simply inherited from self-attention, but is modulated by normalization, activation gates, weight norms, and residual pathways.

Finally, our experiments validate the formulas against automatic differentiation up to numerical precision and demonstrate substantial speedups for componentwise Jacobian computations. 
\bibliographystyle{plain}
\bibliography{references}

%%%%%%%%%%%%%%%%%%%%%%%%%%%%%%%%%%%%%%%%%%%%%%%%%%%%%%%%%%%%

\newpage
\appendix
\section{Additional experiments}\label{app:exps}

We complement the formula-validation experiments from Section~\ref{sec:experiments} with a small curvature-visualization study. We use the same post-norm Transformer-block structure analyzed in the main text, instantiated inside a simple ViT-style \citep{dosovitskiy2020image} classifier for MNIST \citep{deng2012mnist}: images are patchified by a linear projection, passed through a single Transformer block, and the resulting token representations are averaged to produce classification logits. Here we expand the plots into separate parameter-block visualizations.

\begin{figure}[H]
    \centering
    % LEFT PANEL
    \begin{minipage}[H]{0.48\textwidth}
        \centering
        \includegraphics[width=\linewidth]{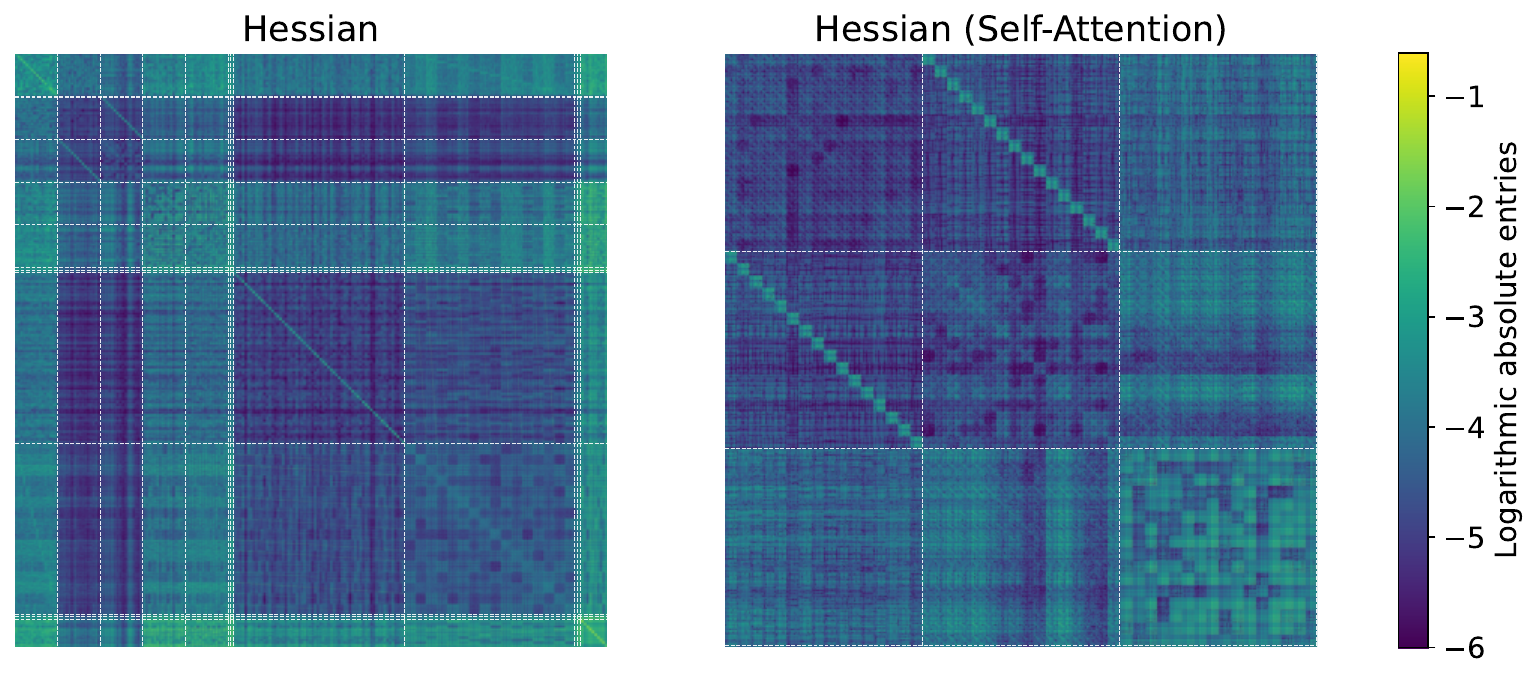}
        \caption{Hessian entries for an \textbf{initialized model}. Note the magnitude heterogeneity; the \textcolor{olive}{Values} corresponding blocks have larger values.}
        \label{fig:hessian_entries}
    \end{minipage}
    \hfill
    % RIGHT PANEL
    \begin{minipage}[H]{0.48\textwidth}
        \centering
        \includegraphics[width=\linewidth]{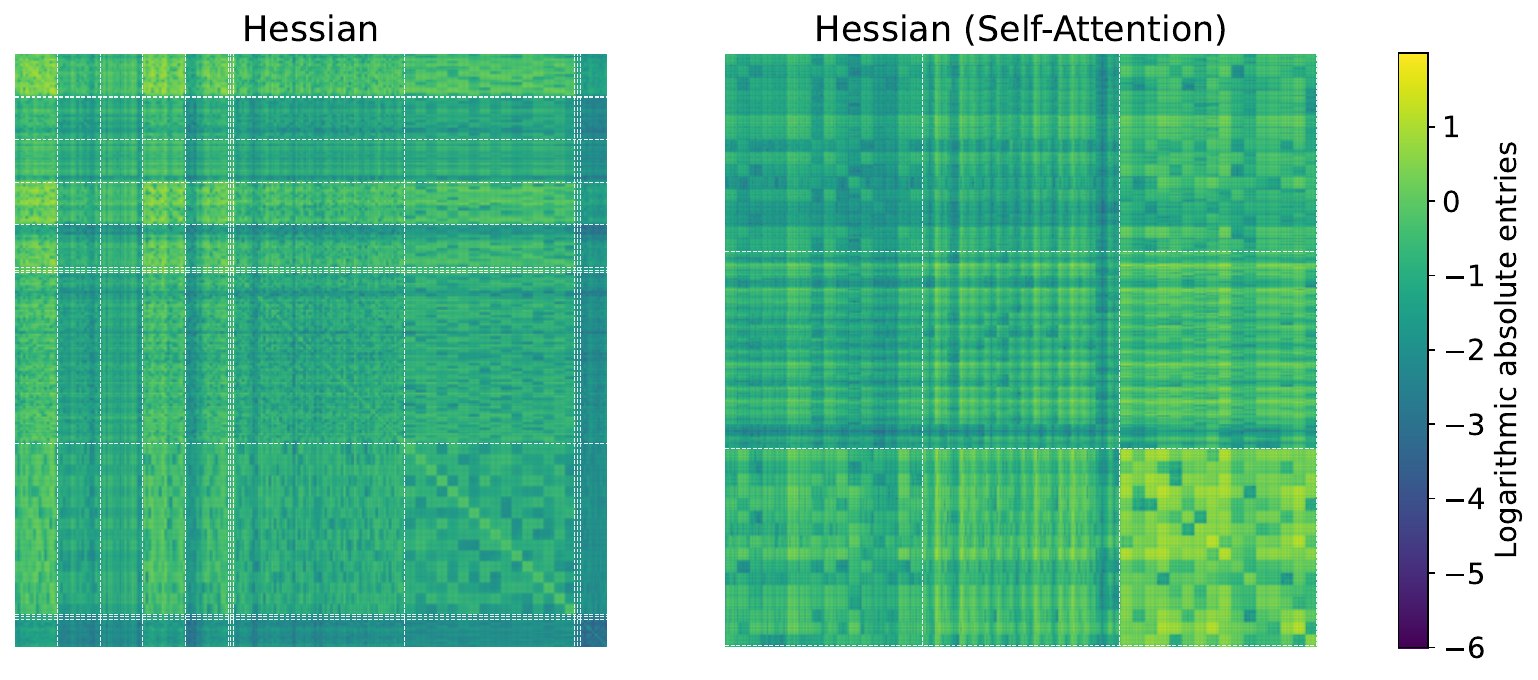}
        \caption{Hessian entries after \textbf{training} for several epochs. While magnitudes increase, the \textcolor{olive}{Values-Values} blocks remain the high-curvature dominant terms.}
        \label{fig:hessian_entries_trained}
    \end{minipage}
    \label{fig:hessian_viz}
\end{figure}

At initialization (Figure~\ref{fig:hessian_entries}), the spectrum is highly anisotropic, with Value interactions clearly distinguishable from the background noise of FFN curvature. After training the model for several iterations until convergence (Figure~\ref{fig:hessian_entries_trained}), we observe that while the overall scale of the Hessian increases, the relative structural dominance of the $\mathbf{W}_V$-$\mathbf{W}_V$ block persists. This confirms \cite{ormaniec2024attentionhessian} and that the specific polynomial scaling factors identified in Theorem~\ref{thm:self_attention_hessian_estimation} correctly predict the principal sources of curvature in the Transformer landscape. Additional layer-wise Hessians visualizations are present below.

\begin{figure}[H]
    \centering
    \includegraphics[width=\linewidth]{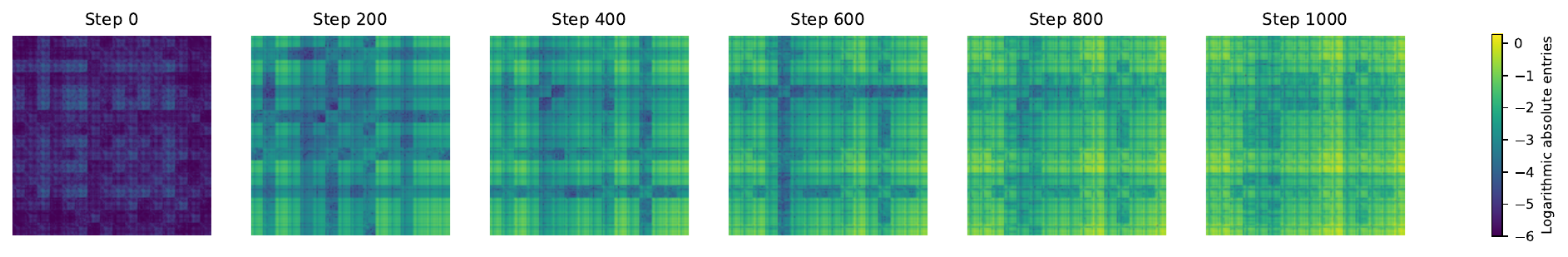}
    \caption{Queries entries visualization.}
    \label{fig:hessian_entries_queries}
\end{figure}

\begin{figure}[H]
    \centering
    \includegraphics[width=\linewidth]{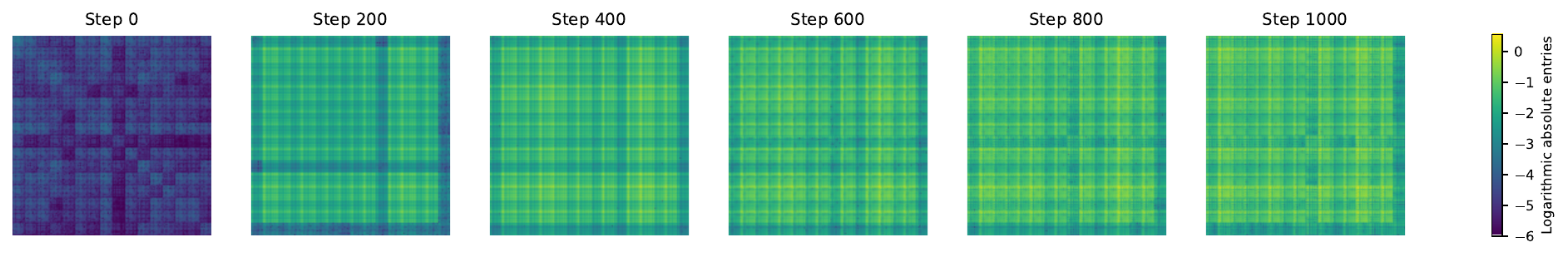}
    \caption{Keys entries visualization.}
    \label{fig:hessian_entries_keys}
\end{figure}

\begin{figure}[H]
    \centering
    \includegraphics[width=\linewidth]{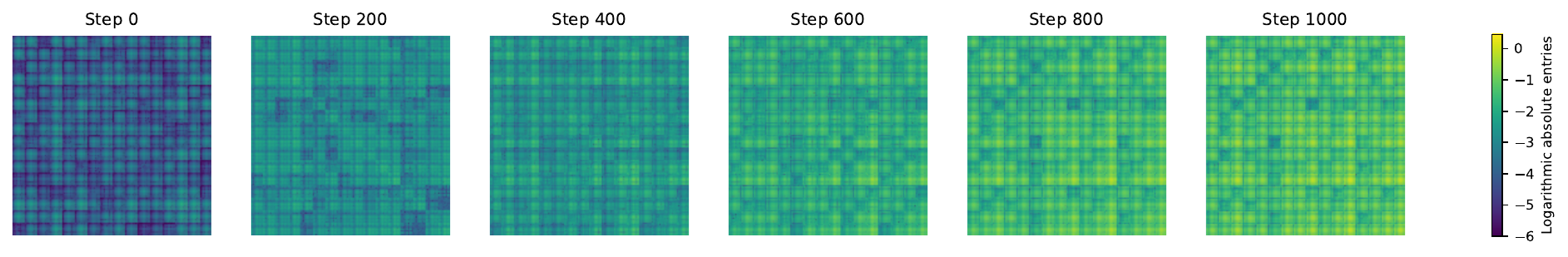}
    \caption{Values entries visualization.}
    \label{fig:hessian_entries_values}
\end{figure}

\begin{figure}[H]
    \centering
    \includegraphics[width=\linewidth]{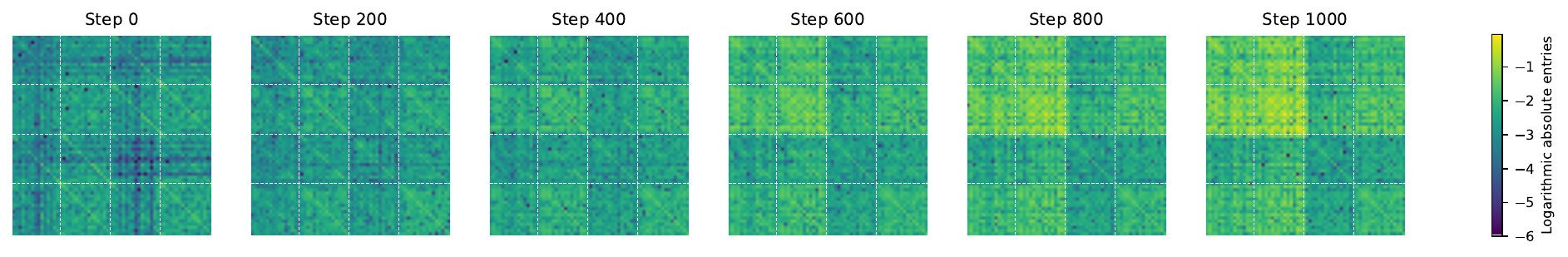}
    \caption{LayerNorm entries visualization.}
    \label{fig:hessian_entries_layernorm}
\end{figure}

\begin{figure}[H]
    \centering
    \includegraphics[width=\linewidth]{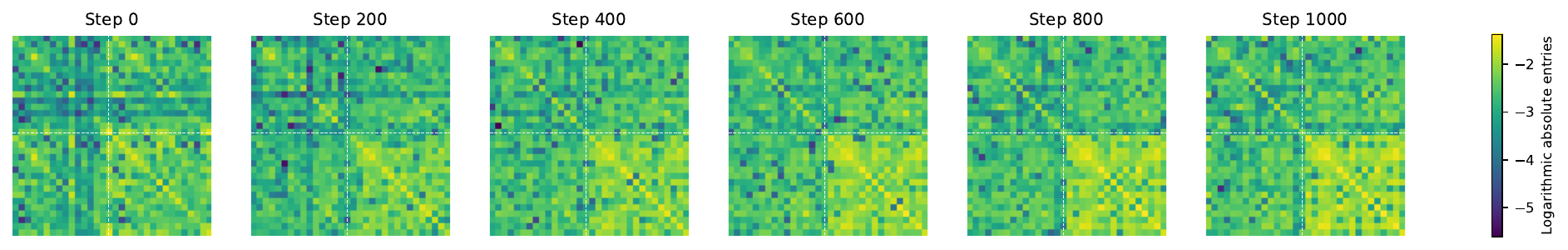}
    \caption{FeedForward entries visualization.}
    \label{fig:hessian_entries_feedforward}
\end{figure}

\section{Matrix calculus preliminaries} \label{app:properties}

\subsection{Basic matrix operations properties}

First, we define the notations and rules that we actively use in the text.

\begin{definition}[Matrix Norms] \label{def:matrix_norms}
For a matrix $\mathbf{A} \in \mathbb{R}^{m \times n}$:
\begin{align*}
\|\mathbf{A}\|_2 &= \sigma_1 \quad &&\text{(Spectral norm, largest singular value)} \\
\|\mathbf{A}\|_F &= \sqrt{\sum_{i=1}^m \sum_{j=1}^n |a_{ij}|^2} = \sqrt{\sum_{i=1}^r \sigma_i^2} \quad &&\text{(Frobenius norm)} \\
\|\mathbf{A}\|_1 &= \max_{1 \leq j \leq n} \sum_{i=1}^m |a_{ij}| \quad &&\text{(Maximum absolute column sum)} \\
\|\mathbf{A}\|_\infty &= \max_{1 \leq i \leq m} \sum_{j=1}^n |a_{ij}| \quad &&\text{(Maximum absolute row sum)} \\
\|\mathbf{A}\|_{\max} &= \max_{i,j} |a_{ij}| \quad &&\text{(Element-wise maximum, not a submultiplicative norm)}
\end{align*}
\end{definition}

\begin{definition}[Vectorization and Element-wise Operations] \label{def:vec_elem_ops}
Let $\mathbf{A}$ be a matrix and $\mathbf{v}$ be a vector.
\begin{itemize}
    \item $\mathrm{vec}_r(\mathbf{A})$ denotes the row-wise vectorization of matrix $\mathbf{A}$.
    \item $\mathbf{A}^{\circ \alpha}$ denotes the element-wise $\alpha$-power of matrix $\mathbf{A}$, i.e., $(\mathbf{A}^{\circ \alpha})_{ij} = (\mathbf{A}_{ij})^\alpha$.
    \item $\textit{diag}(\mathbf{v})$ creates a diagonal matrix with vector $\mathbf{v}$ on its main diagonal.
\end{itemize}
\end{definition}

\begin{property}[Relation between $\mathrm{vec}$ and $\mathrm{vec}_r$] \label{prop:vec_relation}
    Let $\mathbf{A} \in \mathbb{R}^{m \times n}$. The row-wise vectorization operator $\mathrm{vec}_r$ and the standard column-wise vectorization operator $\mathrm{vec}$ are related by the transpose:
\begin{equation*}
\mathrm{vec}_r(\mathbf{A}) = \mathrm{vec}(\mathbf{A}^\top)
\end{equation*}
\end{property}

\begin{definition}[Commutation Matrix] \label{def:commutation_matrix}

The commutation matrix $\mathbf{K}_{m,n} \in \mathbb{R}^{mn \times mn}$ is the unique matrix such that for any matrix $\mathbf{A} \in \mathbb{R}^{m \times n}$ the following holds

\begin{equation*}
\mathbf{K}_{m,n} \mathrm{vec}(\mathbf{A}) = \mathrm{vec}(\mathbf{A}^\top)
\end{equation*}

Using Property \ref{prop:vec_relation}, we immediately have the relationship:

\begin{equation*}
\mathrm{vec}_r(\mathbf{A}) = \mathbf{K}_{m,n} \mathrm{vec}(\mathbf{A}) \quad \text{and} \quad \mathrm{vec}(\mathbf{A}) = \mathbf{K}_{n,m} \mathrm{vec}_r(\mathbf{A})
\end{equation*}

since $\mathbf{K}_{n,m}\mathbf{K}_{m,n} = \mathbf{I}_{mn}$.
\end{definition}

From \citep{magnus1988matrix} we utilize the property

\begin{property} [Row-wise vectorization of matrix product] \label{prop:vec_r_matrix_product}
    Let $\mathbf{X}, \mathbf{A}, \mathbf{B}$ be matrices with appropriate dimensions, then
    \begin{equation*}
        \mathrm{vec}_r(\mathbf{A} \mathbf{X}\mathbf{B}) = (\mathbf{A} \otimes \mathbf{B}^\top) \mathrm{vec}_r(\mathbf{X})
    \end{equation*}
\end{property}

\begin{property}[Row-wise vectorization of Hadamard product] \label{prop:vec_r_hadamard_product}
Let $\mathbf{A}, \mathbf{B} \in \mathbb{R}^{m \times n}$. Then

\[
\mathrm{vec}_r(\mathbf{A} \circ \mathbf{B}) = \textit{diag}(\mathrm{vec}_r(\mathbf{A})) \mathrm{vec}_r(\mathbf{B})
\]

where $\circ$ denotes the Hadamard (element-wise) product.
This result follows directly from \citep{magnus1988matrix}, where the similar result was obtained for column-wise vectorization. 
\end{property}

\begin{proposition} [Identification Theorem for Row-wise Vectorization] \label{prop:identification_theorem_vec_r}

Let $\mathrm{SA}: \mathbb{R}^{m\times n} \rightarrow \mathbb{R}^{p, q}$ be a differentiable matrix-valued function of a matrix $\mathbf{X} \in \mathbb{R}^{m \times n}$. If the differential of $\mathrm{SA}$ can be written as \begin{equation*}
d \mathrm{vec}_r(\mathrm{SA}(\mathbf{X})) = \mathbf{J} \cdot d \mathrm{vec}_r(\mathbf{X})
\end{equation*}

for some matrix $\mathbf{J} \in \mathbb{R}^{pq \times mn}$ that does not depend on $d\mathbf{X}$. Then $\mathbf{J}$ is the Jacobian matrix of the transformation from $\mathbf{X}$ to $\mathrm{SA}(\mathbf{X})$ with respect to row-wise vectorization. We denote this as:

\begin{equation*}
\frac{\partial \mathrm{SA}(\mathbf{X})}{\partial \mathbf{X}} := \frac{\partial \mathrm{vec}_r(\mathrm{SA}(\mathbf{X}))}{\partial (\mathrm{vec}_r(\mathbf{X}))^\top} = \mathbf{J}
\end{equation*}

This is the $\mathrm{vec}_r$ analogue of the fundamental Identification Theorem from \citep{magnus1988matrix} for column-wise vectorization.
    
\end{proposition}

\begin{property}[Element-wise division] \label{prop:elem_wise_division}
    Let $\mathbf{A} \in \mathbb{R}^{m\times n}$ be a matrix and $\mathbf{b} \in \mathbb{R}^{m \times 1}$ be a vector. Then for matrix $\mathbf{C} \in {\in \mathbb{R}^{m \times n}}$, where $c_{i,j} = \frac{a_{i,j}}{b_i}$ is fulfilled that
    \begin{equation*}
        \mathbf{C} = \textit{diag}^{-1}(\mathbf{b})\mathbf{A}
    \end{equation*}
\end{property}

\begin{proposition} [Spectral norm of $\mathbf{1}_{L \times L}$ matrix] \label{prop:1_spectral_norm}
    Let $\mathbf{A} = \mathbf{1}_{L \times L}$ (a matrix full of 1). Then its spectral norm is
    \begin{equation*}
        \| \mathbf{A}\|_2 = L
    \end{equation*}
\end{proposition}

\begin{proof}
    Using basic Linear Algebra properties, we obtain $\mathrm{tr}(\mathbf{A}) = L$ and $\mathrm{rank}(\mathbf{A}) = 1 = \mathrm{dim}(\mathrm{Im}(\mathbf{X}))$. Therefore, using $\mathrm{dim}(\mathrm{Im}(\mathbf{X})) + \mathrm{dim}(\mathrm{Ker}(\mathbf{X})) = L$, we get $\mathrm{dim}(\mathrm{Ker}(\mathbf{X})) = L - 1$. Thus, for $i \in \{2, \dots L\}$ we get $\lambda_i = 0$ and for $\lambda_1 = L$. Then, the only non-null singular value of the matrix $\mathbf{A}$ is $\sqrt{L^2} = L$. Thus, we obtain that $\| \mathbf{A}\|_2 = L$, according to Definition \ref{def:matrix_norms}.
\end{proof}

\subsection{Matrix-valued functions derivative properties}

Next, we introduce the properties for calculating the matrix-valued function derivative.

\begin{property} [Matrix-Product derivative] \label{prop:matrix_product_derivative}
    Let $\mathbf{X}, \mathbf{A}, \mathbf{B}$ be matrices with appropriate dimensions, then
    \begin{equation*}
        \frac{\partial\mathbf{A}\mathbf{X}\mathbf{B}}{\partial\mathbf{X}} = \mathbf{A} \otimes \mathbf{B}^\top
    \end{equation*}
    where $\mathbf{A}$ and $\mathbf{B}$ have no dependence on $\mathbf{X}$.
\end{property}

Detailed proof of this statement can be found in \citep{singh2021analyticinsightsstructurerank}.

\begin{property} [Kronecker-Product derivative] \label{prop:kronecker_product_derivative}
    Let \( \mathbf{X} \in \mathbb{R}^{n \times q} \) and \( \mathbf{Y} \in \mathbb{R}^{p \times r} \). Then
\[
\frac{\partial (\mathbf{X} \otimes \mathbf{Y})}{\partial \mathbf{X}} = (\mathbf{I}_n \otimes \mathbf{K}_{p,q} \otimes \mathbf{I}_r) \left( \mathbf{I}_{nq} \otimes \mathrm{vec}_r \mathbf{Y} \right),
\]
and analogously
\[
\frac{\partial (\mathbf{X} \otimes \mathbf{Y})}{\partial \mathbf{Y}} = (\mathbf{I}_n \otimes \mathbf{K}_{p,q} \otimes \mathbf{I}_r) \left( \mathrm{vec}_r \mathbf{X} \otimes \mathbf{I}_{pr} \right).
\]
\end{property}

The detailed proof is in \citep{ormaniec2024attentionhessian}.

From the properties above, we derive calculations for special cases which we use in this paper.

\begin{proposition} [Matrix-valued functions multiplication derivative] \label{prop:matrix_funcs_product_derivative}

Let  $\mathbf{A}(\mathbf{X}) \in \mathbb{R}^{p \times r}$ and $\mathbf{B}(\mathbf{X}) \in \mathbb{R}^{r \times q}$ be matrix-valued functions of the matrix $\mathbf{X}$, then

\begin{equation*}
    \frac{\partial\mathbf{A}(\mathbf{X}) \mathbf{B}(\mathbf{X})}{\partial \mathbf{X}} = \left(\mathbf{A} \otimes \mathbf{I}_q 
 \right) \frac{\partial\mathbf{B}}{\partial\mathbf{X}} + \left( \mathbf{I}_p \otimes \mathbf{B}^\top\right) \frac{\partial\mathbf{A}}{\partial\mathbf{X}}
\end{equation*}
    
\end{proposition}

\begin{proof}
    First, we apply  a classic chain-rule for calculation a derivative of a complicated function and then combine it with Property \ref{prop:matrix_product_derivative}
    \begin{align*}
        \frac{\partial\mathbf{A}(\mathbf{X}) \mathbf{B}(\mathbf{X})}{\partial \mathbf{X}} &= \frac{\partial \mathbf{A}\mathbf{B}}{\partial\mathbf{B}}\frac{\partial\mathbf{B}}{\partial \mathbf{X}} + \frac{\partial \mathbf{A}\mathbf{B}}{\partial\mathbf{A}}\frac{\partial\mathbf{A}}{\partial \mathbf{X}} = \frac{\partial \mathbf{A}\mathbf{B}\mathbf{I}_q}{\partial\mathbf{B}}\frac{\partial\mathbf{B}}{\partial \mathbf{X}} + \frac{\partial \mathbf{I}_p\mathbf{A}\mathbf{B}}{\partial\mathbf{A}}\frac{\partial\mathbf{A}}{\partial \mathbf{X}} = \\
        &= \left( \mathbf{A} \otimes \mathbf{I}_q \right) \frac{\partial \mathbf{B}}{\partial \mathbf{X}} + \left( \mathbf{I}_p \otimes \mathbf{B}^\top \right) \frac{\partial \mathbf{A}}{\partial \mathbf{X}}
    \end{align*}
\end{proof}

\begin{proposition} [Matrix-valued functions Kronecker product derivative] \label{prop:matrix_funcs_kronecker_product_derivative}

Let  $\mathbf{A}(\mathbf{X}) \in \mathbb{R}^{n \times q}$ and $\mathbf{B}(\mathbf{X}) \in \mathbb{R}^{p \times r}$ be matrix-valued functions of the matrix $\mathbf{X}$, then

\begin{equation*}
    \frac{\partial\mathbf{A}(\mathbf{X}) \otimes \mathbf{B}(\mathbf{X})}{\partial \mathbf{X}} = (\mathbf{I}_n \otimes \mathbf{K}_{p,q} \otimes \mathbf{I}_r) \left( \left( \mathrm{vec}_r \mathbf{A} \otimes \mathbf{I}_{pr} \right) \frac{\partial\mathbf{B}}{\partial \mathbf{X}} + \left( \mathbf{I}_{nq} \otimes \mathrm{vec}_r \mathbf{B} \right) \frac{\partial\mathbf{A}}{\partial \mathbf{X}}\right)
\end{equation*}
    
\end{proposition}

\begin{proof}
    First, we apply a classic chain rule for calculating the derivative of a complicated function and then combine it with Property \ref{prop:kronecker_product_derivative}
    \begin{align*}
        \frac{\partial\mathbf{A}(\mathbf{X}) \otimes \mathbf{B}(\mathbf{X})}{\partial \mathbf{X}} &= \frac{\partial \mathbf{A}\otimes\mathbf{B}}{\partial\mathbf{B}}\frac{\partial\mathbf{B}}{\partial \mathbf{X}} + \frac{\partial \mathbf{A}\otimes\mathbf{B}}{\partial\mathbf{A}}\frac{\partial\mathbf{A}}{\partial \mathbf{X}} = \\
        &= (\mathbf{I}_n \otimes \mathbf{K}_{p,q} \otimes \mathbf{I}_r) \left( \mathrm{vec}_r \mathbf{A} \otimes \mathbf{I}_{pr} \right) \frac{\partial\mathbf{B}}{\partial \mathbf{X}} + (\mathbf{I}_n \otimes \mathbf{K}_{p,q} \otimes \mathbf{I}_r) \left( \mathbf{I}_{nq} \otimes \mathrm{vec}_r \mathbf{B} \right) \frac{\partial\mathbf{A}}{\partial \mathbf{X}} = \\
        &= (\mathbf{I}_n \otimes \mathbf{K}_{p,q} \otimes \mathbf{I}_r) \left( \left( \mathrm{vec}_r \mathbf{A} \otimes \mathbf{I}_{pr} \right) \frac{\partial\mathbf{B}}{\partial \mathbf{X}} + \left( \mathbf{I}_{nq} \otimes \mathrm{vec}_r \mathbf{B} \right) \frac{\partial\mathbf{A}}{\partial \mathbf{X}}\right)
    \end{align*}
\end{proof}

Next, we develop the operations that we introduced above and derive calculations using $\mathrm{\text{vec}}_r$ notation as we do in this paper.

\begin{proposition} [Derivative of the invert matrix] \label{prop:invert_derivative}
For an invertible square matrix $\mathbf{D} \in \mathbb{R}^{n \times n}$, the derivative of its inverse is
\[
    \frac{\partial \mathbf{D}^{-1}}{\partial \mathbf{D}} 
    = -\mathbf{D}^{-1} \otimes \mathbf{D}^{-\top}.
\]
\end{proposition}

\begin{proof}
This is a standard result in matrix calculus. The differential identity 
\[
    d(\mathbf{D}^{-1}) = -\mathbf{D}^{-1} \,(d\mathbf{D})\, \mathbf{D}^{-1}
\]
appears in \citep{petersen2012matrix} and in \citep{magnus1988matrix}.  
Applying the $\mathrm{vec}_r$ operator and using the property \ref{prop:vec_r_matrix_product} yields 

\[
\mathrm{vec}_r(-\mathbf{D}^{-1} \,(d\mathbf{D})\, \mathbf{D}^{-1}) = (-\mathbf{D}^{-1} \otimes \mathbf{D}^{-\top}) \mathrm{vec}_r(d\mathbf{D})
\]

By the definition and the identification theorem from Property \ref{prop:identification_theorem_vec_r} we obtain

\[
\mathrm{vec}_r(d\mathbf{D}^{-1}) = \frac{\partial \mathrm{vec}_r \mathbf{D}^{-1}}{\partial\mathrm{vec}_r\mathbf{D}} \mathrm{vec}_r(d\mathbf{D})
\]

Comparing two results we get $\frac{\partial \mathrm{vec}_r \mathbf{D}^{-1}}{\partial\mathrm{vec}_r\mathbf{D}} = (-\mathbf{D}^{-1} \otimes \mathbf{D}^{-\top})$

\end{proof}

\begin{proposition}[Derivative of $\mathrm{diag}(\cdot)$] \label{prop:diag_derivative}
For $\mathbf{v} \in \mathbb{R}^{L \times 1}$, the derivative of the diagonalization map is
\[
    \frac{\partial \mathrm{diag}(\mathbf{v})}{\partial \mathbf{v}}
    = \big(\mathbf{e}_1 \otimes \mathbf{e}_1 \quad \dots \quad \mathbf{e}_L \otimes \mathbf{e}_L\big),
\]
where $\mathbf{e}_i$ are the standard basis vectors in $\mathbb{R}^L$.
\end{proposition}

\begin{proof}
By Definition \ref{def:vec_elem_ops}, $\mathrm{diag}(\mathbf{v})$ places entry $v_i$ at position $(i,i)$ of the resulting diagonal matrix.

The derivative of $\mathrm{diag}(\mathbf{v})$ w.r.t.\ $v_i$ is the elementary matrix $\mathbf{E}_{ii} = \mathbf{e}_i \mathbf{e}_i^\top$ that has one in position $(i,i)$ and zeros elsewhere.

Applying the row-wise vectorization operator, we obtain
\[
\mathrm{vec}_r(\mathbf{E}_{i,i}) = \mathbf{e}_i \otimes \mathbf{e}_i
\]

by the standard Kronecker–vec identity \ref{prop:vec_r_matrix_product}.

Stacking across $i=1,\dots,L$, the Jacobian becomes

\[
    \frac{\partial \mathrm{diag}(\mathbf{v})}{\partial \mathbf{v}}
    = \big(\mathbf{e}_1 \otimes \mathbf{e}_1 \quad \dots \quad \mathbf{e}_L \otimes \mathbf{e}_L\big),
\]
\end{proof}

\begin{proposition}[Derivative of the Hadamard square] \label{prop:hadamard_square_derivative}
For a matrix $\mathbf{A} \in \mathbb{R}^{m \times n}$, the derivative of the elementwise square is
\[
    \frac{\partial \mathbf{A}^{\circ 2}}{\partial \mathbf{A}}
    = 2 \cdot \mathrm{diag}\!\big(\mathrm{vec}_r(\mathbf{A})\big).
\]
\end{proposition}

\begin{proof}
By Definition \ref{def:vec_elem_ops}, $(\mathbf{A}^{\circ 2}){ij} = (\mathbf{A}{ij})^2$.
Differentiating elementwise gives $d(\mathbf{A}^{\circ 2}) = 2\mathbf{A} \circ d\mathbf{A}$.
Applying the $\mathrm{vec}_r$ operator and using Property~\ref{prop:vec_r_hadamard_product}, we obtain
\[
    \mathrm{vec}_r(d(\mathbf{A}^{\circ 2})) = 2 \textit{diag}(\mathrm{vec}_r(\mathbf{A})) \mathrm{vec}_r (d\mathbf{A})
\]

By the identification theorem from Property \ref{prop:identification_theorem_vec_r}, this implies

\[
    \frac{\partial \mathbf{A}^{\circ 2}}{\partial \mathbf{A}}
    = \frac{\partial \mathrm{vec}_r(\mathbf{A}^{\circ 2})}{\partial \mathrm{vec}_r(\mathbf{A})} = 2 \cdot \mathrm{diag}\!\big(\mathrm{vec}_r(\mathbf{A})\big)
\]

which establishes the result.
\end{proof}

\begin{proposition}[Derivative of the Hadamard root]\label{prop:hadamard_root_derivative}
For $\mathbf{A} \in \mathbb{R}^{m \times n}$ with positive entries, the derivative of the elementwise square root is
\[
    \frac{\partial \mathbf{A}^{\circ \frac{1}{2}}}{\partial \mathbf{A}}
    = \tfrac{1}{2}\, \mathrm{diag}^{-1}\!\big(\mathrm{vec}_r^{\circ \frac{1}{2}}(\mathbf{A})\big).
\]
\end{proposition}
\begin{proof}
Similarly to the proof of Proposition \ref{prop:hadamard_square_derivative}, we obtain $d(\mathbf{A}^{\circ 1/2}) = \frac{1}{2}\mathbf{A}^{\circ -1/2} \circ d\mathbf{A}$ 
Thus, writing in vectorized form gives
\[
    \frac{\partial \mathbf{A}^{\circ \frac{1}{2}}}{\partial \mathbf{A}}
    = \frac{\partial \mathrm{vec}_r(\mathbf{A}^{\circ \frac{1}{2}})}{\partial \mathrm{vec}_r(\mathbf{A})} = \tfrac{1}{2}\, \mathrm{diag}^{-1}\!\big(\mathrm{vec}_r^{\circ \frac{1}{2}}(\mathbf{A})\big).
\]
\end{proof}

\begin{proposition} [Transposed Matrix derivative] \label{prop:transposed_matrix_derivative}

Let $\mathbf{A} \in \mathbb{R}^{m \times n}$, then the following holds:

\begin{equation*}
    \frac{\partial \mathbf{A}^\top}{\partial \mathbf{A}} = \mathbf{K}_{n, m}
\end{equation*}
    
\end{proposition} 

\begin{proof}
    Combining a similar property from \citep{magnus1988matrix} for column-wise vectorization with the column-row connection rule \ref{prop:vec_relation} and \ref{def:commutation_matrix} we obtain the theorem statement.
\end{proof}

\subsection{Matrix norm properties}

Similarly to \citep{petersen2012matrix} we introduce a matrix norms table comparison.

\begin{property}[Matrix norm inequalities] \label{prop:matrix_norm_inequalities}
Let $\mathbf{A} \in \mathbb{R}^{m \times n}$. Then the following inequalities hold between different matrix norms:

\begin{center}
\renewcommand{\arraystretch}{1.5}
\begin{tabular}{c|ccccc}
\diagbox[height=2em, width=4em]{X}{Y} & $\|\mathbf{A}\|_{\max}$ & $\|\mathbf{A}\|_1$ & $\|\mathbf{A}\|_\infty$ & $\|\mathbf{A}\|_2$ & $\|\mathbf{A}\|_F$ \\
\hline
$\|\mathbf{A}\|_{\max}$ &  & 1 & 1 & 1 & 1 \\
$\|\mathbf{A}\|_1$ & $m$ &  & $m$ & $\sqrt{m}$ & $\sqrt{m}$ \\
$\|\mathbf{A}\|_\infty$ & $n$ & $n$ &  & $\sqrt{n}$ & $\sqrt{n}$ \\
$\|\mathbf{A}\|_2$ & $\sqrt{mn}$ & $\sqrt{n}$ & $\sqrt{m}$ &  & 1 \\
$\|\mathbf{A}\|_F$ & $\sqrt{mn}$ & $\sqrt{n}$ & $\sqrt{m}$ & $\sqrt{d}$ & \\
\end{tabular}
\end{center}

where $d = \operatorname{rank}(\mathbf{A})$. The table should be read as: for any two norms $\|\cdot\|_X$ and $\|\cdot\|_Y$,
\[
\|\mathbf{A}\|_X \leq c \cdot \|\mathbf{A}\|_Y
\]
where $c$ is the constant found at the intersection of row $X$ and column $Y$.
\end{property}

\begin{property} [Matrix sum norm] \label{prop:matrix_sum_norm}

Let $\mathbf{A}$ and $\mathbf{B}$ be matrices from $\mathbb{R}^{m \times n}$, then

\begin{equation}
    \| \mathbf{A} + \mathbf{B}\|_2 \leq \| \mathbf{A}\|_2 + \| \mathbf{B}\|_2
\end{equation}
    
\end{property}

\begin{property} [Kronecker product norm] \label{prop:kronecker_product_norm}

Let $\mathbf{A} \in \mathbb{R}^{m \times n}$ and $\mathbf{B} \in \mathbb{R}^{p \times q}$, then the following holds

\begin{equation*}
    \| \mathbf{A} \otimes \mathbf{B}\|_2 = \| \mathbf{A}\|_2 \| \mathbf{B}\|_2
\end{equation*}
    
\end{property}

\begin{property} [Matrix product norm] \label{prop:matrix_product_norm}

Let $\mathbf{A} \in \mathbb{R}^{m \times n}$ and $\mathbf{B} \in \mathbb{R}^{n \times q}$, then the following holds

\begin{equation*}
    \| \mathbf{A} \mathbf{B}\|_2 \leq \| \mathbf{A}\|_2 \| \mathbf{B}\|_2
\end{equation*}
    
\end{property}

The properties above can be found in \citep{magnus1988matrix}.

\begin{property} [Block-matrix norm inequality] \label{prop:block_matrix_norm}

Let $\mathbf{A} \in \mathbb{R}^{m \times n}$ be a block-matrix, each block of which is a matrix $\mathbf{B}_{i,j}$, thus the following holds

\begin{equation*}
    \| \mathbf{A} \|_2 \leq \sqrt{m n} \max\limits_{i,j}\| \mathbf{B}_{i,j}\|_2
\end{equation*}

Note, if matrix $\mathbf{A}$ is block-diagonal, then the strict equality holds $\| \mathbf{A} \|_2 = \max\limits_{i}\| \mathbf{B}_{i,i}\|_2$.
\end{property}

\begin{property} [Transposed matrix norm] \label{prop:transposed_matrix_norm}

Let $\mathbf{A} \in \mathbb{R}^{m \times n}$, then

\begin{equation*}
    \| \mathbf{A}\|_2 = \| \mathbf{A}^\top\|_2
\end{equation*}
    
\end{property}

\section{Proofs of the Theorems}\label{app:B}
\subsection{Proof of Theorem~\ref{thm:self_attention_hessian_estimation}}\label{app:proof_self_attention_hessian_estimation}

\begin{proof}

From Lemma A.3 \citep{noci2022signalpropagationtransformerstheoretical} and using Properties \ref{prop:matrix_product_norm} and \ref{prop:kronecker_product_norm}

\begin{align*}
    \| \frac{\partial \mathbf{A}}{\partial \mathbf{T}}\|_2 = \frac{1}{L} \| \mathbf{I}_L\|_2 \| \mathbf{I}_L - \frac{1}{L}\mathbf{1}_{L \times L}\|_2 \leq \frac{1}{L}
\end{align*}

Here we used that $\frac{1}{L}\mathbf{1}_{L \times L}$ is a projection matrix, therefore $\mathbf{I}_L - \frac{1}{L}\mathbf{1}_{L \times L}$ is a projection matrix and it's norm is $\| \mathbf{I}_L - \frac{1}{L}\mathbf{1}_{L \times L}\|_2 \leq 1$.

Next we estimate the $\mathbf{Z}_1$  norm, utilizing the same Properties \ref{prop:matrix_product_norm} and \ref{prop:kronecker_product_norm}

\begin{align*}
    \| \mathbf{Z}_1\|_2 \leq \| \mathbf{I}_L \otimes \mathbf{X}^{\top}\|_2 \| \frac{\partial \mathbf{A}}{\partial \mathbf{T}}\|_2 \| \mathbf{X} \otimes \mathbf{X}\|_2 \leq \| \mathbf{X}\|_2 \frac{1}{L} \|\mathbf{X}\|^2_2 = \frac{1}{L} \|\mathbf{X}\|^3_2
\end{align*}

where we used Property \ref{prop:transposed_matrix_norm} for $\| \mathbf{X}\|_2 = \| \mathbf{X}^\top\|_2$.

Now we calculate estimations for the outer-product Hessian part.

But before that we estimate $\| \mathbf{A}\|_2$. This block itself is a row-wise softmax matrix. Thus, each element $ \mathbf{A}_{i,j} \leq 1$. Next we use Property \ref{prop:matrix_norm_inequalities} and obtain $ \|\mathbf{A}\|_{\max} \leq \|\mathbf{A}\|_2 \leq \sqrt{LL} \|\mathbf{A}\|_{\max} = L\| \mathbf{A}\|_{max} \leq L$. Therefore, the $\|\mathbf{M}_1\|_2 = \|\mathbf{A}\mathbf{X}\|_2 \leq L \|\mathbf{X}\|_2$.

Thus, the $\|\mathbf{H}_o (\mathbf{W}_i, \mathbf{W}_j) \|_2$ is estimated below:
\begin{align*}
    \| \mathbf{H}_o(\mathbf{W}_V, \mathbf{W}_V)\|_2 \leq \frac{2}{L d_V} \| \mathbf{M}_1\|^2_2 1 \leq \frac{2}{L d_V}\| \mathbf{A}\|^2_2 \| \mathbf{X}\|^2_2 \leq \frac{2}{L d_V} L^2 \| \mathbf{X}\|^2_2 = \frac{2L}{d_V}\| \mathbf{X}\|^2_2
\end{align*}
\begin{align*}
    \| \mathbf{H}_o(\mathbf{W}_Q, \mathbf{W}_Q)\|_2 &\leq \| \frac{2}{L d_V d_K} (\mathbf{I}_{d_V} \otimes \mathbf{W}^\top_K) \mathbf{Z}^\top_1 (\mathbf{I}_{L} \otimes \mathbf{W}_V\mathbf{W}^\top_V)\ \mathbf{Z}_1 (\mathbf{I}_{d_V} \otimes \mathbf{W}_K)\|_2 \\
    &\leq \frac{2}{L d_V d_K} \| \mathbf{W}_K\|_2^2 \|\mathbf{Z}_1 \|^2_2 \| \mathbf{W}_V\|^2_2 \leq \frac{2}{L d_V d_K}\| \mathbf{W}_K\|_2^2\| \mathbf{W}_V\|^2_2 \frac{1}{L^2} \| \mathbf{X}\|^6_2 =\\
    &= \frac{2}{L^3 d_V d_K} \| \mathbf{W}_K\|_2^2\| \mathbf{W}_V\|^2_2 \mathbf{X}\|^6_2
\end{align*}
\begin{align*}
    \| \mathbf{H}_o(\mathbf{W}_V, \mathbf{W}_Q)\|_2 &\leq \frac{2}{L d_V \sqrt{d_K}} \|\mathbf{M}_1^\top \otimes \mathbf{W}_V^\top \|_2 \| \mathbf{Z}_1\|_2 \| \mathbf{I}_{d_V} \otimes \mathbf{W}_K \|_2 \\
    &\leq \frac{2}{L d_V \sqrt{d_K}} L \| \mathbf{X}\|_2 \| \mathbf{W}_V\|_2 \frac{1}{L} \| \mathbf{X}\|^3_2 \| \mathbf{W}_K \|_2 \\
    &= \frac{2}{L d_V \sqrt{d_K}} \| \mathbf{W}_V\|_2 \| \mathbf{W}_K \|_2 \| \mathbf{X}\|^4_2
\end{align*}
\begin{align*}
    \| \mathbf{H}_o(\mathbf{W}_Q, \mathbf{W}_K)\|_2 &\leq \frac{2}{L d_V d_K} \|(\mathbf{I}_{d_V} \otimes \mathbf{W}^\top_K) \mathbf{Z}_1^\top (\mathbf{I}_L \otimes \mathbf{W}_V \mathbf{W}^\top_V) \mathbf{Z}_1 (\mathbf{W}_Q \otimes \mathbf{I}_{d_V}) \mathbf{K}{d_K, d_V}\|_2 \\
    &\leq \frac{2}{L^3 d_V d_K} \|\mathbf{W}_K\|_2 \|\mathbf{W}_Q\|_2 \| \mathbf{W}_V\|_2^2 \|\mathbf{X} \|^6_2
\end{align*}

where we use Properties \ref{prop:matrix_product_norm}, \ref{prop:kronecker_product_norm} and $\| \mathbf{K}_{d_V d_K} \|_2 = 1$, because $\mathbf{K}_{m,n}$ is a commutation matrix from Definition \ref{def:commutation_matrix}. 

Next we derive functional-part estimation.
First we provide analysis for $\mathbf{R}_m = \mathrm{vec}_r (\mathrm{SA} (\mathbf{X}) - \textbf{Target})^T \otimes \mathbf{I}_m$ from Theorem 3.2 from \citep{ormaniec2024attentionhessian}. Since $\mathrm{vec}_r(\cdot)$ is a vectorization procedure $\|\mathrm{vec}_r(\mathrm{SA}(\mathbf{X}) - \textbf{Target})\|_2 = \| \mathrm{SA}(\mathbf{X}) - \textbf{Target} \|_F \leq \sqrt{\mathrm{rank}(\mathrm{SA}(\mathbf{X}) - \textbf{Target} )} \| \mathrm{SA}(\mathbf{X}) - \textbf{Target} \|_2$ according to Property \ref{prop:matrix_norm_inequalities}. Therefore, we obtain 
\begin{align*}
    \| \mathbf{R}_m\| &\leq \sqrt{\mathrm{rank}(\mathrm{SA}(\mathbf{X}) - \textbf{Target})} \|\mathrm{SA}(\mathbf{X}) - \textbf{Target}\|_2 \\ 
    &\leq \sqrt{\mathrm{rank}(\mathrm{SA}(\mathbf{X}) - \textbf{Target})} (\| \mathbf{A} \|_2 \|\mathbf{X}\|_2 \|\mathbf{W}_V \|_2 + \|\textbf{Target}\|_2) \\
    &\leq \sqrt{\mathrm{rank}(\mathrm{SA}(\mathbf{X}) - \textbf{Target})} (L \|\mathbf{X}\|_2 \|\mathbf{W}_V \|_2 + \|\textbf{Target}\|_2)
\end{align*}

where we used Properties \ref{prop:matrix_product_norm}, \ref{prop:matrix_sum_norm}

Next we estimate the shuffling matrix norm, utilizing standard properties
\begin{align*}
    \| \mathbf{S}\|_2 = \|(\mathbf{I}_{d_V} \otimes \mathbf{K}_{d_V,d_V}) (\mathrm{vec}_r (\mathbf{I}_{d_V}) \otimes \mathbf{I}_{d_V})\|_2 \leq \| \mathrm{vec}_r (\mathbf{I}_{d_V}) \|_2 = \| \mathbf{I}_{d_V}\|_{F} = \sqrt{d_V}
\end{align*}

Next challenging part is computing bounds for $\| \frac{\partial^2 \mathbf{A}}{\partial \mathbf{T}^2} \|_2$. In Lemma C1 from \citep{ormaniec2024attentionhessian} the a block form of this expression is provided: \begin{equation*}
\frac{\partial^2 \mathbf{A}_{i,j}}{\partial \mathbf{T}_{i,:} \partial \mathbf{T}_{i,:}} = \mathbf{A}_{i,j} \left(2\mathbf{A}_{i,:}\mathbf{A}_{i,:}^{\top} + \mathbf{E}_{j,j}^{L,L} - \text{diag}(\mathbf{A}_{i,:}) - \mathbf{e}_j \mathbf{A}_{i,:}^{\top} - \mathbf{A}_{i,:}\mathbf{e}_j^{\top}\right) \in \mathbb{R}^{L \times L},
\end{equation*}

where $\mathbf{E}_{j,j}^{L, L} = \mathbf{e}_j \mathbf{e}_j^{\top} \in \mathbb{R}^{L \times L}$ therefore it contains only one non-zero element that equals 1 in $(j, j)$ position. Additionally, it's explicitly said that the second derivative of the row-wise softmax has a block-diagonal structure. Thus, we use block matrix Property \ref{prop:block_matrix_norm}: $\left\|\frac{\partial^2 \mathbf{A}}{\partial \mathbf{T}^2}\right\|_2 = \max_{i,j} \left\|\frac{\partial^2 \mathbf{A}_{i,j}}{\partial \mathbf{T}_{i,:} \partial \mathbf{T}_{i,:}}\right\|_2$. Thus, we conduct $\|\frac{\partial^2 \mathbf{A}_{i,j}}{\partial \mathbf{T}_{i,:} \partial \mathbf{T}_{i,:}} \|_2$ estimation. As we stated before $\mathbf{A}_{i,j} \leq 1$. Now $\| \mathbf{A}_{i,:}\mathbf{A}_{i,:}^{\top} \|_2$: as soon as $\mathbf{A}_{i,:}$ is a row in a softmax matrix, values in it sum up to 1. Thus, we can use the vector-matrix inequalities to obtain: $\| \mathbf{A}_{i,:}\mathbf{A}_{i,:}^{\top} \|_2 \leq \|\mathbf{A}_{i,:}\|^2_2 \leq \|\mathbf{A}_{i,:}\|_1^2 = 1$. After that we conduct $ \|\mathbf{E}_{j,j}^{m,n}\|_2 = \| \mathbf{e}_j \mathbf{e}_j^{\top}\|_2 \leq 1$. Then we estimate $\|diag(\mathbf{A}_{i,:})\|_2$. For diagonal matrices we can easily obtain that $\|diag(\mathbf{A}_{i,:})\|_2 = \max \limits_j \mathbf{A}_{i,j} \leq 1$. Next we estimate $\mathbf{e}_j \mathbf{A}_{i,:}^{\top}$ and $\mathbf{A}_{i,:}\mathbf{e}_j^{\top}$ norms: the matrices $\mathbf{e}_j \mathbf{A}_{i,:}^{\top}$ and $\mathbf{A}_{i,:}\mathbf{e}_j^{\top}$ are rank-1 matrices with only one non-zero row and one non-zero column respectively, containing elements of $\mathbf{A}_{i,:}$. Their spectral norms can be estimated $\|\mathbf{A}_{i,:}\|_2 \leq 1$.

Therefore, we provide an estimation:
\begin{align*}
    \| \frac{\partial^2 \mathbf{A}}{\partial \mathbf{T}^2} \|_2 \leq 6 
\end{align*}

In this way we can easily obtain the $\| \mathbf{Z}_2 \|_2$ estimation
\begin{align*}
    \| \mathbf{Z}_2 \|_2 = \| \left(\mathbf{I}_L \otimes \mathbf{X}^\top \otimes \mathbf{X}^\top \otimes \mathbf{X}^\top\right) \left(\partial^2\mathbf{A}/\partial\mathbf{T}^2\right) \left(\mathbf{X} \otimes \mathbf{X}\right) \|_2 \leq \| \mathbf{X}\|^5_2 \| \frac{\partial^2\mathbf{A}}{\partial\mathbf{T}^2} \|_2 \leq 6 \| \mathbf{X}\|^5_2
\end{align*}

After that, we proceed to the estimation of the functional Hessian norms.
\begin{align*}
\|\mathbf{H}_\mathrm{f}(\mathbf{W}_V, \mathbf{W}_V)\|_2 = 0
\end{align*}
\begin{align*}
\|\mathbf{H}_\mathrm{f}(\mathbf{W}_Q, \mathbf{W}_Q)\|_2 &= \frac{2}{Ld_V d_K} \|\mathbf{R}_{d_V d_K} \left(\mathbf{I}_L \otimes \mathbf{W}_V^\top \otimes \mathbf{I}_{d_V} \otimes \mathbf{W}_K^\top\right) \mathbf{Z}_2 \left(\mathbf{I}_{d_V} \otimes \mathbf{W}_K\right)\|_2, \\
&\leq \frac{2}{Ld_V d_K} \| \mathbf{R}_{d_V d_K} \|_2 \| \mathbf{W}_V \|_2 \| \mathbf{W}_K\|_2 \|\mathbf{Z}_2 \|_2 \| \mathbf{W}_K\|_2 \\
&\leq  \frac{2}{Ld_V d_K}6\sqrt{\mathrm{rank}(\mathrm{SA}(\mathbf{X}) - \textbf{Target})} (L \|\mathbf{X}\|_2 \|\mathbf{W}_V \|_2 + \|\textbf{Target}\|_2)\| \mathbf{W}_V \|_2 \| \mathbf{W}_K\|^2_2\| \mathbf{X}\|^5_2 =\\
&= \frac{12}{d_V d_K}\sqrt{\mathrm{rank}(\mathrm{SA}(\mathbf{X}) - \textbf{Target})} (L \|\mathbf{X}\|_2 \|\mathbf{W}_V \|_2 + \|\textbf{Target}\|_2)\| \mathbf{W}_V \|_2 \| \mathbf{W}_K\|^2_2\| \mathbf{X}\|^5_2
\end{align*}
\begin{align*}
\|\mathbf{H}_\mathrm{f}(\mathbf{W}_V, \mathbf{W}_Q)\|_2 &= \frac{2}{Ld_V\sqrt{d_K}} \|\mathbf{R}_{d_V^2} \left(\mathbf{I}_L \otimes \mathbf{S}\right) \mathbf{Z}_1 \left(\mathbf{I}_{d_V} \otimes \mathbf{W}_K\right)\|_2 \leq \\
&\leq \frac{2}{Ld_V\sqrt{d_K}} \| \mathbf{R}_{d_V^2}\|_2 \| \mathbf{S} \|_2 \|\mathbf{Z}_1 \|_2 \| \mathbf{W}_K\|_2 \leq\\
&\leq \frac{2}{Ld_V\sqrt{d_K}} \sqrt{\mathrm{rank}(\mathrm{SA}(\mathbf{X}) - \textbf{Target})} (L \|\mathbf{X}\|_2 \|\mathbf{W}_V \|_2 + \|\textbf{Target}\|_2) \sqrt{d_V} \frac{1}{L} \|\mathbf{X}\|^3_2\|\mathbf{W}_K\|_2 = \\
&= \frac{2\sqrt{\mathrm{rank}(\mathrm{SA}(\mathbf{X}) - \textbf{Target})}}{L^2\sqrt{d_Vd_K}}(L \|\mathbf{X}\|_2 \|\mathbf{W}_V \|_2 + \|\textbf{Target}\|_2)\|\mathbf{W}_K\|_2 \|\mathbf{X}\|^3_2
\end{align*}
\begin{align*}
\|\mathbf{H}_\mathrm{f}(\mathbf{W}_Q, \mathbf{W}_K)\| &\leq \frac{2}{Ld_V d_K}\| \mathbf{R}_{d_V d_K} \left(\mathbf{I}_L \otimes \mathbf{W}_V^\top \otimes \mathbf{I}_{d_V} \otimes \mathbf{W}_K^\top\right) \mathbf{Z}_2 \left(\mathbf{W}_Q \otimes \mathbf{I}_{d_V}\right) \mathbf{K}_{d_K, d_V}\|_2 + \\
&\quad + \frac{2}{Ld_V\sqrt{d_K}} \|\mathbf{R}_{d_V} \left(\mathbf{I}_L \otimes \mathbf{W}_V^\top \otimes \mathbf{I}_{d_V}\right) \left(\mathbf{Z}_1 \otimes \mathbf{I}_{d_V}\right) \mathbf{S} \otimes \mathbf{I}_{d_K}\|_2 \leq\\
&\leq  \frac{2}{Ld_V d_K}\sqrt{\mathrm{rank}(\mathrm{SA}(\mathbf{X}) - \textbf{Target})} (L \|\mathbf{X}\|_2 \|\mathbf{W}_V \|_2 + \|\textbf{Target}\|_2) \cdot \\ &\cdot\|\mathbf{W}_V\|_2 \|\mathbf{W}_K\|_2 \|\mathbf{W}_Q\|_26 \| \mathbf{X}\|^5_2 + \\
&+\frac{2}{Ld_V\sqrt{d_K}}\sqrt{\mathrm{rank}(\mathrm{SA}(\mathbf{X}) - \textbf{Target})} (L \|\mathbf{X}\|_2 \|\mathbf{W}_V \|_2 + \|\textbf{Target}\|_2) \|\mathbf{W}_V \|_2 \frac{1}{L} \|\mathbf{X}\|^3_2 \sqrt{d_V} =\\
&=\frac{2\sqrt{\mathrm{rank}(\mathrm{SA}(\mathbf{X}) - \textbf{Target})} (L \|\mathbf{X}\|_2 \|\mathbf{W}_V \|_2 + \|\textbf{Target}\|_2)}{Ld_V\sqrt{d_Vd_K}} \|\mathbf{W}_V\|_2 \cdot \\
& \cdot \Big(3L\|\mathbf{W}_K\|_2 \|\mathbf{W}_Q\|_2 \| \mathbf{X}\|^5_2 + \frac{d_V}{L} \|\mathbf{X}\|^3_2\Big),
\end{align*}

Therefore we can obtain the final hessian estimation according to Property \ref{prop:matrix_norm_inequalities}, where we used number of block equal to 3 from $\{K, Q, V\}$:

\begin{align*}
    &\| \mathbf{H} (\mathbf{W}_i, \mathbf{W}_j)\|_2 \leq 3\max \limits_{i,j \in \{Q, K, V\}} \Big(\|\mathbf{H}_o(\mathbf{W}_i, \mathbf{W}_j)\|_2 + \|\mathbf{H}_f(\mathbf{W}_i, \mathbf{W}_j)\|_2\Big)
\end{align*}

And now after substituting results :

\begin{align*}
    &\| \mathbf{H} (\mathbf{W}_i, \mathbf{W}_j)\|_2 \leq\\
    & \leq 3 \max \Bigg(\frac{2L}{d_V} \| \mathbf{X}\|^2_2, \\
    &\quad \frac{2}{L^3 d_V d_K} \| \mathbf{W}_K\|_2^2 \| \mathbf{W}_V\|^2_2 \| \mathbf{X}\|^6_2 + \\
    &+ \frac{12}{d_V d_K} \sqrt{\mathrm{rank}(\mathrm{SA}(\mathbf{X}) - \textbf{Target})} (L \|\mathbf{X}\|_2 \|\mathbf{W}_V \|_2 + \|\textbf{Target}\|_2) \| \mathbf{W}_V \|_2 \| \mathbf{W}_K\|^2_2 \| \mathbf{X}\|^5_2, \\
    &\quad \frac{2}{L d_V \sqrt{d_K}} \| \mathbf{W}_V\|_2 \| \mathbf{W}_K \|_2 \| \mathbf{X}\|^4_2 + \frac{2\sqrt{\mathrm{rank}(\mathrm{SA}(\mathbf{X}) - \textbf{Target})}}{L^2\sqrt{d_V d_K}} (L \|\mathbf{X}\|_2 \|\mathbf{W}_V \|_2 + \|\textbf{Target}\|_2) \|\mathbf{W}_K\|_2 \|\mathbf{X}\|^3_2,\\
    &\quad \frac{2}{L^3 d_V d_K} \|\mathbf{W}_K\|_2 \|\mathbf{W}_Q\|_2 \| \mathbf{W}_V\|^2_2 \|\mathbf{X} \|^6_2 + \\
    &+\frac{2\sqrt{\mathrm{rank}(\mathrm{SA}(\mathbf{X}) - \textbf{Target})} (L \|\mathbf{X}\|_2 \|\mathbf{W}_V \|_2 + \|\textbf{Target}\|_2)}{L d_V \sqrt{d_V d_K}} \|\mathbf{W}_V\|_2 \Big(3L \|\mathbf{W}_K\|_2 \|\mathbf{W}_Q\|_2 \| \mathbf{X}\|^5_2 + \frac{d_V}{L} \|\mathbf{X}\|^3_2\Big)\Bigg)
\end{align*}

The obtained expression we denote as $M$. The obtained inequalities can be simplified by $\mathrm{rank}(\mathrm{SA}(\mathbf{X}) - \textbf{Target}) \le \min(L, d_V)$. That ends the proof. 
\end{proof}

\subsection{Proof of Theorem~\ref{thm:layernorm_derivative}}\label{app:proof_layernorm_derivative}

\begin{proof}
We represent LayerNorm layer as
\begin{equation*}
    \text{LayerNorm}(\mathbf{X}) = \mathbf{P}(\mathbf{X})\mathbf{M}(\mathbf{X})
\end{equation*}

where $\mathbf{P}(\mathbf{X}) = \mathbf{D}^{-1}, \text{ where } \mathbf{D} = \textit{diag}(\sigma(\mathbf{X}))$ and $\mathbf{M}(\mathbf{X}) = (\mathbf{X} - \mu(\mathbf{X})\mathbf{1}^\top_{d_V})$ according to Property \ref{prop:elem_wise_division}.

Using the matrix-product derivative rule from Property \ref{prop:matrix_funcs_product_derivative} we obtain:

\begin{equation*}
    \frac{\partial\text{LayerNorm}(\mathbf{X})}{\partial \mathbf{X}} = ( \mathbf{P}(\mathbf{X}) \otimes \mathbf{I}_{d_V}) \frac{\partial \mathbf{M}}{\partial \mathbf{X}} + (\mathbf{I}_L\otimes \mathbf{M}^\top)\frac{\partial \mathbf{P}}{\partial \mathbf{X}}
\end{equation*}

Let's start with $\frac{\partial \mathbf{M}}{\partial \mathbf{X}}$. Using simple matrix calculus properties we can obtain $\mathbf{M}(\mathbf{X}) = (\mathbf{X} - \mu(\mathbf{X}) \mathbf{1}^{\top}_{d_V}) = (\mathbf{X} - \frac{1}{d_V}\mathbf{X} \mathbf{1}_{d_V} \mathbf{1}^{\top}_{d_V}) = (\mathbf{X} - \frac{1}{d_V}\mathbf{X} \mathbf{1}_{d_V \times d_V})$. Thus, the derivative is

\begin{equation*}
   \frac{\partial \mathbf{M}}{\partial \mathbf{X}} = \frac{\partial  (\mathbf{X} - \frac{1}{d_V}\mathbf{X} \mathbf{1}_{d_V \times d_V})}{\partial \mathbf{X}} = (\mathbf{I}_L \otimes \mathbf{I}_{d_V}) - \frac{1}{d_V}(\mathbf{I}_L \otimes \mathbf{1}_{d_V \times d_V})
\end{equation*}

Next, we calculate the $\frac{\partial \mathbf{P}}{\partial \mathbf{X}}$. First, we start with the transformation of $\sigma(\mathbf{X})$ expression. We can rewrite it in the matrix terms $\sigma(\mathbf{X}) = (\frac{1}{d_V} (\mathbf{X} - \mu(X)\mathbf{1}_{d_V}^\top)^{\circ 2} \mathbf{1}_{d_V})^{\circ \frac{1}{2}} = \frac{1}{\sqrt{d_V}} \left(\mathbf{M}(\mathbf{X})^{\circ 2} \mathbf{1}_{d_V}\right)^{\circ{\frac{1}{2}}}$. Here, $\circ \alpha$ operation is element-wise $\alpha$-powering from Definition \ref{def:vec_elem_ops}.

Therefore, we can apply chain rule and get

\begin{align*}
    \frac{\partial \mathbf{P}}{\partial \mathbf{X}} &= \frac{\partial \mathbf{D}^{-1}}{\partial \mathbf{D}} \frac{\partial\textit{diag}(\sigma(\mathbf{X}))}{\partial \sigma(\mathbf{X})} \frac{\partial \sigma(\mathbf{X})}{\partial \mathbf{X}}
\end{align*}

Therefore, by utilizing Properties \ref{prop:hadamard_square_derivative}, \ref{prop:hadamard_root_derivative} and \ref{prop:matrix_product_derivative} we can find
\begin{equation*}
    \frac{\partial \sigma(\mathbf{X})}{\partial \mathbf{X}} = \frac{1}{\sqrt{d_V}} \frac{\partial \tau^{\circ \frac{1}{2}}}{\partial \tau} \frac{\partial \tau}{\partial \mathbf{Q}} \frac{\partial \mathbf{Q}}{\partial {\mathbf{X}}},
\end{equation*}
Here $\tau = \mathbf{Q}\cdot\mathbf{1}_L$ and $\mathbf{Q} = \mathbf{M}^{\circ{2}}$. Thus, we can continue calculations and obtain

\begin{align*}
    \frac{\partial \sigma(\mathbf{X})}{\partial \mathbf{X}} &= \frac{1}{\sqrt{d_V}} \frac{\partial \tau^{\circ \frac{1}{2}}}{\partial \tau} \frac{\partial \mathbf{Q}\cdot\mathbf{1}_{d_V}}{\partial \mathbf{Q}} \frac{\partial \mathbf{M}^{\circ{2}}}{\partial \mathbf{M}} \frac{\partial \mathbf{M}}{\partial \mathbf{X}} = \\
    \\ &= \frac{1}{\sqrt{d_V}} \frac{1}{2} \textit{diag}^{-1}(\mathrm{vec}_r^{\circ \frac{1}{2}}(\tau)) (\mathbf{I}_L \otimes \mathbf{1}^T_{d_V}) 2\cdot \textit{diag}(\mathrm{vec}_r (\mathbf{M}))\frac{\partial \mathbf{M}}{\partial \mathbf{X}} = \\
    &= \frac{1}{\sqrt{d_V}}\textit{diag}^{-1}(\mathrm{vec}_r^{\circ \frac{1}{2}}(\mathbf{M}^{\circ{2}}\cdot\mathbf{1}_{d_V}))\cdot (\mathbf{I}_L \otimes \mathbf{1}^T_{d_V})\cdot \textit{diag}(\mathrm{vec}_r (\mathbf{M})) \frac{\partial \mathbf{M}}{\partial \mathbf{X}}
\end{align*}

Therefore, by applying \ref{prop:invert_derivative} and \ref{prop:diag_derivative} for the first and second multiplier, we obtain

\begin{align*}
    \frac{\partial \mathbf{P}}{\partial \mathbf{X}} &= \frac{1}{\sqrt{d_V}}\left(-\mathbf{D}^{-1} \otimes \mathbf{D}^{-\top} \right) \Big(\mathbf{e}_1 \otimes \mathbf{e}_1 \quad \dots  \quad \mathbf{e}_L \otimes \mathbf{e}_L\Big) \cdot \\
    &\left(\textit{diag}^{-1}(\mathrm{vec}_r^{\circ \frac{1}{2}}(\mathbf{M}^{\circ{2}}\cdot\mathbf{1}_{d_V}))\cdot (\mathbf{I}_L \otimes \mathbf{1}^T_{d_V})\cdot \textit{diag}(\mathrm{vec}_r (\mathbf{M}))\frac{\partial \mathbf{M}}{\partial \mathbf{X}}\right)
\end{align*}

Therefore, we found the first derivative of the LayerNorm function:

\begin{align*}
    \frac{\partial\text{LayerNorm}(\mathbf{X})}{\partial \mathbf{X}} &= ( \mathbf{P}(\mathbf{X}) \otimes \mathbf{I}_{d_V}) \frac{\partial \mathbf{M}}{\partial \mathbf{X}} + (\mathbf{I}_L\otimes \mathbf{M}^\top)\frac{\partial \mathbf{P}}{\partial \mathbf{X}} = \\
    & = ( \mathbf{P}(\mathbf{X}) \otimes \mathbf{I}_{d_V}) \frac{\partial \mathbf{M}}{\partial \mathbf{X}} +\\
    &+ (\mathbf{I}_L\otimes \mathbf{M}^\top)\frac{1}{\sqrt{d_V}}\left(-\mathbf{D}^{-1} \otimes \mathbf{D}^{-\top} \right) \Big(\mathbf{e}_1 \otimes \mathbf{e}_1 \quad \dots  \quad \mathbf{e}_L \otimes \mathbf{e}_L\Big) \cdot \\
    &\cdot \left(\textit{diag}^{-1}(\mathrm{vec}_r^{\circ \frac{1}{2}}(\mathbf{M}^{\circ{2}}\cdot\mathbf{1}_{d_V}))\cdot (\mathbf{I}_L \otimes \mathbf{1}^T_{d_V})\cdot \textit{diag}(\mathrm{vec}_r (\mathbf{M}))\frac{\partial \mathbf{M}}{\partial \mathbf{X}}\right)
\end{align*}

where $\mathbf{M}(\mathbf{X}) = (\mathbf{X} - \frac{1}{d_V}\mathbf{X} \mathbf{1}_{d_V \times d_V})$, $\mathbf{P}(\mathbf{X}) = \textit{diag}^{-1}(\sigma(\mathbf{X}))$ and $\frac{\partial \mathbf{M}}{\partial \mathbf{X}} = (\mathbf{I}_L \otimes \mathbf{I}_{d_V}) - \frac{1}{d_V}(\mathbf{I}_L \otimes \mathbf{1}_{d_V \times d_V})$

That ends the proof.

\end{proof}

\subsection{Proof of Theorem~\ref{thm:layernorm_second_derivative}}\label{app:proof_layernorm_second_derivative}

\begin{proof}
    Now, we calculate the second derivative $\frac{\partial^2\text{LayerNorm}}{\partial \mathbf{X}^2}$. Using the matrix product derivative property \ref{prop:matrix_product_derivative}, we obtain:
\begin{align*}
    \frac{\partial^2\text{LayerNorm}}{\partial \mathbf{X}^2} &= \left( (\mathbf{P}(\mathbf{X})  \otimes  \mathbf{I}_{d_V}) \otimes \mathbf{I}_{Ld_V} \right)\frac{\partial^2 \mathbf{M}}{\partial \mathbf{X}^2} + \left( \mathbf{I}_{Ld_V} \otimes  (\frac{\partial \mathbf{M}}{\partial \mathbf{X}})^\top \right) \frac{\partial (\mathbf{P}(\mathbf{X}) \otimes \mathbf{I}_{d_V})}{\partial \mathbf{X}} + \\
    &+ \left( (\mathbf{I}_L \otimes \mathbf{M}^\top ) \otimes \mathbf{I}_{L d_V}  \right) \frac{\partial^2 \mathbf{P}}{\partial \mathbf{X}^2} + \left( \mathbf{I}_{L d_V} \otimes (\frac{\partial \mathbf{P}}{\partial \mathbf{X}})^\top \right) \frac{\partial (\mathbf{I}_L \otimes \mathbf{M}^\top )}{\partial\mathbf{X}}
\end{align*}

Here, we have $\mathbf{P} \in \mathbb{R}^{L \times L}$, $\mathbf{M} \in \mathbb{R}^{L \times d_V}$, $\frac{\partial \mathbf{M}}{\partial \mathbf{X}} \in \mathbb{R}^{Ld_V \times Ld_V}$, $\frac{\partial \mathbf{P}}{\partial \mathbf{X}} \in \mathbb{R}^{L^2 \times Ld_V}$

Next, we can easily obtain, using Properties \ref{prop:kronecker_product_derivative}, \ref{prop:transposed_matrix_derivative}: 
\begin{align*}
    \frac{\partial^2 \mathbf{M}}{\partial \mathbf{X}^2} &= 0 \\
    \frac{\partial (\mathbf{P}(\mathbf{X}) \otimes \mathbf{I}_{d_V} )}{\partial \mathbf{X}} &= \frac{\partial (\mathbf{P} \otimes \mathbf{I}_L)}{\partial \mathbf{P}} \frac{\partial \mathbf{P}}{\partial\mathbf{X}} = \left(\mathbf{I}_L \otimes \mathbf{K}_{L, L} \otimes \mathbf{I}_L \right) \left(\mathbf{I}_{L^2} \otimes  \mathrm{vec}_r(\mathbf{I}_L)  \right) \frac{\partial \mathbf{P}}{\partial \mathbf{X}} \\
    \frac{\partial (\mathbf{I}_L \otimes \mathbf{M}^\top )}{\partial\mathbf{X}} &= \frac{\partial ( \mathbf{I}_L \otimes \mathbf{M}^\top)}{\partial \mathbf{M}^\top} \frac{\partial \mathbf{M}^\top}{\partial \mathbf{M}} \frac{\partial \mathbf{M}}{\partial \mathbf{X}} = \left(\mathbf{I}_{L} \otimes \mathbf{K}_{d_V,L} \otimes \mathbf{I}_L \right) \left(\mathrm{vec}_r(\mathbf{I}_L) \otimes  \mathbf{I}_{L d_V} \right) \mathbf{K}_{d_V, L} \frac{\partial \mathbf{M}}{\partial \mathbf{X}}
\end{align*}

Now, we analyze the second-order derivative of the $\mathbf{P}$ matrix. To derive correct calculations we need to write the dimensions of each multiplier in the calculated first derivative out. Matrix $\mathbf{D}$ is a $\textit{diag}(\sigma(\mathbf{X}))$, the size of vector $\sigma(\mathbf{X})$ is $L\times 1$, therefore, $\mathbf{D} \in \mathbb{R}^{L\times L}$ and the part $\left(-\mathbf{D}^{-1} \otimes \mathbf{D}^{-\top} \right) \in \mathbb{R}^{L^2 \times L^2}$. Next, we note that the size of each basis vector $\mathbf{e}_i$ is $L\times1$, thus we obtain $\mathbf{e}_i \otimes \mathbf{e}_i \in \mathbb{R}^{L^2 \times 1}$ and $ \Big(\mathbf{e}_1 \otimes \mathbf{e}_1 \quad \dots  \quad \mathbf{e}_L \otimes \mathbf{e}_L\Big) \in \mathbb{R}^{L^2 \times L}$. As we discussed earlier, $\mathbf{M}(\mathbf{X}) \in \mathbb{R}^{L \times d_V}$, then $M\cdot\mathbf{1}_{d_V} \in \mathbb{R}^{L \times 1}$, and we can derive the size of $\textit{diag}^{-1}(\mathrm{vec}_r^{\circ \frac{1}{2}}(\mathbf{M}^{\circ{2}}\cdot\mathbf{1}_{d_V}))$, which is $L \times L$. Next multipliers are $(\mathbf{I}_L \otimes \mathbf{1}^T_{d_V}) \in \mathbb{R}^{L \times Ld_V}$ and $\textit{diag}(\mathrm{vec}_r (M)) \in \mathbb{R}^{Ld_V \times Ld_V}$. The last one is $\frac{\partial \mathbf{M}}{\partial \mathbf{X}}$, which we have already calculated, it's size is $Ld_V \times Ld_V$. Therefore, the whole derivative $\frac{\partial \mathbf{P}}{\partial \mathbf{X}}$ is from $\mathbb{R}^{L^2 \times Ld_V}$.
 
We start with $\frac{\partial \mathbf{P}}{\partial \mathbf{X}} = \frac{1}{\sqrt{d_V}} \mathbf{A}_1(\mathbf{X})\cdot \mathbf{B}_1(\mathbf{X})$, where $\mathbf{A}_1 = \left(-\mathbf{D}^{-1} \otimes \mathbf{D}^{-\top} \right)$ and $\mathbf{B}_1$ is the other multiplier.

Therefore, using Property \ref{prop:matrix_funcs_product_derivative} we obtain 
\begin{align*}
    \frac{\partial^2 \mathbf{P}}{\partial \mathbf{X}^2} = \frac{1}{\sqrt{d_V}} \frac{\partial \mathbf{A}_1(\mathbf{X})\cdot \mathbf{B}_1(\mathbf{X})}{\partial \mathbf{X}} = \frac{1}{\sqrt{d_V}} \left(\mathbf{A}_1 \otimes  \mathbf{I}_{Ld_V}  \right) \frac{\partial \mathbf{B}_1}{\partial \mathbf{X}} + \left( \mathbf{I}_{L^2} \otimes  \mathbf{B}_1^\top \right) \frac{\partial \mathbf{A}_1}{\partial \mathbf{X}}
\end{align*}

Now we focus on calculating $\frac{\partial \mathbf{A}_1}{\partial \mathbf{X}}$ on the current step. Utilising the rule \ref{prop:matrix_funcs_kronecker_product_derivative} we can simply get:

\begin{align*}
    \frac{\partial \mathbf{A}_1}{\partial \mathbf{X}} = \frac{\partial \left(-\mathbf{D}^{-1} \otimes \mathbf{D}^{-\top} \right)}{\partial \mathbf{X}} = \left(\mathbf{I}_L \otimes \mathbf{K}_{L, L} \otimes \mathbf{I}_L \right) &\Big((\mathbf{I}_{L^2} \otimes \mathrm{vec}_r(\mathbf{\mathbf{D}^{-\top}})) \cdot \frac{\partial -\mathbf{\mathbf{D}^{-1}}}{\partial \mathbf{X}}+ \\
    &+ (\mathrm{vec}_r(-\mathbf{\mathbf{D}^{-1}}) \otimes \mathbf{I}_{L^2}) \cdot \frac{\partial \mathbf{\mathbf{D}^{-\top}}}{\partial \mathbf{X}} \Big)
\end{align*}

By using the transposed matrix and the invert matrix derivative properties \ref{prop:transposed_matrix_derivative}, \ref{prop:invert_derivative}, we obtain: $\frac{\partial -\mathbf{\mathbf{D}^{-1}}}{\partial \mathbf{X}} = \frac{\partial -\mathbf{\mathbf{D}^{-1}}}{\partial \mathbf{D}}\frac{\partial \mathbf{D}}{\partial \mathbf{X}} = \left( \mathbf{D}^{-1} \otimes \mathbf{D}^{-\top}\right) \frac{\partial \mathbf{D}}{\partial \mathbf{X}}$ and $\frac{\partial \mathbf{\mathbf{D}^{-\top}}}{\partial \mathbf{X}} = \frac{\partial \mathbf{\mathbf{D}^{-\top}}}{\partial \mathbf{D}^{-1}}\frac{\partial \mathbf{\mathbf{D}^{-1}}}{\partial \mathbf{D}} \frac{\partial \mathbf{\mathbf{D}}}{\partial \mathbf{X}} = \mathbf{K}_{L, L} \left(-\mathbf{D}^{-1} \otimes \mathbf{D}^{-\top} \right) \frac{\partial \mathbf{\mathbf{D}}}{\partial \mathbf{X}}$, where we the $\frac{\partial \mathbf{\mathbf{D}}}{\partial \mathbf{X}}$ as we calculated earlier, while computing the first LayerNorm's derivative is $\frac{\partial \mathbf{\mathbf{D}}}{\partial \mathbf{X}} = \Big(\mathbf{e}_1 \otimes \mathbf{e}_1 \quad \dots  \quad \mathbf{e}_L \otimes \mathbf{e}_L\Big)\left(\textit{diag}^{-1}(\mathrm{vec}_r^{\circ \frac{1}{2}}(\mathbf{M}^{\circ{2}}\cdot\mathbf{1}_{d_V}))\cdot (\mathbf{I}_L \otimes \mathbf{1}^T_{d_V})\cdot \textit{diag}(\mathrm{vec}_r (\mathbf{M}))\frac{\partial \mathbf{M}}{\partial \mathbf{X}}\right)$

And now we proceed to the calculations of the remaining part derivative. 

We first assign new $\mathbf{A}_2$ and $\mathbf{B}_2$ for clear calculations. We have $\mathbf{B}_1 = \Big(\mathbf{e}_1 \otimes \mathbf{e}_1 \quad \dots  \quad \mathbf{e}_L \otimes \mathbf{e}_L\Big)\left(\textit{diag}^{-1}(\mathrm{vec}_r^{\circ \frac{1}{2}}(\mathbf{M}^{\circ{2}}\cdot\mathbf{1}_{d_V}))\cdot (\mathbf{I}_L \otimes \mathbf{1}^T_{d_V})\cdot \textit{diag}(\mathrm{vec}_r (\mathbf{M}))\frac{\partial \mathbf{M}}{\partial \mathbf{X}}\right)$ and we assign new $\mathbf{A}_2$ and new $\mathbf{B}_2$ as $\mathbf{A}_2 = \textit{diag}^{-1}(\mathrm{vec}_r^{\circ \frac{1}{2}}(\mathbf{M}^{\circ{2}}\cdot\mathbf{1}_{d_V}))$, $\mathbf{B}_2 = (\mathbf{I}_L \otimes \mathbf{1}^T_{d_V})\cdot \textit{diag}(\mathrm{vec}_r (\mathbf{M}))\frac{\partial \mathbf{M}}{\partial \mathbf{X}}$ and we denote $\mathbf{E} = \Big(\mathbf{e}_1 \otimes \mathbf{e}_1 \quad \dots  \quad \mathbf{e}_L \otimes \mathbf{e}_L\Big)$. Thus, $\mathbf{B}_1 = \mathbf{E} \mathbf{A}_2 \mathbf{B}_2$

While $\mathbf{E}$ is a constant matrix we can apply the simplified matrix product derivative rule \ref{prop:matrix_funcs_product_derivative} and obtain

\begin{align*}
    \frac{\partial \mathbf{B}_1}{\partial \mathbf{X}} &= \frac{\partial \mathbf{E} \mathbf{A}_2 \mathbf{B}_2}{\partial (\mathbf{A}_2 \mathbf{B}_2)} \frac{\partial \mathbf{A}_2 \mathbf{B}_2}{\partial \mathbf{X}} = \left( \mathbf{E} \otimes \mathbf{I}_{L d_V}\right)\frac{\partial \mathbf{A}_2 \mathbf{B}_2}{\partial \mathbf{X}} \\
    &= \left( \mathbf{E} \otimes \mathbf{I}_{L d_V}\right) \left( (\mathbf{A}_2\otimes \mathbf{I}_{L d_V} )\frac{\partial \mathbf{B}_2}{\partial \mathbf{X}} + (\mathbf{I}_L \otimes \mathbf{B}_2^\top) \frac{\partial \mathbf{A}_2}{\partial \mathbf{X}}\right)
\end{align*}

Now, we introduce the last $\mathbf{A}_3$ and $\mathbf{B}_3$ assignment. We represent $\mathbf{B}_2$ as $\mathbf{B}_2 = \mathbf{J} \mathbf{A}_3 \mathbf{B}_3$, where $\mathbf{J} = (\mathbf{I}_L \otimes \mathbf{1}^T_{d_V})$, $\mathbf{A}_3 = \textit{diag}(\mathrm{vec}_r (\mathbf{M}))$ and $\mathbf{B}_3 = \frac{\partial \mathbf{M}}{\partial \mathbf{X}}$.

Similarly to the previous step we firstly apply simplified matrix product derivative rule \ref{prop:matrix_funcs_product_derivative} and get

\begin{align*}
    \frac{\partial \mathbf{B}_2}{\partial \mathbf{X}} &= \frac{\partial \mathbf{J} \mathbf{A}_3 \mathbf{B}_3}{\partial (\mathbf{A}_3 \mathbf{B}_3)} \frac{\partial \mathbf{A}_3 \mathbf{B}_3}{\partial \mathbf{X}} = \left( \mathbf{J} \otimes \mathbf{I}_{L d_V}\right)\frac{\partial \mathbf{A}_3 \mathbf{B}_3}{\partial \mathbf{X}} \\
    &= \left( \mathbf{J} \otimes \mathbf{I}_{L d_V}\right) \left((\mathbf{A}_3 \otimes \mathbf{I}_{Ld_V} ) \frac{\partial \mathbf{B}_3}{\partial \mathbf{X}} + ( \mathbf{I}_{Ld_V} \otimes \mathbf{B}_3^\top)\frac{\partial\mathbf{A}_3}{\partial \mathbf{X}}\right)
\end{align*}

Where both Jacobian matrices can be found easily $\frac{\partial\mathbf{A}_3}{\partial \mathbf{X}} = \frac{\partial \textit{diag}(\mathrm{vec}_r(\mathbf{M}))}{\partial \mathbf{X}} = \frac{\partial \textit{diag}(\mathbf{v})}{\partial (\mathbf{v})} \frac{\partial \mathrm{vec}_r(\mathbf{M})}{\partial \mathbf{M}} \frac{\partial \mathbf{M}}{\partial \mathbf{X}}$

Where we have already calculated $\frac{\partial \textit{diag}(\mathbf{v})}{\partial (\mathbf{v})} = \Big(\mathbf{e}_1 \otimes \mathbf{e}_1 \quad \dots  \quad \mathbf{e}_L \otimes \mathbf{e}_L\Big)$ according to the property \ref{prop:diag_derivative}, here $\mathbf{e}_i \in \mathbb{R}^{Ld_V \times 1}$, additionally $\frac{\partial \mathrm{vec}_r(\mathbf{M})}{\partial \mathbf{M}}$ is simply $\mathbf{I}_{L d_V}$. As for $\frac{\partial \mathbf{B}_3}{\partial \mathbf{X}}$ for current $\mathbf{B}$ it is $\frac{\partial \mathbf{B}_3}{\partial \mathbf{X}} = \frac{\partial^2 \mathbf{M}}{\partial\mathbf{X}^2} = 0$

The last step in our analysis is putting every part of our calculations together. In our notation we can simplify the expression

\begin{align*}
    \frac{\partial^2 \mathbf{P}}{\partial \mathbf{X}^2} &= \frac{1}{\sqrt{d_V}} \left(\mathbf{A}_1 \otimes  \mathbf{I}_{Ld_V}  \right) \frac{\partial \mathbf{B}_1}{\partial \mathbf{X}} + \left( \mathbf{I}_{L^2} \otimes  \mathbf{B}_1^\top \right) \frac{\partial \mathbf{A}_1}{\partial \mathbf{X}}
\end{align*}

where $\frac{\partial \mathbf{B}_1}{\partial \mathbf{X}}$, $\frac{\partial \mathbf{A}_1}{\partial \mathbf{X}} \mathbf{B}_1$ and it's definitions $\mathbf{A}_1$, $\mathbf{B}_1$ are given above.

The last step in the proof is simply combining all together and substituting all calculated derivatives into the LayerNorm's Hessian.

That ends the proof.
\end{proof}

\subsection{Proof of Theorem~\ref{thm:transformer_derivative}}\label{app:proof_transformer_derivative}
\begin{proof}
It's worth noting that in our notation $\mathbf{X} \in R^{L \times d_V}, \mathbf{Y} \in R^{L\times d_V}, \mathbf{W}_1 \in R^{d_V \times d_{ff}}, \text{ReLU}(\mathbf{Y\mathbf{W}_1}) \in R^{L \times d_{ff}}, \mathbf{W}_2 \in R^{d_{ff} \times d_V} $.
 
We consider the Transformer block as it's defined in \ref{eq:transformer}, explicitly:

\[
\mathbf{Y} = \text{LayerNorm}(\mathrm{SA}(\mathbf{X}) + \mathbf{X}),
\]
\[
\mathbf{Z} = \text{LayerNorm}(\text{FFN}(\mathbf{Y}) + \mathbf{Y}),
\]

We derive calculations for the first derivative of the whole transformer block $\frac{\partial\mathbf{Z}}{\partial \mathbf{W}_i}$. 

For $i \in \{1,2\}$:
\begin{equation*}
    \frac{\partial\mathbf{Z}}{\partial \mathbf{W}_i} = \frac{\partial \text{LayerNorm}(\text{FFN}(\mathbf{Y}) + \mathbf{Y})}{\partial (\text{FFN}(\mathbf{Y}) + \mathbf{Y})} \frac{\partial (\text{FFN}(\mathbf{Y}) + \mathbf{Y})}{\partial \mathbf{W}_i}
\end{equation*}

where 

\begin{equation*}
    \frac{\partial (\text{FFN}(\mathbf{Y}) + \mathbf{Y})}{\partial \mathbf{W}_i} = \frac{\partial (\text{FFN}(\mathbf{Y}))}{\partial \mathbf{W}_i} = \frac{\partial \mathbf{I}_L \sigma(\mathbf{Y}\mathbf{W}_1)\mathbf{W}_2 \mathbf{I}_{d_V}}{\partial \mathbf{W}_i}
\end{equation*}

Therefore, using Property \ref{prop:matrix_product_derivative}:

\begin{align*}
    \text{for } i = 2:&
    \frac{\partial \mathbf{I}_L \sigma(\mathbf{Y}\mathbf{W}_1)\mathbf{W}_2 \mathbf{I}_{d_V}}{\partial \mathbf{W}_i} = \sigma(\mathbf{Y} \mathbf{W}_1) \otimes \mathbf{I}_{d_V}\\
    \text{for } i = 1:& \frac{\partial \mathbf{I}_L \sigma(\mathbf{Y}\mathbf{W}_1)\mathbf{W}_2 \mathbf{I}_{d_V}}{\partial \mathbf{W}_i} = \frac{\partial \sigma(\mathbf{Y}\mathbf{W}_1) \mathbf{W}_2}{\partial \sigma(\mathbf{Y}\mathbf{W}_1)} \frac{\partial \sigma(\mathbf{Y}\mathbf{W}_1)}{\partial \mathbf{Y}\mathbf{W}_1} \frac{\partial \mathbf{Y}\mathbf{W}_1}{\partial \mathbf{W}_1} \\
    &= \left(\mathbf{I}_L \otimes \mathbf{W}_2^\top \right) \frac{\partial \sigma(\mathbf{Y}\mathbf{W}_1)}{\partial \mathbf{Y}\mathbf{W}_1} \left( \mathbf{I}_L \otimes \mathbf{W}_1^\top\right)
\end{align*}

According to Lemma \ref{lemma:relu_derivative_hessian}, we obtain

\begin{align*}
    \text{for } i = 1:& \frac{\partial \mathbf{I}_L \sigma(\mathbf{Y}\mathbf{W}_1)\mathbf{W}_2 \mathbf{I}_{d_V}}{\partial \mathbf{W}_i} = \left(\mathbf{I}_L \otimes \mathbf{W}_2^\top \right) \mathrm{diag}\!\big(\mathrm{vec}_r(\mathbf{1}_{\{\mathbf{X}>0\}})\big) \left( \mathbf{Y} \otimes \mathbf{I}_{d_{ff}}\right)
\end{align*}

Thus for $i \in \{1, 2\}$ the following holds:

\begin{equation*}
    \frac{\partial (\text{FFN}(\mathbf{Y}) + \mathbf{Y})}{\partial \mathbf{W}_i} = \begin{cases}
        \left(\mathbf{I}_L \otimes \mathbf{W}_2^\top \right) \mathrm{diag}\!\big(\mathrm{vec}_r(\mathbf{1}_{\{\mathbf{X}>0\}})\big) \left( \mathbf{Y} \otimes \mathbf{I}_{d_{ff}}\right), \text{for } i = 1 \\
        \sigma(\mathbf{Y} \mathbf{W}_1) \otimes \mathbf{I}_{d_V}, \text{for } i = 2
    \end{cases}
\end{equation*}

and the whole Transformer block derivative can be calculated as:

\begin{equation*}
    \frac{\partial\mathbf{Z}}{\partial \mathbf{W}_i} = \begin{cases}
        \frac{\partial \text{LayerNorm}(\text{FFN}(\mathbf{Y}) + \mathbf{Y})}{\partial (\text{FFN}(\mathbf{Y}) + \mathbf{Y})}\left(\mathbf{I}_L \otimes \mathbf{W}_2^\top \right) \mathrm{diag}\!\big(\mathrm{vec}_r(\mathbf{1}_{\{\mathbf{X}>0\}})\big) \left( \mathbf{Y} \otimes \mathbf{I}_{d_{ff}}\right), \text{for } i = 1 \\
        \frac{\partial \text{LayerNorm}(\text{FFN}(\mathbf{Y}) + \mathbf{Y})}{\partial (\text{FFN}(\mathbf{Y}) + \mathbf{Y})}\sigma(\mathbf{Y} \mathbf{W}_1) \otimes \mathbf{I}_{d_V}, \text{for } i = 2
    \end{cases}
\end{equation*}

where according to Theorem \ref{thm:layernorm_derivative}

\begin{align*}
    &\frac{\partial\text{LayerNorm}(\text{FFN}(\mathbf{Y}) + \mathbf{Y})}{\partial (\text{FFN}(\mathbf{Y}) + \mathbf{Y})} = ( \mathbf{P}(\text{FFN}(\mathbf{Y}) + \mathbf{Y}) \otimes \mathbf{I}_{d_V}) \frac{\partial \mathbf{M}}{\partial (\text{FFN}(\mathbf{Y}) + \mathbf{Y})} +\\
    &+ (\mathbf{I}_L\otimes \mathbf{M}^\top)\frac{1}{\sqrt{d_V}}\left(-\mathbf{D}^{-1} \otimes \mathbf{D}^{-\top} \right) \Big(\mathbf{e}_1 \otimes \mathbf{e}_1 \quad \dots  \quad \mathbf{e}_L \otimes \mathbf{e}_L\Big) \cdot\\
    &\cdot \left(\textit{diag}^{-1}(\mathrm{vec}_r^{\circ \frac{1}{2}}(\mathbf{M}^{\circ{2}}\cdot\mathbf{1}_{d_V}))\cdot (\mathbf{I}_L \otimes \mathbf{1}^T_{d_V})\cdot \textit{diag}(\mathrm{vec}_r (\mathbf{M}))\frac{\partial \mathbf{M}}{\partial (\text{FFN}(\mathbf{Y}) + \mathbf{Y})}\right)
\end{align*}

where $\mathbf{M}(\text{FFN}(\mathbf{Y}) + \mathbf{Y}) = ((\text{FFN}(\mathbf{Y}) + \mathbf{Y}) - \frac{1}{d_V}(\text{FFN}(\mathbf{Y}) + \mathbf{Y}) \mathbf{1}_{d_V \times d_V})$, $\mathbf{P}((\text{FFN}(\mathbf{Y}) + \mathbf{Y})) = \textit{diag}^{-1}(\sigma(\text{FFN}(\mathbf{Y}) + \mathbf{Y})$ and $\frac{\partial \mathbf{M}}{\partial(\text{FFN}(\mathbf{Y}) + \mathbf{Y})} = (\mathbf{I}_L \otimes \mathbf{I}_{d_V}) - \frac{1}{d_V}(\mathbf{I}_L \otimes \mathbf{1}_{d_V \times d_V})$, and here $\sigma$ is simply calculated according to the LayerNorm definition.

Next, we derive calculations for $i \in \{ K, Q, V\}$

\begin{align*}
    \frac{\partial\mathbf{Z}}{\partial \mathbf{W}_i} = \frac{\partial \text{LayerNorm}(\text{FFN}(\mathbf{Y}) + \mathbf{Y})}{\partial (\text{FFN}(\mathbf{Y}) + \mathbf{Y})} \frac{\partial (\text{FFN}(\mathbf{Y}) + \mathbf{Y})}{\partial \mathbf{Y}} \frac{\partial \mathbf{Y}}{\partial \mathbf{W}_i}
\end{align*}

Utilizing Property \ref{prop:matrix_product_derivative} and Lemma \ref{lemma:relu_derivative_hessian}, we obtain:

\begin{align*}
    \frac{\partial (\text{FFN}(\mathbf{Y}) + \mathbf{Y})}{\partial \mathbf{Y}} &= \frac{\partial \text{FFN}(\mathbf{Y}) }{\partial \mathbf{Y}} + \frac{\partial \mathbf{Y}}{\partial \mathbf{Y}} = \frac{\partial \text{FFN}(\mathbf{Y}) }{\partial \mathbf{Y}} + \left( \mathbf{I}_L \otimes \mathbf{I}_{d_V}\right) = \frac{\partial \sigma(\mathbf{Y} \mathbf{W}_1) \mathbf{W}_2}{\partial \mathbf{Y}} +\left( \mathbf{I}_L \otimes \mathbf{I}_{d_V}\right)=\\ 
    &= \left( \mathbf{I}_L \otimes \mathbf{W}_2^\top\right) \frac{\partial \sigma (\mathbf{Y} \mathbf{W}_1)}{\partial \mathbf{Y}\mathbf{W}_1} \frac{\partial \mathbf{Y}\mathbf{W}_1}{\partial \mathbf{Y}} + \left( \mathbf{I}_L \otimes \mathbf{I}_{d_V}\right) = \\
    & = \left( \mathbf{I}_L \otimes \mathbf{W}_2^\top\right) \mathrm{diag}\!\big(\mathrm{vec}_r(\mathbf{1}_{\{\mathbf{X}>0\}})\big) \left( \mathbf{I}_L \otimes \mathbf{W}_1^\top \right) + \left( \mathbf{I}_L \otimes \mathbf{I}_{d_V}\right)
\end{align*}

and for calculating $\frac{\partial \mathbf{Y}}{\partial \mathbf{W}_i}$ we use Lemma A.2 from \citep{noci2022signalpropagationtransformerstheoretical}: 
\begin{align*}
\frac{\partial \mathrm{SA}}{\partial \mathbf{W}_V} &= \text{softmax}\left(\frac{\mathbf{X}\mathbf{W}_Q\mathbf{W}_{K}^{\top}\mathbf{X}^\top}{\sqrt{d_K}}\right) \mathbf{X} \otimes \mathbf{I}_{d_V}\\
\frac{\partial \mathrm{SA}}{\partial \mathbf{W}_Q} &= \left(\mathbf{I}_L \otimes \mathbf{W}_{V}^{\top}\mathbf{X}^\top\right) \frac{\partial \mathbf{A}}{\partial \mathbf{M}} \left(\frac{\mathbf{X} \otimes \mathbf{X}\mathbf{W}_K}{\sqrt{d_K}}\right),
\end{align*}

\textit{where:}
\begin{equation*}
\frac{\partial \mathbf{A}}{\partial \mathbf{M}} = \text{blockdiag}\left(\frac{\partial \mathbf{A}_i}{\partial \mathbf{M}_i^\top}\right)
\end{equation*}

\textit{and } $\frac{\partial \mathbf{A}_i}{\partial \mathbf{M}_i^\top} = \text{diag}(\mathbf{A}_i) - \mathbf{A}_i\mathbf{A}_i^\top$, where $\mathbf{A}_i$ \text{is the i-th row of} $\mathbf{A}$ in a column vector format. Finally, under the uniform-attention assumption it simplifies to:
\begin{equation*}
\frac{\partial \mathbf{A}}{\partial \mathbf{M}} = \frac{1}{n}\mathbf{I}_L \otimes \left(\mathbf{I}_L - \frac{1}{L}\mathbf{1}_{L \times L}\right)
\end{equation*}

Additionally, we can easily expand the result on $\mathbf{W}_K$, where we apply the property \ref{prop:transposed_matrix_derivative}, therefore:  

\begin{align*}
\frac{\partial \mathrm{SA}}{\partial \mathbf{W}_K} &= \left(\mathbf{I}_L \otimes \mathbf{W}_{V}^{\top}\mathbf{X}^\top\right) \frac{\partial \mathbf{A}}{\partial \mathbf{M}} \left(\frac{(\mathbf{X} \mathbf{W}_Q \otimes \mathbf{X})\mathbf{K}_{d_V d_K}}{\sqrt{d_k}}\right),
\end{align*}

Thus $\frac{\partial \mathbf{Y}}{\partial\mathbf{W}_i}$ can be calculated as follows:

\begin{align*}
    \frac{\partial \mathbf{Y}}{\partial\mathbf{W}_i} = \frac{\partial \text{LayerNorm}(\mathrm{SA} (\mathbf{X}) + \mathbf{X})}{\partial \mathbf{W}_i} = \frac{\partial \text{LayerNorm}(\mathrm{SA} (\mathbf{X}) + \mathbf{X})}{\partial (\mathrm{SA} (\mathbf{X}) + \mathbf{X})} \frac{\partial \mathrm{SA}(\mathbf{X})}{\partial \mathbf{W}_i}
\end{align*}

where $\frac{\partial \mathrm{SA}(\mathbf{X})}{\partial \mathbf{W}_i}$ is calculated according to Lemma A.2 from \citep{noci2022signalpropagationtransformerstheoretical}, which we mentioned earlier above and $\frac{\partial \text{LayerNorm}(\mathrm{SA} (\mathbf{X}) + \mathbf{X})}{\partial (\mathrm{SA} (\mathbf{X}) + \mathbf{X})}$ is calculated according to Theorem \ref{thm:layernorm_derivative}.

Substituting the expressions ends the proof.
\end{proof}

\subsection{Proof of Theorem~\ref{thm:transformer_hessian}}\label{app:proof_transformer_hessian}
\begin{proof}
We differentiate the Jacobian from Theorem~\ref{thm:transformer_derivative} using Proposition~\ref{prop:matrix_funcs_product_derivative} (matrix-product derivative), Proposition~\ref{prop:kronecker_product_derivative} (Kronecker-product derivative), Proposition~\ref{prop:transposed_matrix_derivative}, the Identification Theorem~\ref{prop:identification_theorem_vec_r}, and Lemma~\ref{lemma:relu_derivative_hessian}.

Step 1. For any $i \in \{1,2,K,Q,V\}$ we have
\[\frac{\partial \mathbf{Z}}{\partial \mathbf{W}_i} \;=\; \mathbf{J}_Z \, \mathbf{B}_i,\qquad\mathbf{J}_Z \in \mathbb{R}^{L d_V \times L d_V},\]
where $\mathbf{B}_i := \frac{\partial \mathbf{S}}{\partial \mathbf{W}_i}$ is given casewise by
\[\mathbf{B}_1 = (\mathbf{I}_L \otimes \mathbf{W}_2^\top)\, \mathbf{D}_\sigma\, (\mathbf{Y} \otimes \mathbf{I}_{d_{ff}}) \in \mathbb{R}^{L d_V \times n_1},\quad\mathbf{B}_2 = \sigma(\mathbf{Y}\mathbf{W}_1) \otimes \mathbf{I}_{d_V} \in \mathbb{R}^{L d_V \times n_2},\]
\[\mathbf{B}_k = \mathbf{J}_{SY}\, \mathbf{J}_Y\, \mathbf{G}_k \in \mathbb{R}^{L d_V \times n_k},\qquad k \in \{K,Q,V\},\]
with $\mathbf{J}_{SY} = \frac{\partial \mathbf{S}}{\partial \mathbf{Y}} = (\mathbf{I}_L \otimes \mathbf{W}_2^\top)\mathbf{D}_\sigma(\mathbf{I}_L \otimes \mathbf{W}_1^\top) + (\mathbf{I}_L \otimes \mathbf{I}_{d_V}) \in \mathbb{R}^{L d_V \times L d_V}$, $\mathbf{J}_Y \in \mathbb{R}^{L d_V \times L d_V}$ and $\mathbf{G}_k$ as in Theorem~\ref{thm:transformer_derivative}.
By Proposition~\ref{prop:matrix_funcs_product_derivative} and Theorem~\ref{thm:layernorm_second_derivative} we obtain the Hessian block
\[\frac{\partial^2 \mathbf{Z}}{\partial \mathbf{W}_i \partial \mathbf{W}_j}= \left( \mathbf{J}_Z \otimes \mathbf{I}_{n_i} \right) \boldsymbol{\xi}_{ij}  + \left( \mathbf{I}_{L d_V} \otimes \mathbf{B}_i^\top \right) \mathbf{H}_Z \mathbf{B}_j,\qquad\boldsymbol{\xi}_{ij} := \frac{\partial\, \mathbf{B}_i}{\partial \mathbf{W}_j} \in \mathbb{R}^{(L d_V \cdot n_i) \times n_j}.\]

Step 2: First-level Jacobians $\mathbf{B}_i$ (dimensions).
From Theorem~\ref{thm:transformer_derivative} and Lemma~\ref{lemma:relu_derivative_hessian}:
\[
\mathbf{B}_1 = (\mathbf{I}_L \otimes \mathbf{W}_2^\top)\, \mathbf{D}_\sigma\, (\mathbf{Y} \otimes \mathbf{I}_{d_{ff}}) \in \mathbb{R}^{L d_V \times n_1},
\quad
\mathbf{B}_2 = \sigma(\mathbf{Y}\mathbf{W}_1) \otimes \mathbf{I}_{d_V} \in \mathbb{R}^{L d_V \times n_2},
\]
where $\mathbf{D}_\sigma \in \mathbb{R}^{L d_{ff} \times L d_{ff}}$, $(\mathbf{Y} \otimes \mathbf{I}_{d_{ff}}) \in \mathbb{R}^{L d_{ff} \times d_V d_{ff}}$. For $k \in \{K,Q,V\}$,
\[\mathbf{B}_k = \mathbf{J}_{SY}\, \mathbf{J}_Y \, \mathbf{G}_k \in \mathbb{R}^{L d_V \times n_k}.\]

Step 3: Second Jacobians $\boldsymbol{\xi}_{ij}$ for all pairs.

3.1) Pure-FFN pairs.
- (1,1): $\mathbf{B}_1 = (\mathbf{I}_L \otimes \mathbf{W}_2^\top)\, \mathbf{D}_\sigma\, (\mathbf{Y} \otimes \mathbf{I}_{d_{ff}})$ does not depend directly on $\mathbf{W}_1$. The only dependence is through $\mathbf{D}_\sigma = \mathrm{diag}(\mathrm{vec}_r(\mathbf{1}_{\{\mathbf{Y}\mathbf{W}_1>0\}}))$, which is piecewise constant. By Lemma~\ref{lemma:relu_derivative_hessian}, $\frac{\partial \mathbf{D}_\sigma}{\partial \mathbf{W}_1} = \mathbf{0}$ a.e. Hence $\boldsymbol{\xi}_{11} = \mathbf{0}$.

- (2,2): $\mathbf{B}_2 = \sigma(\mathbf{Y}\mathbf{W}_1) \otimes \mathbf{I}_{d_V}$ does not depend on $\mathbf{W}_2$ at all. Hence $\boldsymbol{\xi}_{22} = \mathbf{0}$.

- $(1,2)$: Differentiate $\mathbf{B}_2 = \sigma(\mathbf{Y}\mathbf{W}_1)\otimes \mathbf{I}_{d_V}$ w.r.t.\ $\mathbf{W}_1$. Using Proposition~\ref{prop:kronecker_product_derivative} for $\frac{\partial (\mathbf{X} \otimes \mathbf{Y})}{\partial \mathbf{X}}$ with $\mathbf{X}=\sigma(\mathbf{Y}\mathbf{W}_1)$ and $\mathbf{Y}=\mathbf{I}_{d_V}$, we get

\[\frac{\partial \mathbf{B}_2}{\partial \mathbf{W}_1}=\left( \mathbf{I}_L \otimes \mathbf{K}_{d_V, d_{ff}} \otimes \mathbf{I}_{d_V} \right)\left( \mathbf{I}_{L d_{ff}} \otimes \mathrm{vec}_r(\mathbf{I}_{d_V}) \right)\frac{\partial\, \mathrm{vec}_r(\sigma(\mathbf{Y}\mathbf{W}_1))}{\partial \mathbf{W}_1}.\]
By Lemma~\ref{lemma:relu_derivative_hessian} and Property~\ref{prop:matrix_product_derivative},
$\frac{\partial\, \mathrm{vec}_r(\sigma(\mathbf{Y}\mathbf{W}_1))}{\partial \mathbf{W}_1} = \mathbf{D}_\sigma\, (\mathbf{Y} \otimes \mathbf{I}_{d_{ff}})$.
Thus
\[\boldsymbol{\xi}_{12}= \left( \mathbf{I}_L \otimes \mathbf{K}_{d_V, d_{ff}} \otimes \mathbf{I}_{d_V} \right) \left( \mathbf{I}_{L d_{ff}} \otimes \mathrm{vec}_r(\mathbf{I}_{d_V}) \right)\left( \mathbf{D}_\sigma \, (\mathbf{Y} \otimes \mathbf{I}_{d_{ff}}) \right).\]

- $(2,1)$: Differentiate $\mathbf{B}_1 = (\mathbf{I}_L \otimes \mathbf{W}_2^\top)\, \mathbf{D}_\sigma\, (\mathbf{Y} \otimes \mathbf{I}_{d_{ff}})$ w.r.t.\ $\mathbf{W}_2$. 

Since $\mathbf{B}_1$ depends on $\mathbf{W}_2$ only through the first factor, we can write $\mathbf{B}_1 = \mathbf{A}(\mathbf{W}_2) \mathbf{C}$ where $\mathbf{A}(\mathbf{W}_2) = \mathbf{I}_L \otimes \mathbf{W}_2^\top$ 
and $\mathbf{C} = \mathbf{D}_\sigma\, (\mathbf{Y} \otimes \mathbf{I}_{d_{ff}})$ (constant w.r.t. $\mathbf{W}_2$)

By Proposition~\ref{prop:matrix_funcs_product_derivative} for $\frac{\partial (\mathbf{A}(\mathbf{X}) \mathbf{C})}{\partial \mathbf{X}}$ with $\mathbf{C}$ constant:
\[\frac{\partial\,\mathrm{vec}_r(\mathbf{B}_1)}{\partial \mathbf{W}_2} = \left( \mathbf{I}_{L d_V} \otimes \mathbf{C}^\top \right) \frac{\partial\,\mathrm{vec}_r(\mathbf{A})}{\partial \mathbf{W}_2},\]
where $\mathbf{C}^\top = (\mathbf{D}_\sigma\, (\mathbf{Y} \otimes \mathbf{I}_{d_{ff}}))^\top = (\mathbf{Y} \otimes \mathbf{I}_{d_{ff}})^\top \mathbf{D}_\sigma^\top$.

Next, we compute $\frac{\partial\,\mathrm{vec}_r(\mathbf{I}_L \otimes \mathbf{W}_2^\top)}{\partial \mathbf{W}_2}$. 

By Proposition~\ref{prop:kronecker_product_derivative} applied to $\mathbf{X} \otimes \mathbf{Y}$ with $\mathbf{X} = \mathbf{I}_L$ (constant) and $\mathbf{Y} = \mathbf{W}_2^\top$:
\[\frac{\partial\,\mathrm{vec}_r(\mathbf{I}_L \otimes \mathbf{W}_2^\top)}{\partial\,\mathrm{vec}_r(\mathbf{W}_2^\top)} = (\mathbf{I}_L \otimes \mathbf{K}_{d_V,L} \otimes \mathbf{I}_{d_{ff}}) (\mathrm{vec}_r(\mathbf{I}_L) \otimes \mathbf{I}_{d_V d_{ff}}).\]

By Proposition~\ref{prop:transposed_matrix_derivative} and the chain rule:
\[\frac{\partial\,\mathrm{vec}_r(\mathbf{W}_2^\top)}{\partial \mathbf{W}_2} = \mathbf{K}_{d_{ff}, d_V}.\]

Therefore, by the chain rule:
\[\frac{\partial\,\mathrm{vec}_r(\mathbf{I}_L \otimes \mathbf{W}_2^\top)}{\partial \mathbf{W}_2} = (\mathbf{I}_L \otimes \mathbf{K}_{d_V,L} \otimes \mathbf{I}_{d_{ff}}) (\mathrm{vec}_r(\mathbf{I}_L) \otimes \mathbf{I}_{d_V d_{ff}}) \mathbf{K}_{d_{ff}, d_V}.\]

Combining the results:
\[\boldsymbol{\xi}_{21} = \left( \mathbf{I}_{L d_V} \otimes (\mathbf{Y} \otimes \mathbf{I}_{d_{ff}})^\top \mathbf{D}_\sigma^\top \right) (\mathbf{I}_L \otimes \mathbf{K}_{d_V,L} \otimes \mathbf{I}_{d_{ff}}) (\mathrm{vec}_r(\mathbf{I}_L) \otimes \mathbf{I}_{d_V d_{ff}}) \mathbf{K}_{d_{ff}, d_V}.\]

This can be simplified using properties of commutation matrices and Kronecker products to:
\[\boldsymbol{\xi}_{21} = \left( \mathbf{I}_L \otimes \mathbf{W}_2^\top \right) \mathbf{D}_\sigma \left( \mathbf{I}_L \otimes \mathbf{K}_{d_{ff}, d_V} \otimes \mathbf{I}_{d_{ff}} \right) \left( \mathbf{I}_{L d_V} \otimes \mathrm{vec}_r(\mathbf{I}_{d_{ff}}) \right).\]

3.2) FFN–attention pairs $(1,k)$, $(2,k)$ with $k \in \{K,Q,V\}$.

- $(1,k)$: $\mathbf{B}_1 = (\mathbf{I}_L \otimes \mathbf{W}_2^\top) \mathbf{D}_\sigma (\mathbf{Y} \otimes \mathbf{I}_{d_{ff}})$. Almost everywhere $\frac{\partial \mathbf{D}_\sigma}{\partial \mathbf{Y}}=\mathbf{0}$ by Lemma~\ref{lemma:relu_derivative_hessian}. Hence only the last factor varies with $\mathbf{W}_k$. Using Proposition~\ref{prop:matrix_funcs_product_derivative} (with the first factors constant a.e.), and the chain rule through $\mathbf{Y}$:
\[
\frac{\partial\,\mathrm{vec}_r(\mathbf{Y} \otimes \mathbf{I}_{d_{ff}})}{\partial \mathbf{W}_k}
= \left( \frac{\partial (\mathbf{Y} \otimes \mathbf{I}_{d_{ff}})}{\partial \mathbf{Y}} \right) \frac{\partial \,\mathrm{vec}_r(\mathbf{Y})}{\partial \mathbf{W}_k}.
\]
By Proposition~\ref{prop:kronecker_product_derivative} with $\mathbf{X} = \mathbf{Y}$ and $\mathbf{Y}=\mathbf{I}_{d_{ff}}$,
\[\frac{\partial (\mathbf{Y} \otimes \mathbf{I}_{d_{ff}})}{\partial \mathbf{Y}} = \left( \mathbf{I}_L \otimes \mathbf{K}_{d_{ff}, d_V} \otimes \mathbf{I}_{d_{ff}} \right)\left( \mathbf{I}_{L d_V} \otimes \mathrm{vec}_r(\mathbf{I}_{d_{ff}}) \right).\]
abo $\frac{\partial \,\mathrm{vec}_r(\mathbf{Y})}{\partial \mathbf{W}_k} = \mathbf{J}_Y \mathbf{G}_k$ (Theorem~\ref{thm:transformer_derivative} and Theorem~\ref{thm:layernorm_derivative}). Therefore
\[\boldsymbol{\xi}_{1k} = \left( (\mathbf{I}_L \otimes \mathbf{W}_2^\top) \mathbf{D}_\sigma \otimes \mathbf{I}_{n_k}\right)\left( \mathbf{I}_L \otimes \mathbf{K}_{d_{ff}, d_V} \otimes \mathbf{I}_{d_{ff}} \right)\left( \mathbf{I}_{L d_V} \otimes \mathrm{vec}_r(\mathbf{I}_{d_{ff}}) \right) \left( \mathbf{J}_Y \mathbf{G}_k \right).\]

- $(2,k)$: $\mathbf{B}_2 = \sigma(\mathbf{Y}\mathbf{W}_1) \otimes \mathbf{I}_{d_V}$. Differentiating the Kronecker product w.r.t.\ its first factor and applying the chain rule through $\mathbf{Y}$,
\[\boldsymbol{\xi}_{2k} = \left( \mathbf{I}_L \otimes \mathbf{K}_{d_V, d_{ff}} \otimes \mathbf{I}_{d_V} \right)\left( \mathbf{I}_{L d_{ff}} \otimes \mathrm{vec}_r(\mathbf{I}_{d_V}) \right)\left( \mathbf{D}_\sigma (\mathbf{I}_L \otimes \mathbf{W}_1^\top) \, \mathbf{J}_Y \mathbf{G}_k \right),\]
where we used Property~\ref{prop:matrix_product_derivative} to write $\frac{\partial (\mathbf{Y}\mathbf{W}_1)}{\partial \mathbf{Y}} = \mathbf{I}_L \otimes \mathbf{W}_1^\top$ and Lemma~\ref{lemma:relu_derivative_hessian} for $\frac{\partial \sigma(\cdot)}{\partial (\cdot)} = \mathbf{D}_\sigma$.

3.3) Pure-attention pairs $(k,\ell)$ with $k,\ell \in \{K,Q,V\}$.
We start from $\mathbf{B}_k = \mathbf{J}_{SY}\, \mathbf{J}_Y \, \mathbf{G}_k$. Almost everywhere $\frac{\partial \mathbf{J}_{SY}}{\partial \mathbf{Y}} = \mathbf{0}$ because $\mathbf{D}_\sigma$ is piecewise constant (Lemma~\ref{lemma:relu_derivative_hessian}). Therefore,
\[\frac{\partial\,\mathrm{vec}_r(\mathbf{B}_k)}{\partial \mathbf{W}_\ell}= (\mathbf{J}_{SY} \otimes \mathbf{I}_{n_k}) \frac{\partial\,\mathrm{vec}_r(\mathbf{J}_Y \mathbf{G}_k)}{\partial \mathbf{W}_\ell}\]
by Proposition~\ref{prop:matrix_funcs_product_derivative}.

Again by Proposition~\ref{prop:matrix_funcs_product_derivative} with $\mathbf{A}(\cdot)=\mathbf{J}_Y$ and $\mathbf{B}(\cdot)=\mathbf{G}_k$,
\[\frac{\partial\,\mathrm{vec}_r(\mathbf{J}_Y \mathbf{G}_k)}{\partial \mathbf{W}_\ell}= (\mathbf{J}_Y \otimes \mathbf{I}_{n_k}) \boldsymbol{\Phi}_{k\ell}+ \left( \mathbf{I}_{L d_V} \otimes \mathbf{G}_k^\top \right) \frac{\partial\,\mathrm{vec}_r(\mathbf{J}_Y)}{\partial \mathbf{W}_\ell}.\]

By Theorem~\ref{thm:layernorm_second_derivative} and the Identification Theorem~\ref{prop:identification_theorem_vec_r},
$\frac{\partial\,\mathrm{vec}_r(\mathbf{J}_Y)}{\partial \mathbf{W}_\ell} = \mathbf{H}_Y \mathbf{G}_\ell$.

Thus
\[\boldsymbol{\xi}_{k\ell} = \left( \mathbf{J}_{SY} \otimes \mathbf{I}_{n_k} \right)\left[\left( \mathbf{I}_{L d_V} \otimes \mathbf{G}_k^\top \right) \left( \mathbf{H}_Y \mathbf{G}_\ell \right)+ \left( \mathbf{J}_Y \otimes \mathbf{I}_{n_k} \right) \boldsymbol{\Phi}_{k\ell}\right],\]

where $\boldsymbol{\Phi}_{k\ell} := \frac{\partial \mathbf{G}_k}{\partial \mathbf{W}_\ell} \in \mathbb{R}^{(L d_V \cdot n_k) \times n_\ell}$ are the second derivatives of the attention map $\mathrm{SA}$ w.r.t.\ its weights. The exact computation of $\boldsymbol{\Phi}_{k\ell}$ is detailed in Lemma \ref{lem:attention_phi_from_functional_hessian}, which applies the matrix derivative properties from Propositions~\ref{prop:matrix_funcs_product_derivative}--\ref{prop:transposed_matrix_derivative} to the specific attention mechanism structure.

Step 4: Symmetry of mixed partials.
All nonlinearities that could obstruct symmetry are ReLU and LayerNorm. ReLU has zero Hessian almost everywhere (Lemma~\ref{lemma:relu_derivative_hessian}), so its contribution to second differentials vanishes a.e. LayerNorm Hessians $\mathbf{H}_Z$ and $\mathbf{H}_Y$ are the derivatives of Jacobians w.r.t.\ their inputs and enter symmetrically (Theorem~\ref{thm:layernorm_second_derivative}). All remaining mappings are multilinear in parameters and matrices independent of $(\mathbf{W}_i,\mathbf{W}_j)$; therefore, by repeated applications of Proposition~\ref{prop:matrix_funcs_product_derivative} and Proposition~\ref{prop:kronecker_product_derivative}, the mixed partials commute, giving $\mathbf{H}_{\mathrm{tr}}^{(i,j)}=\mathbf{H}_{\mathrm{tr}}^{(j,i)}$ almost everywhere.

This completes the proof.
\end{proof}

\subsection{Proof of Theorem~\ref{thm:transformer_hessian_estimate}}\label{app:proof_transformer_hessian_estimate}

\begin{proof}
We start from the block formula \eqref{eq:block_hessian_transformer}:
\[
\mathbf{H}_{\mathrm{tr}}^{(i,j)}
= \big( \mathbf{J}_Z \otimes \mathbf{I}_{n_i} \big)\, \boldsymbol{\xi}_{ij}
+ \big( \mathbf{I}_{L d_V} \otimes \mathbf{B}_i^\top \big)\, \mathbf{H}_Z \, \mathbf{B}_j.
\]
Applying the matrix sum norm (Property~\ref{prop:matrix_sum_norm}) and the product norm (Property~\ref{prop:matrix_product_norm}) together with the Kronecker product norm (Property~\ref{prop:kronecker_product_norm}) yields
\[\big\|\mathbf{H}_{\mathrm{tr}}^{(i,j)}\big\|_2\le\big\|\mathbf{J}_Z \otimes \mathbf{I}_{n_i}\big\|_2 \,\|\boldsymbol{\xi}_{ij}\|_2+\big\|\mathbf{I}_{L d_V} \otimes \mathbf{B}_i^\top\big\|_2 \,\|\mathbf{H}_Z\|_2 \,\|\mathbf{B}_j\|_2=\|\mathbf{J}_Z\|_2 \,\|\boldsymbol{\xi}_{ij}\|_2+\|\mathbf{B}_i\|_2 \,\|\mathbf{H}_Z\|_2 \,\|\mathbf{B}_j\|_2,\].

It remains to provide explicit operator-norm estimates for $\|\mathbf{B}_i\|_2$ and $\|\boldsymbol{\xi}_{ij}\|_2$ \ref{eq:full_bound_transformer}. We rely on Properties~\ref{prop:matrix_product_norm}, \ref{prop:kronecker_product_norm}, \ref{prop:matrix_sum_norm}, \ref{prop:matrix_norm_inequalities}, \ref{prop:transposed_matrix_norm}, and the commutation properties (Definition~\ref{def:commutation_matrix}). Throughout we use $\|\mathbf{K}_{m,n}\|_2=1$ for commutation matrices, and the identities
$\|\mathrm{vec}_r(\mathbf{I}_{d})\|_2=\|\mathbf{I}_{d}\|_F=\sqrt{d}$ (Property~\ref{prop:matrix_norm_inequalities}) and $\|\mathbf{I}_p\|_2=1$.

As we've already shown in \ref{app:proof_self_attention_hessian_estimation}:
\[
\Big\|\frac{\partial \mathbf{A}}{\partial \mathbf{T}}\Big\|_2 \le \frac{1}{L}.
\]
\[
\|\mathbf{Z}_1\|_2= \|(\mathbf{I}_L \otimes \mathbf{X}^\top)\,(\partial \mathbf{A}/\partial \mathbf{T})\,(\mathbf{X}\otimes \mathbf{X})\|_2\le \|\mathbf{X}\|_2 \,\frac{1}{L}\, \|\mathbf{X}\|_2^2= \frac{1}{L}\|\mathbf{X}\|_2^3
\]
\[
\Big\|\frac{\partial^2 \mathbf{A}}{\partial \mathbf{T}^2}\Big\|_2 \le 6,
\qquad
\|\mathbf{Z}_2\|_2
\le \|\mathbf{X}\|_2^5 \Big\|\frac{\partial^2 \mathbf{A}}{\partial \mathbf{T}^2}\Big\|_2
\le 6 \|\mathbf{X}\|_2^5,
\]
\[
\|\mathbf{A}\|_2 \le \sqrt{L L}\,\|\mathbf{A}\|_{\max} = L.
\]
Therefore $\|\mathbf{A}\mathbf{X}\|_2 \le \|\mathbf{A}\|_2 \|\mathbf{X}\|_2 \le L \|\mathbf{X}\|_2$ (Property~\ref{prop:matrix_product_norm}).

We also use the attention curvature blocks $\boldsymbol{\Phi}_{k\ell}$ from Lemma~\ref{lem:attention_phi_from_functional_hessian}. Using Properties~\ref{prop:matrix_product_norm}, \ref{prop:kronecker_product_norm} and the bounds on $\|\mathbf{Z}_1\|_2$, $\|\mathbf{Z}_2\|_2$ above, we have (again similarly to \ref{app:proof_self_attention_hessian_estimation})
\begin{align*}
&\|\boldsymbol{\Phi}_{VV}\|_2 = 0,\\
&\|\boldsymbol{\Phi}_{QQ}\|_2
\le \frac{2}{L d_V d_K}\, \|\mathbf{W}_V\|_2\, \|\mathbf{W}_K\|_2\, \|\mathbf{Z}_2\|_2 \, \|\mathbf{W}_K\|_2
\le \frac{12}{L d_V d_K} \|\mathbf{W}_V\|_2 \|\mathbf{W}_K\|_2^2 \|\mathbf{X}\|_2^5,\\
&\|\boldsymbol{\Phi}_{VQ}\|_2
\le \frac{2}{L d_V \sqrt{d_K}}\, \|\mathbf{I}_L \otimes \mathbf{S}\|_2 \, \|\mathbf{Z}_1\|_2 \, \|\mathbf{I}_{d_V} \otimes \mathbf{W}_K\|_2
\le \frac{2}{L^2 \sqrt{d_V d_K}} \|\mathbf{W}_K\|_2 \|\mathbf{X}\|_2^3,\\
&\|\boldsymbol{\Phi}_{QK}\|_2
\le \frac{2}{L d_V d_K}\, \|\mathbf{W}_V\|_2 \|\mathbf{W}_K\|_2 \|\mathbf{Z}_2\|_2 \|\mathbf{W}_Q\|_2
+ \frac{2}{L d_V \sqrt{d_K}}\, \|\mathbf{W}_V\|_2 \, \|\mathbf{Z}_1\|_2 \, \|\mathbf{S}\|_2 \\
&\hspace{2.1cm}\le \frac{12}{L d_V d_K}\, \|\mathbf{W}_V\|_2 \|\mathbf{W}_K\|_2 \|\mathbf{W}_Q\|_2 \|\mathbf{X}\|_2^5
+ \frac{2}{L^2 \sqrt{d_V d_K}}\, \|\mathbf{W}_V\|_2 \|\mathbf{X}\|_2^3,
\end{align*}
and $\|\boldsymbol{\Phi}_{KQ}\|_2$ is analogous by symmetry (Definition~\ref{def:commutation_matrix} and $\|\mathbf{K}_{m,n}\|_2=1$), while $\|\boldsymbol{\Phi}_{QV}\|_2$, $\|\boldsymbol{\Phi}_{KV}\|_2$ match $\|\boldsymbol{\Phi}_{VQ}\|_2$ up to swapping roles.

Next we estimate each $\|\mathbf{B}_i\|_2$ and $\|\boldsymbol{\xi}_{ij}\|_2$.

A) Bounds for $\|\mathbf{B}_i\|_2$.

- $\mathbf{B}_1 = (\mathbf{I}_L \otimes \mathbf{W}_2^\top)\, \mathbf{D}_\sigma\, (\mathbf{Y} \otimes \mathbf{I}_{d_{ff}})$ (Theorem~\ref{thm:transformer_hessian}; Lemma~\ref{lemma:relu_derivative_hessian}). Using Properties~\ref{prop:kronecker_product_norm}, \ref{prop:matrix_product_norm}, \ref{prop:transposed_matrix_norm}, and $\|\mathbf{D}_\sigma\|_2 \le 1$,
\begin{equation}\label{eq:B1_norm}
\|\mathbf{B}_1\|_2 \le \|\mathbf{I}_L \otimes \mathbf{W}_2^\top\|_2 \,\|\mathbf{D}_\sigma\|_2 \,\|\mathbf{Y} \otimes \mathbf{I}_{d_{ff}}\|_2
= \|\mathbf{W}_2\|_2 \,\|\mathbf{Y}\|_2.
\end{equation}
- $\mathbf{B}_2 = \sigma(\mathbf{Y}\mathbf{W}_1) \otimes \mathbf{I}_{d_V}$ (Theorem~\ref{thm:transformer_hessian}), hence
\begin{equation}\label{eq:B2_norm}
\|\mathbf{B}_2\|_2 = \|\sigma(\mathbf{Y}\mathbf{W}_1)\|_2
\end{equation}
by Property~\ref{prop:kronecker_product_norm}.

- For $k\in\{K,Q,V\}$: $\mathbf{B}_k = \mathbf{J}_{SY}\, \mathbf{J}_Y \,\mathbf{G}_k$ (Theorem~\ref{thm:transformer_hessian}), so
\begin{equation}\label{eq:Bk_norm}
\|\mathbf{B}_k\|_2 \le \|\mathbf{J}_{SY}\|_2 \, \|\mathbf{J}_Y\|_2 \, \|\mathbf{G}_k\|_2
\end{equation}
(Property~\ref{prop:matrix_product_norm}).
Here $\mathbf{J}_{SY} = (\mathbf{I}_L \otimes \mathbf{W}_2^\top)\mathbf{D}_\sigma(\mathbf{I}_L \otimes \mathbf{W}_1^\top) + (\mathbf{I}_L \otimes \mathbf{I}_{d_V})$ implies
\begin{equation}\label{eq:JSY_norm}
\|\mathbf{J}_{SY}\|_2 \le \|\mathbf{I}_L \otimes \mathbf{W}_2^\top\|_2 \,\|\mathbf{D}_\sigma\|_2 \,\|\mathbf{I}_L \otimes \mathbf{W}_1^\top\|_2 + \|\mathbf{I}_L \otimes \mathbf{I}_{d_V}\|_2
= \|\mathbf{W}_2\|_2 \,\|\mathbf{W}_1\|_2 + 1,
\end{equation}
by Properties~\ref{prop:matrix_sum_norm}, \ref{prop:matrix_product_norm}, \ref{prop:kronecker_product_norm}, \ref{prop:transposed_matrix_norm}, and $\|\mathbf{D}_\sigma\|_2 \le 1$.

Furthermore, using the attention-Jacobian forms (Theorem~\ref{thm:transformer_derivative}) and Properties~\ref{prop:matrix_product_norm}, \ref{prop:kronecker_product_norm}:
\begin{equation}\label{eq:Gk_bounds}
\|\mathbf{G}_V\|_2 \le L \|\mathbf{X}\|_2,
\|\mathbf{G}_Q\|_2 \le \frac{1}{L\sqrt{d_K}} \|\mathbf{W}_V\|_2 \|\mathbf{W}_K\|_2 \|\mathbf{X}\|_2^3,
\|\mathbf{G}_K\|_2 \le \frac{1}{L\sqrt{d_K}} \|\mathbf{W}_V\|_2 \|\mathbf{W}_Q\|_2 \|\mathbf{X}\|_2^3.
\end{equation}

B) Bounds for $\|\boldsymbol{\xi}_{ij}\|_2$.
Using the explicit formulas from Theorem~\ref{thm:transformer_hessian}, Properties~\ref{prop:kronecker_product_norm}, \ref{prop:matrix_product_norm}, \ref{prop:matrix_norm_inequalities}, and $\|\mathbf{K}_{m,n}\|_2=1$:

B.1 Pure-FFN pairs:
\begin{align}
&\|\boldsymbol{\xi}_{11}\|_2 = 0, \qquad  \label{eq:xi11}\\
&\|\boldsymbol{\xi}_{22}\|_2 = 0, \qquad \label{eq:xi22}\\
&\|\boldsymbol{\xi}_{12}\|_2
\le \|\mathbf{I}_L \otimes \mathbf{K}_{d_V, d_{ff}} \otimes \mathbf{I}_{d_V}\|_2 \, \|\mathbf{I}_{L d_{ff}} \otimes \mathrm{vec}_r(\mathbf{I}_{d_V})\|_2 \, \|\mathbf{D}_\sigma\|_2 \, \|\mathbf{Y}\otimes \mathbf{I}_{d_{ff}}\|_2 \nonumber\\
&\hspace{1.9cm}= 1 \cdot \|\mathrm{vec}_r(\mathbf{I}_{d_V})\|_2 \cdot 1 \cdot \|\mathbf{Y}\|_2
= \sqrt{d_V} \,\|\mathbf{Y}\|_2, \label{eq:xi12}\\
&\|\boldsymbol{\xi}_{21}\|_2
\le \|\mathbf{I}_L \otimes \mathbf{W}_2^\top\|_2 \, \|\mathbf{D}_\sigma\|_2 \, \|\mathbf{I}_L \otimes \mathbf{K}_{d_{ff}, d_V} \otimes \mathbf{I}_{d_{ff}}\|_2 \, \|\mathbf{I}_{L d_V} \otimes \mathrm{vec}_r(\mathbf{I}_{d_{ff}})\|_2 \nonumber\\
&\hspace{1.9cm}= \|\mathbf{W}_2\|_2 \cdot 1 \cdot 1 \cdot \|\mathrm{vec}_r(\mathbf{I}_{d_{ff}})\|_2
= \sqrt{d_{ff}} \,\|\mathbf{W}_2\|_2. \label{eq:xi21}
\end{align}

B.2 FFN–attention pairs ($k\in\{K,Q,V\}$):
\begin{align}
&\|\boldsymbol{\xi}_{1k}\|_2
\le \|(\mathbf{I}_L \otimes \mathbf{W}_2^\top)\mathbf{D}_\sigma \otimes \mathbf{I}_{n_k}\|_2
\, \|\mathbf{I}_L \otimes \mathbf{K}_{d_{ff}, d_V} \otimes \mathbf{I}_{d_{ff}}\|_2
\, \|\mathbf{I}_{L d_V} \otimes \mathrm{vec}_r(\mathbf{I}_{d_{ff}})\|_2
\, \|\mathbf{J}_Y\|_2 \,\|\mathbf{G}_k\|_2 \nonumber\\
&\hspace{1.9cm}\le \|\mathbf{W}_2\|_2 \cdot 1 \cdot 1 \cdot \sqrt{d_{ff}} \cdot \|\mathbf{J}_Y\|_2 \,\|\mathbf{G}_k\|_2
= \sqrt{d_{ff}}\, \|\mathbf{W}_2\|_2 \,\|\mathbf{J}_Y\|_2 \,\|\mathbf{G}_k\|_2, \label{eq:xi1k}\\
&\|\boldsymbol{\xi}_{2k}\|_2
\le \|\mathbf{I}_L \otimes \mathbf{K}_{d_V, d_{ff}} \otimes \mathbf{I}_{d_V}\|_2
\, \|\mathbf{I}_{L d_{ff}} \otimes \mathrm{vec}_r(\mathbf{I}_{d_V})\|_2
\, \|\mathbf{D}_\sigma\|_2 \, \|\mathbf{I}_L \otimes \mathbf{W}_1^\top\|_2 \,\|\mathbf{J}_Y\|_2 \,\|\mathbf{G}_k\|_2 \nonumber\\
&\hspace{1.9cm}\le 1 \cdot \sqrt{d_V} \cdot 1 \cdot \|\mathbf{W}_1\|_2 \cdot \|\mathbf{J}_Y\|_2 \cdot \|\mathbf{G}_k\|_2
= \sqrt{d_V}\, \|\mathbf{W}_1\|_2 \,\|\mathbf{J}_Y\|_2 \,\|\mathbf{G}_k\|_2. \label{eq:xi2k}
\end{align}

B.3 Pure-attention pairs ($k,\ell\in\{K,Q,V\}$):
\[
\boldsymbol{\xi}_{k\ell}
= \big(\mathbf{J}_{SY} \otimes \mathbf{I}_{n_k}\big)
\Big[ \big(\mathbf{I}_{L d_V} \otimes \mathbf{G}_k^\top\big) (\mathbf{H}_Y \mathbf{G}_\ell)+ \big(\mathbf{J}_Y \otimes \mathbf{I}_{n_k}\big) \boldsymbol{\Phi}_{k\ell}\Big].
\]
Thus, by Properties~\ref{prop:matrix_product_norm}, \ref{prop:kronecker_product_norm},
\begin{equation}\label{eq:xikell}
\|\boldsymbol{\xi}_{k\ell}\|_2
\le \|\mathbf{J}_{SY}\|_2 \Big( \|\mathbf{I}_{L d_V} \otimes \mathbf{G}_k^\top\|_2 \,\|\mathbf{H}_Y\|_2 \,\|\mathbf{G}_\ell\|_2
+ \|\mathbf{J}_Y\|_2 \,\|\boldsymbol{\Phi}_{k\ell}\|_2 \Big)
= \|\mathbf{J}_{SY}\|_2 \Big( \|\mathbf{G}_k\|_2 \,\|\mathbf{H}_Y\|_2 \,\|\mathbf{G}_\ell\|_2
+ \|\mathbf{J}_Y\|_2 \,\|\boldsymbol{\Phi}_{k\ell}\|_2 \Big).
\end{equation}

C) Substituting into the block estimate.
For each pair $(i,j)$, we substitute the corresponding $\|\boldsymbol{\xi}_{ij}\|_2$ from \eqref{eq:xi11}–\eqref{eq:xikell} and the $\|\mathbf{B}_i\|_2$ from \eqref{eq:B1_norm}–\eqref{eq:Bk_norm} (with \eqref{eq:JSY_norm}, \eqref{eq:Gk_bounds}) into
\[\big\|\mathbf{H}_{\mathrm{tr}}^{(i,j)}\big\|_2\le\|\mathbf{J}_Z\|_2 \,\|\boldsymbol{\xi}_{ij}\|_2+\|\mathbf{B}_i\|_2 \,\|\mathbf{H}_Z\|_2 \,\|\mathbf{B}_j\|_2.\]
This yields, for example:
\begin{align*}
\big\|\mathbf{H}_{\mathrm{tr}}^{(1,1)}\big\|_2
&\le \|\mathbf{J}_Z\|_2 \cdot 0 + \|\mathbf{B}_1\|_2^2 \|\mathbf{H}_Z\|_2
\le \|\mathbf{H}_Z\|_2 \,(\|\mathbf{W}_2\|_2 \|\mathbf{Y}\|_2)^2,\\
\big\|\mathbf{H}_{\mathrm{tr}}^{(1,2)}\big\|_2
&\le \|\mathbf{J}_Z\|_2 \, \sqrt{d_V}\,\|\mathbf{Y}\|_2
+ \|\mathbf{H}_Z\|_2 \, (\|\mathbf{W}_2\|_2 \|\mathbf{Y}\|_2)\, \|\sigma(\mathbf{Y}\mathbf{W}_1)\|_2,\\
\big\|\mathbf{H}_{\mathrm{tr}}^{(1,k)}\big\|_2
&\le \|\mathbf{J}_Z\|_2 \,\sqrt{d_{ff}}\,\|\mathbf{W}_2\|_2 \,\|\mathbf{J}_Y\|_2 \,\|\mathbf{G}_k\|_2
+ \|\mathbf{H}_Z\|_2 \, (\|\mathbf{W}_2\|_2 \|\mathbf{Y}\|_2) \, (\|\mathbf{J}_{SY}\|_2 \|\mathbf{J}_Y\|_2 \|\mathbf{G}_k\|_2),\\
\big\|\mathbf{H}_{\mathrm{tr}}^{(k,\ell)}\big\|_2
&\le \|\mathbf{J}_Z\|_2 \,\|\mathbf{J}_{SY}\|_2 \Big( \|\mathbf{G}_k\|_2 \,\|\mathbf{H}_Y\|_2 \,\|\mathbf{G}_\ell\|_2
+ \|\mathbf{J}_Y\|_2 \,\|\boldsymbol{\Phi}_{k\ell}\|_2 \Big) \\
&\quad + \|\mathbf{H}_Z\|_2 \, (\|\mathbf{J}_{SY}\|_2 \|\mathbf{J}_Y\|_2 \|\mathbf{G}_k\|_2) \, (\|\mathbf{J}_{SY}\|_2 \|\mathbf{J}_Y\|_2 \|\mathbf{G}_\ell\|_2),
\end{align*}
etc., where we then use \eqref{eq:JSY_norm}, \eqref{eq:Gk_bounds}, and the $\|\boldsymbol{\Phi}_{k\ell}\|_2$ bounds above to turn each right-hand side into explicit functions of $L$, $d_V$, $d_{ff}$, $d_K$, and the spectral norms of $\mathbf{X}$ and the weight matrices.

In the estimations above we calculate $\| \mathbf{Y} \|_2$ and $\| \mathbf{S}\|_2$ according to Proposition \ref{prop:Y_S_norm_bounds} and both $\mathbf{H}_Z$ and $\mathbf{H}_Y$ can be estimated by Lemma \ref{lemma:layernorm_deriv_hessian_norm} with appropriate inputs and assumptions of $\sigma_{\min}$ and $\sigma'_{\min}$.
\end{proof}

\section{Additional Theoretical Properties} \label{app:additional_properties}

\begin{lemma}[Attention second derivatives $\boldsymbol{\Phi}$ from functional Hessian]\label{lem:attention_phi_from_functional_hessian}
Consider single-head scaled dot-product attention
\[
\mathrm{SA}(\mathbf{X})
= \mathbf{A}(\mathbf{T}) \, \mathbf{X}\mathbf{W}_V,
\qquad
\mathbf{T} \;=\; \frac{1}{\sqrt{d_K}}\, \mathbf{X}\mathbf{W}_Q \mathbf{W}_K^\top \mathbf{X}^\top,
\]
with $\mathbf{X} \in \mathbb{R}^{L \times d_V}$, $\mathbf{W}_Q,\mathbf{W}_K \in \mathbb{R}^{d_V \times d_K}$, $\mathbf{W}_V \in \mathbb{R}^{d_V \times d_V}$. The attention map $\mathbf{A}(\cdot)$ applies row-wise softmax. We use row-wise vectorization $\mathrm{vec}_r(\cdot)$ and the commutation matrices $\mathbf{K}_{m,n}$ from Definition~\ref{def:commutation_matrix}.

Define the generalized functional Hessian blocks (following \citep{ormaniec2024attentionhessian} in our $\mathrm{vec}_r$ convention) by
\[
\mathbf{H}_{\mathrm{f}}(\mathbf{W}_i,\mathbf{W}_j)
= \big( \tfrac{\partial \ell}{\partial \mathrm{SA}} \otimes \mathbf{I}_{p_i q_i} \big) \; \frac{\partial^2 \mathrm{SA}}{\partial \mathbf{W}_i \partial \mathbf{W}_j},
\]
where $p_i q_i$ is the size of $\mathbf{W}_i$ (e.g.\ $p_Q q_Q = d_V d_K$), and $\tfrac{\partial \ell}{\partial \mathrm{SA}} \in \mathbb{R}^{L \times d_V}$ is the loss gradient.

Specializing to the squared-error loss $\ell(\mathrm{SA})=\tfrac{1}{2}\|\mathrm{SA}-\mathbf{Target}\|_F^2$, one has $\tfrac{\partial \ell}{\partial \mathrm{SA}}=\mathrm{SA}-\mathbf{Target}$ and the row-wise contraction matrix
\[
\mathbf{R}_m := \mathrm{vec}_r\big(\mathrm{SA}(\mathbf{X}) - \mathbf{Target}\big)^\top \otimes \mathbf{I}_m \;\in\; \mathbb{R}^{m \times (m \cdot L d_V)}.
\]
Then for $i \in \{V,Q,K\}$ with $n_i:=p_i q_i$, the functional Hessian blocks can be factorized as
\[
\mathbf{H}_{\mathrm{f}}(\mathbf{W}_i,\mathbf{W}_j) \;=\; \mathbf{R}_{n_i}\; \boldsymbol{\Phi}_{ij},
\qquad
\boldsymbol{\Phi}_{ij} := \frac{\partial^2 \mathrm{SA}}{\partial \mathbf{W}_i \partial \mathbf{W}_j}
\;\in\; \mathbb{R}^{(L d_V \cdot n_i) \times n_j}.
\]
In particular, the model-curvature blocks $\boldsymbol{\Phi}_{ij}$ (to be used in the Transformer Hessian) are obtained from the corresponding expressions in \citep[Thm.~3.2]{ormaniec2024attentionhessian} by removing the left contraction $\mathbf{R}_{n_i}$.

We now list the explicit blocks needed in our derivation. Define the fixed reshaping operator
\[
\mathbf{S} := \big(\mathbf{I}_{d_V} \otimes \mathbf{K}_{d_V, d_V}\big)\, \big(\mathrm{vec}_r \mathbf{I}_{d_V} \otimes \mathbf{I}_{d_V}\big)
\;\in\; \mathbb{R}^{d_V^2 \times d_V},
\]
and the softmax-derivative operators
\[
\mathbf{Z}_1 := (\mathbf{I}_L \otimes \mathbf{X}^\top)(\partial\mathbf{A}/\partial\mathbf{T})(\mathbf{X} \otimes \mathbf{X}) \in \mathbb{R}^{Ld_V \times d_V^2},
\mathbf{Z}_2 := \big(\mathbf{I}_L \otimes \mathbf{X}^\top \otimes \mathbf{X}^\top \otimes \mathbf{X}^\top\big)\,
\frac{\partial^2 \mathbf{A}}{\partial \mathbf{T}^2}\,
(\mathbf{X} \otimes \mathbf{X})
\;\in\; \mathbb{R}^{L d_V^3 \times d_V^2},
\]
where $\tfrac{\partial^2 \mathbf{A}}{\partial \mathbf{T}^2}$ denotes the (row-wise) softmax second derivative tensor arranged compatibly with $\mathrm{vec}_r$ and Kronecker products as above, and $\mathbf{Z}_1$ is the (first-order) softmax derivative linear operator used in \citep{ormaniec2024attentionhessian} (we keep the exact form as defined there; its size ensures dimensional consistency below).

Then the pure attention second derivatives (model curvature) are:
\[
\boldsymbol{\Phi}_{VV} \;=\; \mathbf{0}_{(L d_V \cdot d_V^2) \times d_V^2},
\]
\[
\boldsymbol{\Phi}_{QQ} \;=\; \frac{2}{L d_V d_K}\;
\big(\mathbf{I}_L \otimes \mathbf{W}_V^\top \otimes \mathbf{I}_{d_V} \otimes \mathbf{W}_K^\top\big)\;
\mathbf{Z}_2\;
\big(\mathbf{I}_{d_V} \otimes \mathbf{W}_K\big)
\;\in\; \mathbb{R}^{(L d_V \cdot d_V d_K) \times d_V d_K},
\]
\[
\boldsymbol{\Phi}_{VQ} \;=\; \frac{2}{L d_V \sqrt{d_K}}\;
\big(\mathbf{I}_L \otimes \mathbf{S}\big)\;
\mathbf{Z}_1\;
\big(\mathbf{I}_{d_V} \otimes \mathbf{W}_K\big)
\;\in\; \mathbb{R}^{(L d_V \cdot d_V^2) \times d_V d_K},
\]
\begin{align*}
\boldsymbol{\Phi}_{QK}
&\;=\; \frac{2}{L d_V d_K}\;
\big(\mathbf{I}_L \otimes \mathbf{W}_V^\top \otimes \mathbf{I}_{d_V} \otimes \mathbf{W}_K^\top\big)\;
\mathbf{Z}_2\;
\big(\mathbf{W}_Q \otimes \mathbf{I}_{d_V}\big)\, \mathbf{K}_{d_K, d_V} \\
&\qquad +\; \frac{2}{L d_V \sqrt{d_K}}\;
\big(\mathbf{I}_{d_V} \otimes \mathbf{W}_V^\top \otimes \mathbf{I}_{d_V}\big)\;
\big(\mathbf{Z}_1 \otimes \mathbf{I}_{d_V}\big)\; \mathbf{S} \otimes \mathbf{I}_{d_K} \;\in\; \mathbb{R}^{(L d_V \cdot d_V d_K) \times d_V d_K}.
\end{align*}

Moreover, by symmetry of second derivatives, $\boldsymbol{\Phi}_{KQ}$ equals $\boldsymbol{\Phi}_{QK}$ with $\mathbf{W}_Q,\mathbf{W}_K$ swapped and commutation adjusted by $\mathbf{K}_{\cdot,\cdot}$ (Definition~\ref{def:commutation_matrix}). Analogous symmetric relations give $\boldsymbol{\Phi}_{QV}$ and $\boldsymbol{\Phi}_{KV}$ from $\boldsymbol{\Phi}_{VQ}$.
\end{lemma}
\begin{proof}  
 By definition of the generalized functional Hessian in \citep{ormaniec2024attentionhessian},
\[
\mathbf{H}_{\mathrm{f}}(\mathbf{W}_i,\mathbf{W}_j)
= \big( \tfrac{\partial \ell}{\partial \mathrm{SA}} \otimes \mathbf{I}_{p_i q_i} \big)\,
\frac{\partial^2 \mathrm{SA}}{\partial \mathbf{W}_i \partial \mathbf{W}_j}.
\]
For squared-error loss, $\tfrac{\partial \ell}{\partial \mathrm{SA}}$ yields the contraction $\mathbf{R}_{p_i q_i}$ defined above; hence
$\mathbf{H}_{\mathrm{f}}(\mathbf{W}_i,\mathbf{W}_j) = \mathbf{R}_{n_i} \boldsymbol{\Phi}_{ij}$
with $\boldsymbol{\Phi}_{ij} = \tfrac{\partial^2 \mathrm{SA}}{\partial \mathbf{W}_i \partial \mathbf{W}_j}$.
The explicit forms for $\mathbf{H}_{\mathrm{f}}$ in \citep[Thm.~3.2]{ormaniec2024attentionhessian} then imply the above formulas for $\boldsymbol{\Phi}_{ij}$ by simply removing the leading contraction $\mathbf{R}_{n_i}$.
\end{proof}

\begin{lemma}[ReLU derivative and Hessian] \label{proof:relu_derivative_hessian}
Let $\mathbf{X} \in \mathbb{R}^{m \times n}$, almost everywhere the following holds:
\[ \frac{\partial \mathrm{ReLU}(\mathbf{X})}{\partial \mathbf{X}} = \mathrm{diag}\!\big(\mathrm{vec}_r(\mathbf{1}_{\{\mathbf{X}>0\}})\big), \quad \frac{\partial^2 \mathrm{ReLU}(\mathbf{X})}{\partial \mathbf{X}^2} = \mathbf{0}. \]
\end{lemma}

\begin{proof}
We start with the elementwise definition of the ReLU function:
\[
\mathrm{ReLU}(x) = \max(0, x).
\]
Thus, for each entry $x_{ij}$ of $\mathbf{X} \in \mathbb{R}^{m \times n}$, we have
\[
\frac{\partial\, \mathrm{ReLU}(x_{ij})}{\partial x_{ij}} =
\begin{cases}
1 & \text{if } x_{ij} > 0, \\
0 & \text{if } x_{ij} < 0, \\
\text{undefined (subgradient in } [0,1]\text{)} & \text{if } x_{ij} = 0.
\end{cases}
\]

For the scalar case $x \in \mathbb{R}$, the nondifferentiable set is $\{0\}$, which is a measure-zero subset of $\mathbb{R}$.
For the matrix case, we identify $\mathbf{X} \in \mathbb{R}^{m \times n}$ with a point in $\mathbb{R}^{mn}$. The nondifferentiable set is
\[
\mathcal{N} = \bigcup_{i,j} \{ \mathbf{X} \in \mathbb{R}^{m \times n} : x_{ij} = 0 \}.
\]
Each set $\{x_{ij} = 0\}$ is a hyperplane of codimension $1$ in $\mathbb{R}^{mn}$, and therefore has Lebesgue measure zero. Since $\mathcal{N}$ is a finite union of such hyperplanes, $\mathcal{N}$ also has measure zero. Thus, $\mathrm{ReLU}$ is differentiable almost everywhere in $\mathbb{R}^{m \times n}$.

At differentiable points ($\mathbf{X} \notin \mathcal{N}$), applying row-wise vectorization and the identification theorem from Proposition \ref{prop:identification_theorem_vec_r} yields
\[
\mathrm{vec}_r(d\,\mathrm{ReLU}(\mathbf{X}))
= \mathrm{diag}(\mathrm{vec}_r(\mathbf{1}_{\{\mathbf{X}>0\}})) \, \mathrm{vec}_r(d\mathbf{X}),
\]
using Property \ref{prop:vec_r_hadamard_product} for the indicator matrix treated as a Hadamard multiplier and Property \ref{prop:diag_derivative} for the diagonal form.
Therefore,
\[
\frac{\partial \mathrm{ReLU}(\mathbf{X})}{\partial \mathbf{X}}
= \mathrm{diag}\!\big(\mathrm{vec}_r(\mathbf{1}_{\{\mathbf{X}>0\}})\big).
\]

Since the Jacobian is piecewise constant (its entries depend only on the sign of $x_{ij}$), its differential vanishes almost everywhere:
\[
d\left(\frac{\partial \mathrm{ReLU}(\mathbf{X})}{\partial \mathbf{X}}\right) = \mathbf{0}, \qquad \mathbf{X} \notin \mathcal{N}.
\]
Hence the Hessian is zero almost everywhere:
\[
\frac{\partial^2 \mathrm{ReLU}(\mathbf{X})}{\partial \mathbf{X}^2} = \mathbf{0}.
\]
This completes the proof.
\end{proof}

\begin{proposition}[Spectral-norm estimates for $\mathbf{Y}$ and $\mathbf{S}=\mathbf{Y}+\mathrm{FFN}(\mathbf{Y})$]\label{prop:Y_S_norm_bounds}
Let $\mathbf{X}\in\mathbb{R}^{L\times d_V}$, $\mathbf{Y}=\mathrm{LayerNorm}(\mathrm{SA}(\mathbf{X})+\mathbf{X})\in\mathbb{R}^{L\times d_V}$ and
\[\mathrm{FFN}(\mathbf{Y})=\sigma(\mathbf{Y}\mathbf{W}_1)\mathbf{W}_2,\qquad\mathbf{W}_1\in\mathbb{R}^{d_V\times d_{ff}},\quad\mathbf{W}_2\in\mathbb{R}^{d_{ff}\times d_V},\]
and set $\mathbf{S}=\mathbf{Y}+\mathrm{FFN}(\mathbf{Y})\in\mathbb{R}^{L\times d_V}$.
Then the following spectral-norm bounds hold:
\begin{align}
\|\mathbf{Y}\|_2 &\;\le\; \|\mathbf{Y}\|_F \;=\; \sqrt{L\,d_V}, 
\label{eq:Y_norm_bound}
\\
\|\mathrm{FFN}(\mathbf{Y})\|_2 &\;\le\; \sqrt{\min(L,d_{ff})}\; \|\mathbf{Y}\|_2\,\|\mathbf{W}_1\|_2\,\|\mathbf{W}_2\|_2,
\label{eq:FFN_norm_bound}
\\
\|\mathbf{S}\|_2 \;\le\; \|\mathbf{Y}\|_2 + \|\mathrm{FFN}(\mathbf{Y})\|_2 
&\;\le\; \sqrt{L\,d_V}\;\Big(1+\sqrt{\min(L,d_{ff})}\,\|\mathbf{W}_1\|_2\,\|\mathbf{W}_2\|_2\Big).
\label{eq:S_norm_bound}
\end{align}
\end{proposition}

\begin{proof}
We proceed using only the properties stated in the preliminaries.

1) Bound for $\|\mathbf{Y}\|_2$.
By the LayerNorm definition (Theorem~\ref{thm:layernorm_derivative}), write
\[\mathbf{Y} \;=\; \mathbf{P}(\mathbf{S}_0)\,\mathbf{M}(\mathbf{S}_0), \qquad \mathbf{S}_0:=\mathrm{SA}(\mathbf{X})+\mathbf{X},\]
where $\mathbf{M}(\mathbf{S}_0)=\mathbf{S}_0-\tfrac{1}{d_V}\mathbf{S}_0\mathbf{1}_{d_V}\mathbf{1}_{d_V}^\top$ and 
$\mathbf{P}=\mathrm{diag}^{-1}(\sigma)$ with $\sigma=\tfrac{1}{\sqrt{d_V}}(\mathbf{M}^{\circ 2}\mathbf{1})^{\circ 1/2}$ applied row-wise.
For any row $i$, denote $\mathbf{m}_i$ the $i$-th row of $\mathbf{M}$ and $\sigma_i=\tfrac{1}{\sqrt{d_V}}\|\mathbf{m}_i\|_2$. Then the $i$-th row of $\mathbf{Y}$ is $\mathbf{y}_i=\mathbf{m}_i/\sigma_i$, so
\[\|\mathbf{y}_i\|_2^2 \;=\; \frac{\|\mathbf{m}_i\|_2^2}{\sigma_i^2}\;=\; \frac{\|\mathbf{m}_i\|_2^2}{(1/d_V)\,\|\mathbf{m}_i\|_2^2}\;=\; d_V.\]
Hence every row of $\mathbf{Y}$ has Euclidean norm $\sqrt{d_V}$. Therefore,
\[\|\mathbf{Y}\|_F^2 = \sum_{i=1}^L \|\mathbf{y}_i\|_2^2 = L\,d_V,\qquad\text{so}\qquad\|\mathbf{Y}\|_F = \sqrt{L\,d_V}.\]
By the norm inequality $\|\mathbf{A}\|_2 \le \|\mathbf{A}\|_F$ (Property~\ref{prop:matrix_norm_inequalities}), we obtain \eqref{eq:Y_norm_bound}.

2) Bound for $\|\mathrm{FFN}(\mathbf{Y})\|_2$.
We estimate step-by-step using only matrix norm properties.

First,
\[\|\mathrm{FFN}(\mathbf{Y})\|_2= \|\text{ReLU}(\mathbf{Y}\mathbf{W}_1)\mathbf{W}_2\|_2\;\le\; \|\text{ReLU}(\mathbf{Y}\mathbf{W}_1)\|_2 \,\|\mathbf{W}_2\|_2\qquad\text{(Property~\ref{prop:matrix_product_norm})}.\]
Next, use $\|\cdot\|_2 \le \|\cdot\|_F$ (Property~\ref{prop:matrix_norm_inequalities}) to get
\[\|\text{ReLU}(\mathbf{Y}\mathbf{W}_1)\|_2 \le \|\text{ReLU}(\mathbf{Y}\mathbf{W}_1)\|_F.\]
By Definition~\ref{def:matrix_norms}, $\|\cdot\|_F^2$ is the sum of squares. Entrywise $\sigma(\cdot)$ satisfies $0\le \sigma(a)\le |a|$, hence $\sigma(a)^2 \le a^2$ for each entry $a \in \mathbb{R}$. Therefore,
\[\|\sigma(\mathbf{Y}\mathbf{W}_1)\|_F \le \|\mathbf{Y}\mathbf{W}_1\|_F.\]
Using the inequality $\|\cdot\|_F \le \sqrt{d}\,\|\cdot\|_2$ with $d=\operatorname{rank}(\cdot)$ from Property~\ref{prop:matrix_norm_inequalities} (row $X=\|\cdot\|_F$, column $Y=\|\cdot\|_2$), we obtain
\[\|\mathbf{Y}\mathbf{W}_1\|_F \le \sqrt{\operatorname{rank}(\mathbf{Y}\mathbf{W}_1)}\,\|\mathbf{Y}\mathbf{W}_1\|_2.\]
Since $\mathbf{Y}\mathbf{W}_1 \in \mathbb{R}^{L\times d_{ff}}$, $\operatorname{rank}(\mathbf{Y}\mathbf{W}_1)\le \min(L,d_{ff})$. Thus
\[\|\mathbf{Y}\mathbf{W}_1\|_F \le \sqrt{\min(L,d_{ff})}\,\|\mathbf{Y}\mathbf{W}_1\|_2\le \sqrt{\min(L,d_{ff})}\,\|\mathbf{Y}\|_2 \,\|\mathbf{W}_1\|_2\qquad\text{(Property~\ref{prop:matrix_product_norm})}.\]
Collecting,
\[\|\mathrm{FFN}(\mathbf{Y})\|_2\le \|\sigma(\mathbf{Y}\mathbf{W}_1)\|_F \,\|\mathbf{W}_2\|_2\le \sqrt{\min(L,d_{ff})}\;\|\mathbf{Y}\|_2\,\|\mathbf{W}_1\|_2\,\|\mathbf{W}_2\|_2,\]
which is \eqref{eq:FFN_norm_bound}.

3) Bound for $\|\mathbf{S}\|_2$.
By the sum-norm inequality (Property~\ref{prop:matrix_sum_norm}),
\[\|\mathbf{S}\|_2=\|\mathbf{Y}+\mathrm{FFN}(\mathbf{Y})\|_2\le \|\mathbf{Y}\|_2 + \|\mathrm{FFN}(\mathbf{Y})\|_2.\]
Substituting \eqref{eq:Y_norm_bound} and \eqref{eq:FFN_norm_bound} yields \eqref{eq:S_norm_bound}.
\end{proof}

\begin{lemma} [LayerNorm derivative and Hessian norm estimation] \label{lemma:layernorm_deriv_hessian_norm}

Let $\mathbf{X} \in \mathbb{R}^{m \times n}$. LayerNorm derivative $\mathbf{J}_{\mathrm{LN}}(\mathbf{X}) = \frac{\partial \text{LayerNorm}(\mathbf{X})}{\partial \mathbf{X}}$ is calculated according to Theorem \ref{thm:layernorm_derivative} and its Hessian $\mathbf{H}_{\mathrm{LN}}(\mathbf{X}) = \frac{\partial^2 \text{LayerNorm}(\mathbf{X})}{\partial \mathbf{X}^2}$ is calculated as in Theorem \ref{thm:layernorm_second_derivative}. Then, the following estimation holds:

\begin{align}
\big\|\mathbf{J}_{\mathrm{LN}}(\mathbf{X})\big\|_2
&\le \frac{1}{\sigma_{\min}} \;+\; \frac{\|\mathbf{X}\|_2^2}{\sqrt{n}\,\sigma_{\min}^3}, \label{eq:ln_jac_norm_bound}\\[4pt]
\big\|\mathbf{H}_{\mathrm{LN}}(\mathbf{X})\big\|_2
&\le \frac{\|\mathbf{X}\|_2}{\sigma_{\min}^3}\Big(1+\sqrt{\tfrac{m}{n}}\Big)
\;+\; \frac{\|\mathbf{X}\|_2^2}{\sqrt{n}\,\sigma_{\min}^3}
\;+\; \frac{3\,\|\mathbf{X}\|_2^3}{n\,\sigma_{\min}^5}.
\label{eq:ln_hess_norm_bound}
\end{align}

where $\sigma_{\min}$ denotes $\min \limits_i \|\mathbf{M}_i \|_2$, where $\mathbf{M}(\mathbf{X}) = \mathbf{X}\,(\mathbf{I}_n - \tfrac{1}{n}\mathbf{1}_n\mathbf{1}_n^\top)$
    
\end{lemma}

\begin{proof}
We rely only on the properties established in the preliminaries and on Theorems~\ref{thm:layernorm_derivative}--\ref{thm:layernorm_second_derivative}.

1) LayerNorm Jacobian structure and bound.
By Theorem~\ref{thm:layernorm_derivative} (with $L\!\to\! m$, $d_V\!\to\! n$),
\[\mathbf{J}_{\mathrm{LN}}(\mathbf{X})= (\mathbf{P}\otimes \mathbf{I}_n)\,\mathbf{G}+ (\mathbf{I}_m\otimes \mathbf{M}^\top)\,\mathbf{H},\]
where $\mathbf{G}=\mathbf{I}_{mn}-\tfrac{1}{n}(\mathbf{I}_m\otimes \mathbf{1}_{n\times n})$,
$\mathbf{H}=\tfrac{\partial \mathbf{P}}{\partial \mathbf{X}}$, and $\mathbf{P}=\mathrm{diag}^{-1}(\boldsymbol{\sigma})$.
Using Properties~\ref{prop:kronecker_product_norm}, \ref{prop:matrix_product_norm}, \ref{prop:matrix_sum_norm},
\[\|\mathbf{J}_{\mathrm{LN}}(\mathbf{X})\|_2\le \|\mathbf{P}\otimes \mathbf{I}_n\|_2\,\|\mathbf{G}\|_2 + \|\mathbf{I}_m\otimes \mathbf{M}^\top\|_2\,\|\mathbf{H}\|_2= \|\mathbf{P}\|_2\,\|\mathbf{G}\|_2 + \|\mathbf{M}\|_2\,\|\mathbf{H}\|_2.\]
We now bound each factor:

- $\|\mathbf{G}\|_2\le 1$ since $\tfrac{1}{n}\mathbf{1}_{n\times n}$ is a projection, hence $\|\mathbf{I}_n-\tfrac{1}{n}\mathbf{1}_{n\times n}\|_2\le 1$ and Kronecker preserves the spectral norm bound (Properties~\ref{prop:matrix_product_norm}, \ref{prop:kronecker_product_norm}, Proposition~\ref{prop:1_spectral_norm}).

- $\|\mathbf{P}\|_2 = \|\mathbf{D}^{-1}\|_2 = 1/\sigma_{\min}$, where $\mathbf{D}=\mathrm{diag}(\boldsymbol{\sigma})$.

- $\|\mathbf{M}\|_2 \le \|\mathbf{X}\|_2$, because $\mathbf{M}(\mathbf{X}) = \mathbf{X}\,(\mathbf{I}_n - \tfrac{1}{n}\mathbf{1}_n\mathbf{1}_n^\top)$ and the right factor is a projector with norm $\le 1$ (Property~\ref{prop:matrix_product_norm}).

- For $\|\mathbf{H}\|_2=\big\|\tfrac{\partial \mathbf{P}}{\partial \mathbf{X}}\big\|_2$, Theorem~\ref{thm:layernorm_derivative} plus Propositions~\ref{prop:invert_derivative}, \ref{prop:diag_derivative}, \ref{prop:hadamard_square_derivative}, \ref{prop:hadamard_root_derivative} and Properties~\ref{prop:matrix_product_norm}, \ref{prop:kronecker_product_norm} give (see the same chain as in Theorem~\ref{thm:layernorm_derivative}):
\[
\Big\|\frac{\partial \mathbf{P}}{\partial \mathbf{X}}\Big\|_2
\le \frac{1}{\sqrt{n}}\;\|\mathbf{D}^{-1}\otimes \mathbf{D}^{-\top}\|_2\,
\Big\|\mathrm{diag}^{-1}\!\big(\mathrm{vec}_r^{\circ 1/2}(\mathbf{M}^{\circ 2}\mathbf{1}_n)\big)\Big\|_2\,
\|\mathbf{I}_m\otimes \mathbf{1}_n^\top\|_2\,\|\mathrm{diag}(\mathrm{vec}_r(\mathbf{M}))\|_2\,
\Big\|\frac{\partial \mathbf{M}}{\partial \mathbf{X}}\Big\|_2.
\]
Using $\|\mathbf{D}^{-1}\otimes \mathbf{D}^{-\top}\|_2=\|\mathbf{D}^{-1}\|_2^2=\frac{1}{\sigma_{\min}^{2}}$, 
$\big\|\mathrm{diag}^{-1}(\cdot)\big\|_2 = \frac{1}{\min_i\sqrt{\sum_{v} M_{i,v}^2}} = \frac{1}{\sqrt{n}\,\sigma_{\min}}$,\newline
$\|\mathbf{I}_m\otimes \mathbf{1}^\top\|_2=\sqrt{n}$,
$\|\mathrm{diag}(\mathrm{vec}_r(\mathbf{M}))\|_2=\|\mathbf{M}\|_{\max}\le \|\mathbf{M}\|_2$ (Property~\ref{prop:matrix_norm_inequalities}),
and $\big\|\tfrac{\partial \mathbf{M}}{\partial \mathbf{X}}\big\|_2\le 1$ (projection), we obtain
\[\|\mathbf{H}\|_2 \;\le\; \frac{1}{\sqrt{n}\sigma_{\min}^{2}}\cdot \frac{1}{\sqrt{n}\,\sigma_{\min}}\cdot \sqrt{n}\cdot \|\mathbf{M}\|_2\cdot 1\;\le\; \frac{\|\mathbf{X}\|_2}{\sqrt{n}\,\sigma_{\min}^3}.\]
Collecting the bounds gives \eqref{eq:ln_jac_norm_bound}:
\[
\|\mathbf{J}_{\mathrm{LN}}(\mathbf{X})\|_2
\le \frac{1}{\sigma_{\min}}\cdot 1 + \|\mathbf{X}\|_2\cdot \frac{\|\mathbf{X}\|_2}{\sqrt{n}\,\sigma_{\min}^3}
= \frac{1}{\sigma_{\min}} + \frac{\|\mathbf{X}\|_2^2}{\sqrt{n}\,\sigma_{\min}^3}.
\]

2) LayerNorm Hessian structure and bound.
From Theorem~\ref{thm:layernorm_second_derivative} (with $m,n$), using $\tfrac{\partial^2 \mathbf{M}}{\partial \mathbf{X}^2}=0$,
\[\mathbf{H}_{\mathrm{LN}}(\mathbf{X})=(\mathbf{I}_{mn}\otimes \mathbf{G}^\top)\,\frac{\partial (\mathbf{P}\otimes \mathbf{I}_n)}{\partial \mathbf{X}}+ \big( (\mathbf{I}_m\otimes \mathbf{M}^\top)\otimes \mathbf{I}_{mn}\big)\,\frac{\partial^2 \mathbf{P}}{\partial \mathbf{X}^2}+ (\mathbf{I}_{mn}\otimes \mathbf{H}^\top)\,\frac{\partial (\mathbf{I}_m\otimes \mathbf{M}^\top)}{\partial \mathbf{X}}.\]
We bound the three terms separately with Properties~\ref{prop:matrix_product_norm}, \ref{prop:kronecker_product_norm}.

(i) First term. By Proposition~\ref{prop:kronecker_product_derivative},
\[\frac{\partial (\mathbf{P}\otimes \mathbf{I}_n)}{\partial \mathbf{X}}=(\mathbf{I}_m \otimes \mathbf{K}_{n,m} \otimes \mathbf{I}_n)\,(\mathbf{I}_{m^2} \otimes \mathrm{vec}_r(\mathbf{I}_n))\,\frac{\partial \mathbf{P}}{\partial \mathbf{X}},\]
therefore
\[\Big\|(\mathbf{I}_{mn}\otimes \mathbf{G}^\top)\,\frac{\partial (\mathbf{P}\otimes \mathbf{I}_n)}{\partial \mathbf{X}}\Big\|_2\le \|\mathbf{G}\|_2 \,\|\mathbf{I}_{m^2} \otimes \mathrm{vec}_r(\mathbf{I}_n)\|_2 \,\Big\|\frac{\partial \mathbf{P}}{\partial \mathbf{X}}\Big\|_2= 1 \cdot \sqrt{n} \cdot \frac{\|\mathbf{X}\|_2}{\sqrt{n}\,\sigma_{\min}^3}= \frac{\|\mathbf{X}\|_2}{\sigma_{\min}^3}.\]

(ii) Second term. Using $\|\mathbf{I}_m\otimes \mathbf{M}^\top\|_2=\|\mathbf{M}\|_2\le \|\mathbf{X}\|_2$ and the bound below for $\big\|\tfrac{\partial^2 \mathbf{P}}{\partial \mathbf{X}^2}\big\|_2$,
\[\Big\|\big( (\mathbf{I}_m\otimes \mathbf{M}^\top)\otimes \mathbf{I}_{mn}\big)\,\frac{\partial^2 \mathbf{P}}{\partial \mathbf{X}^2}\Big\|_2\le \|\mathbf{X}\|_2 \,\Big\|\frac{\partial^2 \mathbf{P}}{\partial \mathbf{X}^2}\Big\|_2.\]
We now bound $\big\|\tfrac{\partial^2 \mathbf{P}}{\partial \mathbf{X}^2}\big\|_2$ following the same chain as in the proof of Theorem~\ref{thm:layernorm_second_derivative}:
write $\tfrac{\partial \mathbf{P}}{\partial \mathbf{X}}=\tfrac{1}{\sqrt{n}}\mathbf{A}_1(\mathbf{X})\,\mathbf{E}\,\mathbf{B}_1(\mathbf{X})$ and differentiate using Property~\ref{prop:matrix_product_norm}, while bounding the factors with Propositions~\ref{prop:invert_derivative}, \ref{prop:diag_derivative}, \ref{prop:hadamard_square_derivative}, \ref{prop:hadamard_root_derivative} and Properties~\ref{prop:matrix_product_norm}, \ref{prop:kronecker_product_norm}, \ref{prop:matrix_norm_inequalities}. This yields
\[\Big\|\frac{\partial^2 \mathbf{P}}{\partial \mathbf{X}^2}\Big\|_2\;\le\;\frac{1}{\sqrt{n}\,\sigma_{\min}^3}\,\|\mathbf{X}\|_2\;+\; \frac{3}{n\,\sigma_{\min}^5}\,\|\mathbf{X}\|_2^2.\]
Therefore,
\[\Big\|\big( (\mathbf{I}_m\otimes \mathbf{M}^\top)\otimes \mathbf{I}_{mn}\big)\,\frac{\partial^2 \mathbf{P}}{\partial \mathbf{X}^2}\Big\|_2\le \frac{\|\mathbf{X}\|_2^2}{\sqrt{n}\,\sigma_{\min}^3}+ \frac{3\,\|\mathbf{X}\|_2^3}{n\,\sigma_{\min}^5}.\]

(iii) Third term. By Proposition~\ref{prop:kronecker_product_derivative} and Proposition~\ref{prop:transposed_matrix_derivative},
\[\frac{\partial (\mathbf{I}_m\otimes \mathbf{M}^\top)}{\partial \mathbf{X}}=(\mathbf{I}_m \otimes \mathbf{K}_{n,m} \otimes \mathbf{I}_m)\,(\mathrm{vec}_r(\mathbf{I}_m)\otimes \mathbf{I}_{mn})\,\frac{\partial \mathbf{M}}{\partial \mathbf{X}},\]
so
\[\Big\|(\mathbf{I}_{mn}\otimes \mathbf{H}^\top)\,\frac{\partial (\mathbf{I}_m\otimes \mathbf{M}^\top)}{\partial \mathbf{X}}\Big\|_2\le \|\mathbf{H}\|_2 \,\|\mathrm{vec}_r(\mathbf{I}_m)\otimes \mathbf{I}_{mn}\|_2 \,\Big\|\frac{\partial \mathbf{M}}{\partial \mathbf{X}}\Big\|_2= \frac{\|\mathbf{X}\|_2}{\sqrt{n}\,\sigma_{\min}^3}\cdot \sqrt{m}\cdot 1= \frac{\sqrt{m}}{\sqrt{n}}\,\frac{\|\mathbf{X}\|_2}{\sigma_{\min}^3}.\]

Summing (i)–(iii) with Property~\ref{prop:matrix_sum_norm} yields \eqref{eq:ln_hess_norm_bound}:
\[
\|\mathbf{H}_{\mathrm{LN}}(\mathbf{X})\|_2
\le \frac{\|\mathbf{X}\|_2}{\sigma_{\min}^3}
+ \Big(\frac{\|\mathbf{X}\|_2^2}{\sqrt{n}\,\sigma_{\min}^3}
+ \frac{3\,\|\mathbf{X}\|_2^3}{n\,\sigma_{\min}^5}\Big)
+ \frac{\sqrt{m}}{\sqrt{n}}\,\frac{\|\mathbf{X}\|_2}{\sigma_{\min}^3}
=
\frac{\|\mathbf{X}\|_2}{\sigma_{\min}^3}\Big(1+\sqrt{\tfrac{m}{n}}\Big)
+ \frac{\|\mathbf{X}\|_2^2}{\sqrt{n}\,\sigma_{\min}^3}
+ \frac{3\,\|\mathbf{X}\|_2^3}{n\,\sigma_{\min}^5}.
\]
This completes the proof.
\end{proof}
%%%%%%%%%%%%%%%%%%%%%%%%%%%%%%%%%%%%%%%%%%%%%%%%%%%%%%%%%%%%

\newpage

\end{document}